\documentclass[10pt,twocolumn,letterpaper]{article}

\usepackage{cvpr}                           
\usepackage[table]{xcolor}
\usepackage[accsupp]{axessibility}
\usepackage{pgf}
\usepackage{multirow}
\usepackage{pifont}
\usepackage{pgfplots}
\pgfplotsset{compat=1.17}
\usepackage{pgfplotstable}
\usepackage{tikz}
\usepackage{csvsimple}
\usepackage{graphicx}
\usepackage{xifthen}
\usepackage{arydshln}
\usepackage{adjustbox}
\usepackage{balance}
\usepackage{tcolorbox}
\usepackage{etoolbox}
\usepackage{adjustbox}
\usepackage{nicefrac}
\usetikzlibrary{intersections}
\usepgfplotslibrary{fillbetween}
\usetikzlibrary{shapes,calc,positioning, decorations.text, arrows.meta, calc, shadows.blur, shadings, fit}
\usepackage{ragged2e} 

\definecolor{cvprblue}{rgb}{0.21,0.49,0.74}
\definecolor{cvprpink}{rgb}{1.0,0.4,0.7}
\usepackage{url}
\usepackage[pagebackref,breaklinks,colorlinks,allcolors=cvprblue,urlcolor=cvprpink]{hyperref}

\def\paperID{13224} 
\def\confName{CVPR}
\def\confYear{2026}

\title{Indexing Multimodal Language Models for Large-scale Image Retrieval}

\author{
Bahey Tharwat$^{1}$\thanks{Equal contribution.}
\quad
Giorgos Kordopatis-Zilos$^{2}$\footnotemark[1]
\quad
Pavel Suma$^{2}$
\quad
Ian Reid$^{1}$
\quad
Giorgos Tolias$^{2}$\\
$^{1}$Mohamed bin Zayed University of Artificial Intelligence, Abu Dhabi, UAE\\
$^{2}$VRG, FEE, Czech Technical University in Prague, Prague, Czech Republic
\vspace{-1pt}
}

\begin{document}
\maketitle

\newcommand{\ione}{i\hspace{-.05em}+\hspace{-.07em}1}

\newcommand{\mypartight}[1]{\noindent {\bf #1}}
\newcommand{\myparagraph}[1]{\noindent\textbf{#1}\xspace}

\newcommand{\optional}[1]{{#1}}
\newcommand{\alert}[1]{{\color{red}{#1}}}

\newcommand{\gt}[1]{{\color{purple}{GT: #1}}}
\newcommand{\gtt}[1]{{\color{purple}{#1}}}
\newcommand{\gtr}[2]{{\color{purple}\st{#1} {#2}}}

\newcommand{\gkz}[1]{{\color{cyan}{GKZ: #1}}}
\newcommand{\gkzt}[1]{{\color{cyan}{#1}}}
\newcommand{\gkzr}[2]{{\color{cyan}\st{#1} {#2}}}

\newcommand{\ps}[1]{{\color{brown}{PS: #1}}}
\newcommand{\pst}[1]{{\color{brown}{#1}}}
\newcommand{\psr}[2]{{\color{brown}\st{#1} {#2}}}

\newcommand{\am}[1]{{\color{orange}{AM: #1}}}
\newcommand{\amt}[1]{{\color{orange}{#1}}}
\newcommand{\amr}[2]{{\color{orange}\st{#1} {#2}}}

\newcommand{\och}[1]{{\color{blue}{OCh: #1}}}
\newcommand{\ocht}[1]{{\color{blue}{#1}}}
\newcommand{\ochr}[2]{{\color{blue}\st{#1} {#2}}}

\newcommand{\gray}[1]{{\color{gray}{#1}}}
\definecolor{lightgray}{gray}{0.87} 

\newcolumntype{Y}{>{\centering\arraybackslash}p{4em}}

\def\roxf{$\mathcal{R}$Oxford\xspace}
\def\rox{$\mathcal{R}$Oxf\xspace}
\def\ro{$\mathcal{R}$O\xspace}
\def\rpar{$\mathcal{R}$Paris\xspace}
\def\rpa{$\mathcal{R}$Par\xspace}
\def\rp{$\mathcal{R}$P\xspace}
\def\rdis{$\mathcal{R}$1M\xspace}
\def\rop{$\mathcal{R}$OP+1M\xspace}
\def\gld{GLDv2\xspace}
\def\ilias{\mbox{ILIAS}\xspace}
\def\instre{\mbox{INSTRE}\xspace}
\def\prodonem{\mbox{Prod1M}\xspace}
\def\food{\mbox{Food2k}\xspace}
\def\inat{\mbox{iNaturalist}\xspace}

\newcommand\resnet[3]{\ensuremath{\prescript{#2}{}{\mathtt{R}}{#1}_{\scriptscriptstyle #3}}\xspace}

\newcommand{\ames}{\mbox{AMES}\xspace} %
\newcommand{\chamfer}{\mbox{Chamfer}\xspace} %
\newcommand{\rtf}{$R^2$Former\xspace}
\newcommand{\rrt}{RRT\xspace} %
\newcommand{\cvnet}{CVNet\xspace} %

\newcommand{\stddev}[1]{\scriptsize{$\pm#1$}}

\newcommand{\diffup}[1]{{\color{OliveGreen}{($\uparrow$ #1)}}}
\newcommand{\diffdown}[1]{{\color{BrickRed}{($\downarrow$ #1)}}}

\def\nmsp{\hspace{-6pt}}
\def\nssp{\hspace{-3pt}}
\def\nxssp{\hspace{-1pt}}
\def\zsp{\hspace{0pt}}
\def\xssp{\hspace{1pt}}
\def\ssp{\hspace{3pt}}
\def\msp{\hspace{6pt}}
\def\mlsp{\hspace{9pt}}
\def\lsp{\hspace{12pt}}
\def\xlsp{\hspace{20pt}}

\newcommand{\head}[1]{{\smallskip\noindent\bf #1}}
\newcommand{\equ}[1]{(\ref{equ:#1})\xspace}

\newcommand{\nn}[1]{\ensuremath{\text{NN}_{#1}}\xspace}
\def\l1{\ensuremath{\ell_1}\xspace}
\def\l2{\ensuremath{\ell_2}\xspace}

\newcommand{\tran}{^\top}
\newcommand{\mtran}{^{-\top}}
\newcommand{\zcol}{\mathbf{0}}
\newcommand{\zrow}{\zcol\tran}

\newcommand{\ind}{\mathds{1}}
\newcommand{\expect}{\mathbb{E}}
\newcommand{\nat}{\mathbb{N}}
\newcommand{\zahl}{\mathbb{Z}}
\newcommand{\real}{\mathbb{R}}
\newcommand{\proj}{\mathbb{P}}
\newcommand{\prob}{\mathbf{Pr}}

\newcommand{\mif}{\textrm{if }}
\newcommand{\other}{\textrm{otherwise}}
\newcommand{\minimize}{\textrm{minimize }}
\newcommand{\maximize}{\textrm{maximize }}

\newcommand{\id}{\operatorname{id}}
\newcommand{\const}{\operatorname{const}}
\newcommand{\sgn}{\operatorname{sgn}}
\newcommand{\erf}{\operatorname{erf}}
\newcommand{\var}{\operatorname{Var}}
\newcommand{\mean}{\operatorname{mean}}
\newcommand{\trace}{\operatorname{tr}}
\newcommand{\diag}{\operatorname{diag}}
\newcommand{\vect}{\operatorname{vec}}
\newcommand{\cov}{\operatorname{cov}}

\newcommand{\softmax}{\operatorname{softmax}}
\newcommand{\clip}{\operatorname{clip}}

\newcommand{\defn}{\mathrel{:=}}
\newcommand{\peq}{\mathrel{+\!=}}
\newcommand{\meq}{\mathrel{-\!=}}

\newcommand{\floor}[1]{\left\lfloor{#1}\right\rfloor}
\newcommand{\ceil}[1]{\left\lceil{#1}\right\rceil}
\newcommand{\inner}[1]{\left\langle{#1}\right\rangle}
\newcommand{\norm}[1]{\left\|{#1}\right\|}
\newcommand{\frob}[1]{\norm{#1}_F}
\newcommand{\card}[1]{\left|{#1}\right|\xspace}
\newcommand{\diff}{\mathrm{d}}
\newcommand{\der}[3][]{\frac{d^{#1}#2}{d#3^{#1}}}
\newcommand{\pder}[3][]{\frac{\partial^{#1}{#2}}{\partial{#3^{#1}}}}
\newcommand{\ipder}[3][]{\partial^{#1}{#2}/\partial{#3^{#1}}}
\newcommand{\dder}[3]{\frac{\partial^2{#1}}{\partial{#2}\partial{#3}}}

\newcommand{\wb}[1]{\overline{#1}}
\newcommand{\wt}[1]{\widetilde{#1}}

\newcommand{\cA}{\mathcal{A}}
\newcommand{\cB}{\mathcal{B}}
\newcommand{\cC}{\mathcal{C}}
\newcommand{\cD}{\mathcal{D}}
\newcommand{\cE}{\mathcal{E}}
\newcommand{\cF}{\mathcal{F}}
\newcommand{\cG}{\mathcal{G}}
\newcommand{\cH}{\mathcal{H}}
\newcommand{\cI}{\mathcal{I}}
\newcommand{\cJ}{\mathcal{J}}
\newcommand{\cK}{\mathcal{K}}
\newcommand{\cL}{\mathcal{L}}
\newcommand{\cM}{\mathcal{M}}
\newcommand{\cN}{\mathcal{N}}
\newcommand{\cO}{\mathcal{O}}
\newcommand{\cP}{\mathcal{P}}
\newcommand{\cQ}{\mathcal{Q}}
\newcommand{\cR}{\mathcal{R}}
\newcommand{\cS}{\mathcal{S}}
\newcommand{\cT}{\mathcal{T}}
\newcommand{\cU}{\mathcal{U}}
\newcommand{\cV}{\mathcal{V}}
\newcommand{\cW}{\mathcal{W}}
\newcommand{\cX}{\mathcal{X}}
\newcommand{\cY}{\mathcal{Y}}
\newcommand{\cZ}{\mathcal{Z}}

\newcommand{\vA}{\mathbf{A}}
\newcommand{\vB}{\mathbf{B}}
\newcommand{\vC}{\mathbf{C}}
\newcommand{\vD}{\mathbf{D}}
\newcommand{\vE}{\mathbf{E}}
\newcommand{\vF}{\mathbf{F}}
\newcommand{\vG}{\mathbf{G}}
\newcommand{\vH}{\mathbf{H}}
\newcommand{\vI}{\mathbf{I}}
\newcommand{\vJ}{\mathbf{J}}
\newcommand{\vK}{\mathbf{K}}
\newcommand{\vL}{\mathbf{L}}
\newcommand{\vM}{\mathbf{M}}
\newcommand{\vN}{\mathbf{N}}
\newcommand{\vO}{\mathbf{O}}
\newcommand{\vP}{\mathbf{P}}
\newcommand{\vQ}{\mathbf{Q}}
\newcommand{\vR}{\mathbf{R}}
\newcommand{\vS}{\mathbf{S}}
\newcommand{\vT}{\mathbf{T}}
\newcommand{\vU}{\mathbf{U}}
\newcommand{\vV}{\mathbf{V}}
\newcommand{\vW}{\mathbf{W}}
\newcommand{\vX}{\mathbf{X}}
\newcommand{\vY}{\mathbf{Y}}
\newcommand{\vZ}{\mathbf{Z}}

\newcommand{\va}{\mathbf{a}}
\newcommand{\vb}{\mathbf{b}}
\newcommand{\vc}{\mathbf{c}}
\newcommand{\vd}{\mathbf{d}}
\newcommand{\ve}{\mathbf{e}}
\newcommand{\vf}{\mathbf{f}}
\newcommand{\vg}{\mathbf{g}}
\newcommand{\vh}{\mathbf{h}}
\newcommand{\vi}{\mathbf{i}}
\newcommand{\vj}{\mathbf{j}}
\newcommand{\vk}{\mathbf{k}}
\newcommand{\vl}{\mathbf{l}}
\newcommand{\vm}{\mathbf{m}}
\newcommand{\vn}{\mathbf{n}}
\newcommand{\vo}{\mathbf{o}}
\newcommand{\vp}{\mathbf{p}}
\newcommand{\vq}{\mathbf{q}}
\newcommand{\vr}{\mathbf{r}}
\newcommand{\Vs}{\mathbf{s}}
\newcommand{\vt}{\mathbf{t}}
\newcommand{\vu}{\mathbf{u}}
\newcommand{\vv}{\mathbf{v}}
\newcommand{\vw}{\mathbf{w}}
\newcommand{\vx}{\mathbf{x}}
\newcommand{\vy}{\mathbf{y}}
\newcommand{\vz}{\mathbf{z}}

\newcommand{\vone}{\mathbf{1}}
\newcommand{\vzero}{\mathbf{0}}

\newcommand{\valpha}{{\boldsymbol{\alpha}}}
\newcommand{\vbeta}{{\boldsymbol{\beta}}}
\newcommand{\vgamma}{{\boldsymbol{\gamma}}}
\newcommand{\vdelta}{{\boldsymbol{\delta}}}
\newcommand{\vepsilon}{{\boldsymbol{\epsilon}}}
\newcommand{\vzeta}{{\boldsymbol{\zeta}}}
\newcommand{\veta}{{\boldsymbol{\eta}}}
\newcommand{\vtheta}{{\boldsymbol{\theta}}}
\newcommand{\viota}{{\boldsymbol{\iota}}}
\newcommand{\vkappa}{{\boldsymbol{\kappa}}}
\newcommand{\vlambda}{{\boldsymbol{\lambda}}}
\newcommand{\vmu}{{\boldsymbol{\mu}}}
\newcommand{\vnu}{{\boldsymbol{\nu}}}
\newcommand{\vxi}{{\boldsymbol{\xi}}}
\newcommand{\vomikron}{{\boldsymbol{\omikron}}}
\newcommand{\vpi}{{\boldsymbol{\pi}}}
\newcommand{\vrho}{{\boldsymbol{\rho}}}
\newcommand{\vsigma}{{\boldsymbol{\sigma}}}
\newcommand{\vtau}{{\boldsymbol{\tau}}}
\newcommand{\vupsilon}{{\boldsymbol{\upsilon}}}
\newcommand{\vphi}{{\boldsymbol{\phi}}}
\newcommand{\vchi}{{\boldsymbol{\chi}}}
\newcommand{\vpsi}{{\boldsymbol{\psi}}}
\newcommand{\vomega}{{\boldsymbol{\omega}}}

\newcommand{\rLambda}{\mathrm{\Lambda}}
\newcommand{\rSigma}{\mathrm{\Sigma}}

\makeatletter
\DeclareRobustCommand\onedot{\futurelet\@let@token\@onedot}
\def\@onedot{\ifx\@let@token.\else.\null\fi\xspace}
\def\eg{\emph{e.g}\onedot} \def\Eg{\emph{E.g}\onedot}
\def\ie{\emph{i.e}\onedot} \def\Ie{\emph{I.e}\onedot}
\def\vs{\emph{vs\onedot}}
\def\cf{\emph{cf}\onedot} \def\Cf{\emph{C.f}\onedot}
\def\etc{\emph{etc}\onedot} \def\vs{\emph{vs}\onedot}
\def\wrt{w.r.t\onedot} \def\dof{d.o.f\onedot}
\def\etal{\emph{et al}\onedot}
\makeatother

\newcommand\rurl[1]{%
  \href{https://#1}{\nolinkurl{#1}}%
}

\newcommand{\bentarrow}[1][]{%
  \begin{tikzpicture}[#1]%
    \draw (0,0.7ex) -- (0,0) -- (0.75em,0);
    \draw (0.55em,0.2em) -- (0.75em,0) -- (0.55em,-0.2em);
  \end{tikzpicture}%
}

\definecolor{higha}{HTML}{009b10} 
\definecolor{lowa}{HTML}{ec462e}  
\definecolor{mediuma}{HTML}{FFFFFF} 

\newcommand*{\opacitya}{50} 
\newcommand*{\minvalcolora}{2.5} 
\newcommand*{\midvalcolora}{11.6} 
\newcommand*{\maxvalcolora}{37.3} 
\newcommand{\grca}[1]{
    \ifdim #1pt < \midvalcolora pt
        \pgfmathparse{(#1-\minvalcolora)/(\midvalcolora-\minvalcolora)}
        \let\normalizedval\pgfmathresult
    
        \pgfmathparse{100*(\normalizedval)^(2.0)} 
        \xdef\tempa{\pgfmathresult}
        \pgfmathparse{min(100,max(0,\tempa))}
        \xdef\tempa{\pgfmathresult}
    
        \cellcolor{mediuma!\tempa!lowa!\opacitya} #1
    \else
        \pgfmathparse{(#1-\midvalcolora)/(\maxvalcolora-\midvalcolora)}
        \let\normalizedval\pgfmathresult
    
        \pgfmathparse{100*(\normalizedval)^(2.0)}
        \xdef\tempa{\pgfmathresult}
        \pgfmathparse{min(100,max(0,\tempa))}
        \xdef\tempa{\pgfmathresult}
    
        \cellcolor{higha!\tempa!mediuma!\opacitya} #1
    \fi
}

\definecolor{highb}{HTML}{009b10} 
\definecolor{lowb}{HTML}{ec462e}  
\definecolor{mediumb}{HTML}{FFFFFF} 

\newcommand*{\opacityb}{50} 
\newcommand*{\minvalcolorb}{1.8} 
\newcommand*{\midvalcolorb}{8.6} 
\newcommand*{\maxvalcolorb}{31.3} 
\newcommand{\grcb}[1]{
    \ifdim #1pt < \midvalcolorb pt
        \pgfmathparse{(#1-\minvalcolorb)/(\midvalcolorb-\minvalcolorb)}
        \let\normalizedval\pgfmathresult
    
        \pgfmathparse{100*(\normalizedval)^(2.0)} 
        \xdef\tempa{\pgfmathresult}
        \pgfmathparse{min(100,max(0,\tempa))}
        \xdef\tempa{\pgfmathresult}
    
        \cellcolor{mediumb!\tempa!lowb!\opacityb} #1
    \else
        \pgfmathparse{(#1-\midvalcolorb)/(\maxvalcolorb-\midvalcolorb)}
        \let\normalizedval\pgfmathresult
    
        \pgfmathparse{100*(\normalizedval)^(2.0)}
        \xdef\tempa{\pgfmathresult}
        \pgfmathparse{min(100,max(0,\tempa))}
        \xdef\tempa{\pgfmathresult}
    
        \cellcolor{highb!\tempa!mediumb!\opacityb} #1
    \fi
}

\definecolor{highc}{HTML}{009b10} 
\definecolor{lowc}{HTML}{ec462e}  
\definecolor{mediumc}{HTML}{FFFFFF} 

\newcommand*{\opacityc}{50} 
\newcommand*{\minvalcolorc}{1.7} 
\newcommand*{\midvalcolorc}{5.9} 
\newcommand*{\maxvalcolorc}{20.8} 
\newcommand{\grcc}[1]{
    \ifdim #1pt < \midvalcolorc pt
        \pgfmathparse{(#1-\minvalcolorc)/(\midvalcolorc-\minvalcolorc)}
        \let\normalizedval\pgfmathresult
    
        \pgfmathparse{100*(\normalizedval)^(2.0)} 
        \xdef\tempa{\pgfmathresult}
        \pgfmathparse{min(100,max(0,\tempa))}
        \xdef\tempa{\pgfmathresult}
    
        \cellcolor{mediumc!\tempa!lowc!\opacityc} #1
    \else
        \pgfmathparse{(#1-\midvalcolorc)/(\maxvalcolorc-\midvalcolorc)}
        \let\normalizedval\pgfmathresult
    
        \pgfmathparse{100*(\normalizedval)^(2.0)}
        \xdef\tempa{\pgfmathresult}
        \pgfmathparse{min(100,max(0,\tempa))}
        \xdef\tempa{\pgfmathresult}
    
        \cellcolor{highc!\tempa!mediumc!\opacityc} #1
    \fi
}

\definecolor{highd}{HTML}{009b10} 
\definecolor{lowd}{HTML}{ec462e}  
\definecolor{mediumd}{HTML}{FFFFFF} 

\newcommand*{\opacityd}{50} 
\newcommand*{\minvalcolord}{1.5} 
\newcommand*{\midvalcolord}{7.5} 
\newcommand*{\maxvalcolord}{19.8} 
\newcommand{\grcd}[1]{
    \ifdim #1pt < \midvalcolord pt
        \pgfmathparse{(#1-\minvalcolord)/(\midvalcolord-\minvalcolord)}
        \let\normalizedval\pgfmathresult
    
        \pgfmathparse{100*(\normalizedval)^(2.0)} 
        \xdef\tempa{\pgfmathresult}
        \pgfmathparse{min(100,max(0,\tempa))}
        \xdef\tempa{\pgfmathresult}
    
        \cellcolor{mediumd!\tempa!lowd!\opacityd} #1
    \else
        \pgfmathparse{(#1-\midvalcolord)/(\maxvalcolord-\midvalcolord)}
        \let\normalizedval\pgfmathresult
    
        \pgfmathparse{100*(\normalizedval)^(2.0)}
        \xdef\tempa{\pgfmathresult}
        \pgfmathparse{min(100,max(0,\tempa))}
        \xdef\tempa{\pgfmathresult}
    
        \cellcolor{highd!\tempa!mediumd!\opacityd} #1
    \fi
}

\definecolor{highe}{HTML}{009b10} 
\definecolor{lowe}{HTML}{ec462e}  
\definecolor{mediume}{HTML}{FFFFFF} 

\newcommand*{\opacitye}{50} 
\newcommand*{\minvalcolore}{2.3} 
\newcommand*{\midvalcolore}{10.6} 
\newcommand*{\maxvalcolore}{24.7} 
\newcommand{\grce}[1]{
    \ifdim #1pt < \midvalcolore pt
        \pgfmathparse{(#1-\minvalcolore)/(\midvalcolore-\minvalcolore)}
        \let\normalizedval\pgfmathresult
    
        \pgfmathparse{100*(\normalizedval)^(2.0)} 
        \xdef\tempa{\pgfmathresult}
        \pgfmathparse{min(100,max(0,\tempa))}
        \xdef\tempa{\pgfmathresult}
    
        \cellcolor{mediume!\tempa!lowe!\opacitye} #1
    \else
        \pgfmathparse{(#1-\midvalcolore)/(\maxvalcolore-\midvalcolore)}
        \let\normalizedval\pgfmathresult
    
        \pgfmathparse{100*(\normalizedval)^(2.0)}
        \xdef\tempa{\pgfmathresult}
        \pgfmathparse{min(100,max(0,\tempa))}
        \xdef\tempa{\pgfmathresult}
    
        \cellcolor{highe!\tempa!mediume!\opacitye} #1
    \fi
}

\definecolor{lightred}{RGB}{231, 76, 60}   
\definecolor{lightblue}{RGB}{54, 69, 79}
\definecolor{lightgreen}{RGB}{50, 205, 50} 

\definecolor{appleblue}{RGB}{80,122,255}  
\definecolor{applepink}{RGB}{217,127,174}  
\definecolor{appleteal}{RGB}{85,163,152}  
\definecolor{applepurple}{RGB}{138,107,221}  
\definecolor{applemustard}{RGB}{234,179,77}  
\definecolor{appleorange}{RGB}{255,127,80}  
\definecolor{applegreen}{RGB}{52,199,89}
\definecolor{appleyellow}{RGB}{255,204,0}
\definecolor{applered}{RGB}{255,59,48}  
\definecolor{appleindigo}{RGB}{94,92,230}  
\definecolor{applesand}{RGB}{210,180,140}  
\definecolor{appleslate}{RGB}{142,142,147}  

\definecolor{cvprblue}{rgb}{0.21,0.49,0.74}
\definecolor{materialPurple}{rgb}{0.615, 0.275, 1} 
\definecolor{plotlyBlue}{rgb}{0, 0.482, 1} 
\definecolor{plotlyGreen}{rgb}{0.157, 0.655, 0.271} 
\definecolor{plotlyRed}{rgb}{0.863, 0.208, 0.271} 
\definecolor{plotlyOrange}{rgb}{1, 0.757, 0.027} 
\definecolor{plotlyYellow}{rgb}{1, 0.843, 0} 
\definecolor{plotlyCyan}{rgb}{0.09, 0.745, 0.812} 
\definecolor{plotlyMagenta}{rgb}{0.917, 0.722, 0.894} 
\definecolor{plotlyTeal}{rgb}{0, 0.545, 0.455} 
\definecolor{plotlyNavy}{rgb}{0.4, 0.063, 0.949} 
\definecolor{plotlyPurple}{rgb}{0.58, 0.404, 0.741} 
\definecolor{plotlyBrown}{rgb}{0.82, 0.604, 0.416} 
\definecolor{plotlyPink}{rgb}{0.89, 0.467, 0.761} 
\definecolor{plotlyGray}{rgb}{0.498, 0.498, 0.498} 
\definecolor{plotlyLightBlue}{rgb}{0.122, 0.467, 0.706} 
\definecolor{plotlyLightOrange}{rgb}{1, 0.733, 0.471} 
\definecolor{plotlyLightGreen}{rgb}{0.596, 0.875, 0.541} 

\definecolor{calendarPurple}{rgb}{0.635, 0.349, 1.000} 
\definecolor{calendarRed}{rgb}{1.000, 0.420, 0.420} 
\definecolor{calendarPinkRed}{rgb}{1.000, 0.396, 0.518} 
\definecolor{calendarLightPink}{rgb}{1.000, 0.663, 0.663} 
\definecolor{calendarLightPurple}{rgb}{0.847, 0.627, 1.000} 
\definecolor{calendarDarkBlue}{rgb}{0.424, 0.435, 1.000} 
\definecolor{calendarBlue}{rgb}{0.000, 0.620, 1.000} 
\definecolor{calendarGreen}{rgb}{0.000, 0.714, 0.427} 
\definecolor{calendarLightMint}{rgb}{0.678, 0.910, 0.827} 
\definecolor{calendarYellow}{rgb}{1.000, 0.922, 0.231} 
\definecolor{calendarAmber}{rgb}{1.000, 0.757, 0.027} 
\definecolor{calendarOrange}{rgb}{1.000, 0.596, 0.000} 
\definecolor{calendarBrown}{rgb}{0.490, 0.353, 0.314} 
\definecolor{calendarGray}{rgb}{0.710, 0.710, 0.710} 
\definecolor{calendarGrayBlue}{rgb}{0.482, 0.604, 0.631} 
\definecolor{calendarLightBlue}{rgb}{0.647, 0.847, 0.867} 
\definecolor{calendarDarkPurple}{rgb}{0.416, 0.051, 0.678} 

\definecolor{pecol}{RGB}{234,179,77} 
\definecolor{qwencol}{RGB}{80,122,255}
\definecolor{indexcol}{RGB}{85,163,152}
\definecolor{intercol}{RGB}{142,142,147}
\definecolor{lamracol}{RGB}{217,127,174}
\definecolor{amescol}{RGB}{138,107,221}
\definecolor{qwenthreecol}{RGB}{255,127,80}
\definecolor{qwenrerankcol}{RGB}{255,59,48}
\definecolor{elviscol}{RGB}{210,180,140}

\pgfplotsset{
  normmark/.style={
    only marks,
    mark=*,
    mark size=3.1,
    opacity=1.0,
    line width=1.1pt,
    mark options={draw=black, line width=0.35pt}
  }
}

\pgfplotsset{
  pqmark/.style={
    only marks,
    mark=diamond*,
    mark size=4.2,
    opacity=1.0,
    line width=1.2pt,
    mark options={draw=black, line width=0.35pt}
  }
}

\pgfplotsset{
  line/.style={
    solid,
    line width=1.2pt,
  }
}

\pgfplotsset{
  markline/.style={
    solid,
    mark=*,
    mark size=3.1,
    opacity=1.0,
    line width=1.2pt,
    mark options={draw=black, line width=0.35pt}
  }
}

\pgfplotsset{
  pqline/.style={
    dashed,
    mark=diamond*,
    mark size=4.2,
    opacity=1.0,
    line width=1.2pt,
    mark options={solid, draw=black, line width=0.35pt}
  }
}

\pgfplotsset{
  clustmark/.style={
    only marks,
    mark=triangle*,
    mark size=4.2,
    opacity=1.0,
    line width=1.3pt,
    mark options={draw=black, line width=0.35pt}
  }
}

\pgfplotsset{
  prunmark/.style={
    only marks,
    mark=triangle*,
    mark size=4.2,
    opacity=1.0,
    line width=1.3pt,
    mark options={rotate=180, draw=black, line width=0.35pt}
  }
}

\pgfplotsset{
  poolmark/.style={
    only marks,
    mark=square*,
    mark size=3.0,
    opacity=1.0,
    line width=1.3pt,
    mark options={draw=black, line width=0.35pt}
  }
}

\pgfplotsset{
  samplemark/.style={
    only marks,
    mark=pentagon*,
    mark size=3.5,
    opacity=1.0,
    line width=1.3pt,
    mark options={draw=black, line width=0.35pt}
  }
}

\begin{abstract}
Multimodal Large Language Models (MLLMs) have demonstrated strong cross-modal reasoning capabilities, yet their potential for vision-only tasks remains underexplored. We investigate MLLMs as training-free similarity estimators for instance-level image-to-image retrieval. Our approach prompts the model with paired images and converts next-token probabilities into similarity scores, enabling zero-shot re-ranking within large-scale retrieval pipelines. This design avoids specialized architectures and fine-tuning, leveraging the rich visual discrimination learned during multimodal pre-training. We address scalability by combining MLLMs with memory-efficient indexing and top-$k$ candidate re-ranking. Experiments across diverse benchmarks show that MLLMs outperform task-specific re-rankers outside their native domains and exhibit superior robustness to clutter, occlusion, and small objects. Despite strong results, we identify failure modes under severe appearance changes, highlighting opportunities for future research. Our findings position MLLMs as a promising alternative for open-world large-scale image retrieval. 

\noindent code: \url{http://github.com/gkordo/mllm-ilr}
\vspace{-12pt}
\end{abstract}    
\section{Introduction}
\label{sec:intro}
Multimodal Large Language Models (MLLMs)~\cite{chatgpt, gemini, minigpt4, liu2023visual,llama3} have recently emerged as powerful tools for bridging vision and language. Their ability to jointly process images and text enables rich cross-modal reasoning. From a computer vision perspective, MLLMs offer a unique advantage: they use natural language both as supervision during training and as an interaction mechanism at inference, providing remarkable flexibility. This makes them suitable for diverse tasks such as visual question answering (VQA)~\cite{llava-vl, blip, flamingo}, image captioning~\cite{finecaption, nice}, object recognition~\cite{yolo-world, detclip, dino-x}, and scene understanding~\cite{llmformer, sodano2024open, kirillov2023segment}. In these settings, language serves as an additional source of context, enhancing interpretability and adaptability.

The appeal of MLLMs is twofold: (i) natural language supervision is expressive and diverse, enabling learning from large multimodal corpora without task-specific annotations, and (ii) text-based interaction allows zero-shot adaptation to novel tasks through prompting~\cite{llava-vl, flamingo}. This combination of scalable training and flexible deployment motivates their use beyond traditional multimodal tasks.

\begin{figure}[t!]
    \vspace{0pt}
    \centering
    \pgfplotsset{every tick label/.append style={font=\footnotesize}}

\newcommand{\threecolornode}[7]{
    \node [above] at (axis cs: #1, #2) (TCN) {\footnotesize \textcolor{#4}{#3}};
    \begin{scope}
        \clip (TCN.north east) -- (TCN.south east) -- (TCN.north west) -- cycle;
        \node at (TCN) {\footnotesize\color{#6}{#3}};
    \end{scope}
    \begin{scope}
        \clip (TCN.south west) -- (TCN.north west) -- (TCN.south east) -- cycle;
        \node at (TCN) {\footnotesize\color{#5}{#3}};
    \end{scope}
    \begin{scope}
        \clip ([xshift=#7]TCN.north west) -- ([xshift=#7]TCN.south east) -- 
              ([xshift=-#7]TCN.south east) -- ([xshift=-#7]TCN.north west) -- cycle;
        \node at (TCN) {\footnotesize\color{#4}{#3}};
    \end{scope}
}

\begin{tikzpicture}
\begin{axis}[%
  width=1\linewidth,
  height=0.55\linewidth,
  ylabel={mAP@1k},
  xlabel={time per query (s)},
  grid=both,
  grid style={color=lightgray!60, dash pattern=on 2pt off 2pt},
  xmode=log,
  xmode=log,
  ylabel near ticks, xlabel near ticks, 
  log ticks with fixed point,
  xlabel style={yshift=7pt},
  ymax=55,
  xmin=0.08,
  xmax=6,
  legend columns=1, 
  legend style={
    anchor=south, 
    at={(0.18,0.51)}, 
    cells={anchor=west}, 
    font=\footnotesize, 
    fill opacity=1, 
    inner sep=2pt
    },
  ]

        \addlegendimage{color=qwencol, markline} 
        \addlegendentry{Qwen-C}
        \addlegendimage{color=qwenthreecol, markline} 
        \addlegendentry{Qwen3-C}
        \addlegendimage{color=indexcol, dashed, markline}
        \addlegendentry{Qwen3R-C}
        \addlegendimage{color=amescol, markline} 
        \addlegendentry{AMES}

        \addplot[color=amescol, markline] coordinates {
                                (0.1,  39.38)
                                (0.2,  40.08)
                                (0.4,  40.75)
                                (1,    41.61)
                                (2.5,  42.12)
                                (5,    42.37)
        };
        
        \addplot[color=qwencol, markline] coordinates {
                                (0.124, 35.35)  
                                (0.217, 36.81)  
                                (0.403, 38.85)  
                                (1.023, 42.08) 
                                (2.511, 45.10) 
                                (5.022, 47.14) 
        };

        \addplot[color=qwenthreecol, markline] coordinates {
                                (0.119, 35.80)  
                                (0.208, 37.65)  
                                (0.386, 40.20)  
                                (0.981, 44.28) 
                                (2.407, 48.16) 
                                (4.814, 50.89) 
        };

        \addplot[color=indexcol, dashed, markline] coordinates {
                                (0.119, 35.93)  
                                (0.208, 37.87)  
                                (0.386, 40.72)  
                                (0.981, 45.20)  
                                (2.407, 49.95)  
                                (4.814, 53.44)  
        };

        \threecolornode{0.103}{33.7}{4}{qwenthreecol}{qwencol}{indexcol}{1.2}
        \threecolornode{0.23}{33.2}{7}{qwenthreecol}{qwencol}{indexcol}{1.3}
        \threecolornode{0.43}{35.1}{13}{qwenthreecol}{qwencol}{indexcol}{2.2}
        \threecolornode{0.9}{45.4}{33}{qwenthreecol}{qwencol}{indexcol}{2.2}
        \threecolornode{2.1}{50.0}{81}{qwenthreecol}{qwencol}{indexcol}{2.4}
        \threecolornode{3.8}{52.3}{162}{qwenthreecol}{qwencol}{indexcol}{3}

        \node [above] at (axis cs:  0.1, 39.6) {\footnotesize \textcolor{amescol}{100}};
        \node [above] at (axis cs:  0.2, 40.3) {\footnotesize \textcolor{amescol}{200}};
        \node [above] at (axis cs:  0.38, 41.0) {\footnotesize \textcolor{amescol}{400}};
        \node [above] at (axis cs:  1, 38.0) {\footnotesize \textcolor{amescol}{1000}};
        \node [above] at (axis cs:  2.5, 38.5) {\footnotesize \textcolor{amescol}{2500}};
        \node [above] at (axis cs:  4.7, 38.8) {\footnotesize \textcolor{amescol}{5000}};
        
    \end{axis}
\end{tikzpicture}
    \vspace{-11pt}
    \caption{\footnotesize \textbf{Performance \vs re-ranking time.} MLLM-based re-rankers have higher per-image inference time than methods trained specifically for instance-level image retrieval with re-ranking (AMES), but under a fixed query-time budget, they achieve better retrieval performance, already with as few as 20 re-ranked images, indicated by the colored numbers. All reported methods have roughly the same memory footprint.
    \vspace{-12pt}
    \label{fig:time_map}}
\end{figure}
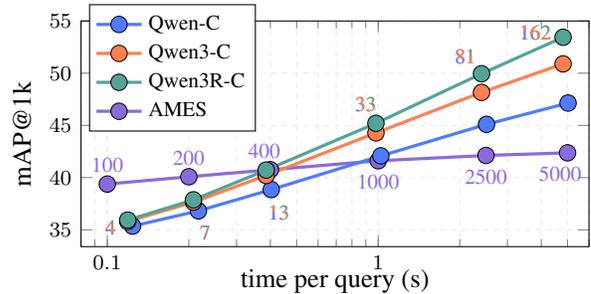

In this work, we explore MLLMs in a training-free manner for a vision-only task, \ie, large-scale image retrieval. Our focus is on instance-level retrieval~\cite{sz03,pci+07}, where the goal is to find the exact object rather than a category. Specifically, we provide the query and candidate images to the model and prompt it with a simple question, \eg ``Do the images show the same object?'', requiring a binary (yes/no) response. We convert the next-token probabilities for ``yes'' and ``no'' into a similarity score, effectively turning the model into a zero-shot classifier for object identity. 

This constitutes an unconventional application of MLLMs, as image similarity does not inherently involve language understanding. Our goal is to repurpose their reasoning capabilities, avoiding specialized architectures~\cite{agt+15,cas20}, domain-specific training data~\cite{wac+20,rtc19}, and costly fine-tuning typically used for instance-level retrieval. We leverage the extensive pre-training of MLLMs across diverse domains, which implicitly equips them with strong visual discrimination in an open-world setting. Interestingly, their internal design often involves comparing image patches, reminiscent of similarity-learning architectures~\cite{tyo+21,ski+24}, albeit without explicit supervision for this task.

We integrate this similarity estimation into a large-scale retrieval pipeline as a re-ranking step on top of an efficient initial ranking using compact image descriptors. Since MLLMs require raw, often high-resolution images, direct application to large collections is impractical. To address this, we adopt indexing-friendly design choices: database images are pre-processed into patch-level features, quantized or pruned, and stored in memory-efficient structures. During retrieval, only the top-$k$ candidates are processed by the MLLM, reducing computational overhead.

Our findings show that MLLMs, when prompted appropriately, serve as effective training-free similarity estimators for instance-level retrieval, offering not only a favorable accuracy-memory tradeoff but also a favorable accuracy-speed tradeoff when operating under a sufficiently generous time budget (see Figure~\ref{fig:time_map}). Compared to task-specific models, MLLMs perform better outside their native training domains, demonstrating superior generalization across diverse object categories.
We show that our indexing scheme is effective not only for training-free usage of generic MLLMs but also for MLLM models specifically fine-tuned as re-rankers.
\looseness=-1

Qualitatively, MLLMs outperform global descriptors and task-specific re-rankers in challenging cases involving small objects, heavy clutter, and occlusion, showcasing strong partial matching capabilities. Despite these advantages, we identify unexpected failure modes in robustness analysis, paving the way for future research to enhance the reliability and scalability of MLLMs for similarity estimation and large-scale retrieval.

\section{Related work}
\label{sec:related}

\myparagraph{Image retrieval.} 
Image retrieval has largely relied on learned global descriptors. Early deep methods repurpose classification networks by pooling activation maps~\cite{bl15, rtc19}, while later approaches train networks explicitly for retrieval using metric learning objectives~\cite{skp+15, rar+19} and advanced sampling strategies~\cite{sohn16, lxz+19}. Although effective, these methods require domain-specific training data, which limits adaptability. Foundation vision models such as CLIP~\cite{clip} introduced generic representations with strong cross-domain generalization, enabling efficient large-scale retrieval. Yet, global descriptors often lack the precision needed for fine-grained and instance-level similarity estimation.

To improve precision, recent work has incorporated re-ranking based on local descriptors. State-of-the-art methods such as RRT~\cite{tyo+21}, $R^2$Former~\cite{zyc+23}, and CVNet~\cite{lsl+22} refine similarity estimation by matching dense features or modeling cross-image correlations with transformer or convolutional architectures. Since such dense approaches often incur high indexing costs, AMES~\cite{ski+24} is designed for an asymmetric setup with binarized descriptors~\cite{ktp+22}, while dense-to-sparse CVNet~\cite{lee2024correlation} sparsifies database descriptors. LOCORE~\cite{xiao2025locore} uses only a fraction of local descriptors and departs from pairwise similarity by fine-tuning an LLM for listwise re-ranking. Despite their strong performance, these methods remain computationally and memory-intensive and rely on domain-specific training. In contrast, we propose a training-free re-ranking approach based on MLLMs, leveraging their open-world generalization while avoiding specialized architectures and domain-dependent training.

\myparagraph{MLLM for image retrieval.}
MLLMs~\cite{llama3, bai2025qwen2} offer reasoning capabilities beyond traditional embedding approaches. In retrieval, they have been studied mainly in settings where text is part of the query, such as composed image retrieval (CIR). Although some methods address CIR in a zero-shot manner~\cite{freedom,tzj+25,hls+25}, instance-level granularity is argued to require dedicated training~\cite{lbl+25,collm} or additional corpora with positive and negative samples~\cite{icir}. 

Another line of work trains MLLMs to produce universal multimodal embeddings, including for image-to-image retrieval. These approaches rely on extensive fine-tuning, typically through contrastive objectives on large-scale paired data~\cite{e5} or massive multimodal collections~\cite{mmembed, vlm2vec,gme,mbeir}. This trend is motivated by the view that pre-trained MLLMs are weak at retrieval. For example, LamRA~\cite{lamra} trains a LoRA~\cite{hu2022lora} module for retrieval and re-ranking, reporting poor performance when MLLMs are used directly, while MM-Embed~\cite{mmembed} introduces modality-aware hard-negative fine-tuning, yet still relies on the pre-trained MLLM for re-ranking. To the best of our knowledge, we are the first to show that a pre-trained MLLM can serve directly as a strong zero-shot re-ranker for pure image-to-image, instance-level retrieval in a fully training-free setting.

\begin{figure*}[t]
    \vspace{-3pt}
    \begin{center}
    \includegraphics[width=.96\linewidth]{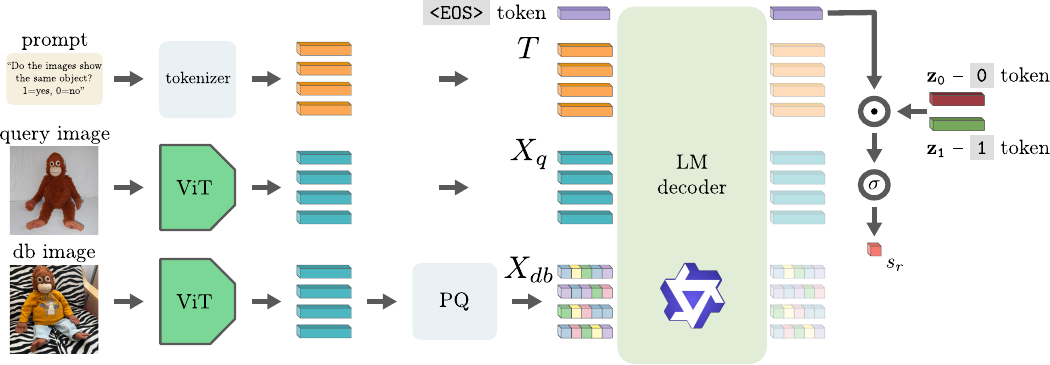}
    \end{center}
    \vspace{-15pt}
    \caption{\textbf{Overview of the proposed MLLM-based re-ranking approach} for instance-level image retrieval. 
    A query image, a database image, and a task-specific textual prompt are jointly used as input for similarity estimation. 
    A vision encoder extracts visual tokens from both images ($X_q$, $X_{db}$), which are concatenated with the textual prompt tokens ($T$) and the end-of-sentence (EOS) token to form the multimodal input to the MLLM. 
    Scalable retrieval is achieved by applying compression and indexing schemes, \eg Product Quantization (PQ), to the database image tokens, enabling memory-efficient storage. 
    The MLLM outputs the updated EOS token representation, which is the joint representation of all inputs and is then compared via dot product with the classifier vectors $\vz_{\texttt{1}}$ and $\vz_{\texttt{0}}$, corresponding to vocabulary tokens ``\texttt{1}'' and ``\texttt{0}'', respectively. 
    The final similarity score ($s_r$) is given by the softmax ($\sigma$) probability assigned to token ``\texttt{1}''.
    \vspace{-12pt}
    \label{fig:overview}}
\end{figure*}

\myparagraph{Improved MLLM efficiency.}
While recent MLLMs can process high-resolution images for fine-grained retrieval~\cite{qwen2-vl,wang2025internvl3_5}, indexing large-scale databases with them remains memory-intensive. FastV~\cite{fastv} attributes much of this cost to redundant visual tokens processed within transformer layers. To address this, prior work has explored visual token pruning within the transformer by identifying and removing less informative tokens, for example, by maximizing token diversity in DivPrune~\cite{divprune} or exploiting attention sparsity in LLaVA-PruMerge~\cite{llava-prumerge}.

Other methods compress visual content into a smaller token set by modifying the feature extractor. TokenPacker~\cite{tokenpacker} merges low-resolution queries with high-resolution details while BRAVE~\cite{brave} uses a Multi-Encoder Querying Transformer to fuse and resample features from multiple vision encoders.
In contrast, our work mainly focuses on training-free token-set compression to reduce the memory footprint.
\section{Method}
\label{sec:method}

\subsection{Problem Formulation}

The goal of image retrieval is to search a database of images given a query image $q$ and return the most relevant results. We focus on the most fine-grained notion of relevance, \ie instance-level retrieval, where two images are considered relevant if they depict the same specific object. Formally, retrieval requires a similarity function $s(q, x) \in \mathbb{R}$ that measures the similarity between the query and a database image $x$, enabling ranking of all candidates by similarity.

In large-scale settings with millions of images, efficiency in both speed and memory is critical. We adopt a two-stage pipeline:  
(i) an initial exhaustive search using compact global descriptors extracted by a vision encoder, producing a coarse ranking (global similarity $s_g(q, x)$)
(ii) a re-ranking step applied to the top-$k$ candidates using a more accurate but computationally intensive similarity estimation (re-ranking similarity $s_r(q, x)$).

Our re-ranking leverages an MLLM to estimate instance-level similarity between image pairs. This approach improves precision while controlling cost through $k$. However, re-ranking typically increases memory requirements, as it demands richer representations than a single global descriptor. Direct access to raw database images, often assumed by MLLMs, is impractical due to storage overhead and prohibitive query-time latency. To address this, we design an indexing-friendly solution that enables memory-efficient storage of image representations while exploiting the reasoning capabilities of MLLMs for re-ranking.

\subsection{MLLM architecture overview}
Given a set of images $\{x_1, \dots, x_C\}$ and a text prompt $p$, the MLLM extracts visual tokens $V$ and textual tokens $T$, and estimates the probability distribution over an output token sequence $y_{1:N}$ from a vocabulary $Z$ as:
\begin{equation}
P(y_{1:N} \mid V, T) = \prod_{n=1}^N P\big(y_n \mid y_{1:n-1}, V, T \big),
\label{eq:mllm}
\end{equation}
where $y_n$ denotes the $n$-th generated token, and $P(\cdot)$ is a categorical distribution over $Z$.

Visual tokens are obtained via a vision encoder $g: \mathbb{R}^{H \times W \times 3} \rightarrow \mathbb{R}^{M \times D}$, which maps an input image of height $H$ and width $W$ into a sequence of $M$ $D$-dimensional embeddings. Each image $x_i$ is mapped to a feature matrix $X_i = g(x_i)$. We employ a Vision Transformer (ViT)~\cite{dbk+21} for this purpose, which divides each image into non-overlapping patches and processes them through a transformer network~\cite{vsp+17}. The visual token sequence is formed by concatenating the encoded representations of all input images, \ie $V = [\, X_1, \dots, X_c \,] \in \mathbb{R}^{L_v \times D}$, where $L_v = C \cdot M$ denotes the total number of visual tokens.\footnote{For simplicity, we assume that all images produce the same number of tokens, which may not hold in practice due to varying image sizes.}

The textual prompt $p$ is tokenized, and the corresponding token embeddings are retrieved from a lookup table, forming $T \in \mathbb{R}^{L_t \times D}$, where $L_t$ is the number of textual prompt tokens. Generated tokens $y_n$ are also mapped into the same embedding space as the prompt tokens.

We denote the language model by $f : \mathbb{R}^{(n + L_v + L_t) \times D} \rightarrow \mathbb{R}^{D}$, which processes the concatenated multimodal token sequence, comprising visual tokens, textual prompt tokens, previously generated tokens, and a special \emph{end-of-sentence} (EOS) token, and maps them to a sequence of vector representations. The model is implemented as a transformer decoder, outputting one vector per input token. All outputs except the one corresponding to the EOS token are discarded; the EOS token's representation serves as the joint embedding of the entire input. Therefore, we represent $f$ as a mapping to a single vector.

This representation is then passed through a linear softmax classifier with $|Z|$ class vectors, one per vocabulary token. Let $\mathbf{z}_i \in \mathbb{R}^D$ denote the class vector for the $i$-th token. The probability of generating token $i$ at position $n$ is given by:
\begin{equation}
P_i(t_n \mid t_{1:n-1}, V, T) = 
\frac{
    e^{\vz_i\tran  f([\,V, T, t_{1:n-1}\,])}
}{
    \sum\limits_{j \in 1\ldots |Z|}
    e^{\vz_j\tran  f([\,V, T, t_{1:n-1}\,])}
}
\label{eq:prob}
\end{equation}

\subsection{Image-to-image similarity with MLLMs.} 
We repurpose the MLLM to estimate the similarity between a query image $q$ and a database image $x_{\text{db}}$. The model is prompted with a textual instruction $p$ that asks whether the two images depict the same object instance:

\noindent
\begingroup
\setlength{\fboxsep}{8pt}%
\ttfamily\footnotesize 
\colorbox{gray!15}{%
\parbox{0.9\linewidth}{%
You are given two images: a query and a candidate. Determine whether the exact same object instance from the query image is present in the candidate image. 
Output strictly a single digit, do not output anything else.: 
\begin{itemize}
    \item 0 = the object instance does not appear. 
    \item 1 = the object instance appears in the candidate. 
\end{itemize}
}}
\endgroup

The model is expected to generate the token ``\texttt{1}'' if the images show the same object, and ``\texttt{0}'' otherwise. Thus, the vocabulary of interest is restricted to these two tokens only, represented by vectors $\vz_{\texttt{0}}$ and $\vz_{\texttt{1}}$.
In this setting, the model does not perform auto-regressive generation, and no multiple output tokens are considered.
Consequently, the MLLM operates as a binary classifier that predicts if the two images depict the same object instance.

The input to the function $f$ consists of the visual tokens extracted from the query and database images, along with the textual prompt tokens. Specifically, the input sequence is $ u=  [\, X_q,\, X_{\text{db}},\, T \,]$, where $X_q = g(x_q)$ and $X_{\text{db}} = g(x_{\text{db}})$ are the visual token sequences for the query and database images, respectively, and $T$ represents the tokenized prompt. The joint representation $\vu = f(u)$ is obtained by processing this input through the MLLM.

To estimate similarity, we evaluate the probability assigned to token  ``\texttt{1}'' using the softmax classifier 
\begin{equation}
s_r(q, x_{\text{db}}) = 
\frac{
    e^{\vz_\texttt{1}\tran \vu}
}{
    e^{\vz_\texttt{0}\tran \vu} + 
    e^{\vz_\texttt{1}\tran \vu}
}.
\label{eq:softmax}
\end{equation}
This probability serves as the re-ranking similarity score between the query image $q$ and the database image $x_{\text{db}}$.

\subsection{Compression and indexing.} 
During re-ranking, the representations of the query image and the textual prompt, denoted as $X_q$ and $T$, are computed once per query and reused across all $k$ candidate images. In contrast, the representations of the top-$k$ database images, $X_{\text{db}} \in \mathbb{R}^{M \times D}$, must be accessed individually. To enable scalable retrieval, we explore strategies for indexing that involve compressing and storing these representations in advance, in a memory-efficient manner. 

Let $\mathbf{x}_m \in \mathbb{R}^{D}$, for $m = 1, \ldots, M$, denote the representation of the $m$-th patch (token) extracted from a database image $x_{\text{db}}$. Our goal is to reduce both the memory footprint of each token and the total number of tokens per image. We investigate the following indexing strategies:

    \textbf{Product Quantization (PQ).} 
    Instead of storing full-precision token embeddings $\mathbf{x}_m \in \mathbb{R}^D$, we apply Product Quantization (PQ)~\cite{jed+11} to obtain compact codes. The $D$-dimensional space is divided into $d$-dimensional subspaces, each associated with a codebook of $K$ centroids, computed offline on a generic image set. Each vector is represented by indices of the nearest centroids in each subspace, resulting in $\nicefrac{D}{d}$ indices per vector. We adopt a standard configuration with $K=256$ centroids per subspace, allowing each index to be stored in one byte, and each vector in $\nicefrac{D}{d}$ bytes. Prior to feeding into the model, compressed tokens are reconstructed by concatenating the corresponding centroids from each subspace, yielding an approximate full-precision vector. PQ enables a trade-off between memory usage and reconstruction fidelity by adjusting $d$.

    \textbf{Token pruning.}
    We adapt DivPrune~\cite{divprune}, a recent token pruning method designed for MLLMs. The algorithm implements a diverse selection, similar to Farthest Point Sampling (FPS). It begins by selecting the token furthest from its closest neighbor, then iteratively adds the token that is furthest from the set of tokens already selected.

    \textbf{$k$-means clustering for token selection.}  
    We apply $k$-means clustering to the tokens of each image, grouping them into $K$ clusters. The centroids are the stored representations, while to define the spatial position of the centroid tokens when fed to the network, we pick the position of the medoids, \ie tokens closest to the cluster centroids. 

    \textbf{Uniform token sampling.}  
    To reduce the token count, we uniformly sample tokens from the original set. Specifically, we retain one token per non-overlapping $2 \times 2$ window on the 2D token grid, effectively downsampling by a factor of 4. The rest tokens are discarded. Selected tokens are preserved in their original form and accompanied by the original spatial positions to retain spatial context.

    \textbf{Average pooling with a sliding window.}  
    Instead of selecting a single token, we compute the average of the four tokens within each non-overlapping $2 \times 2$ window. This reduces the number of tokens by a factor of 4 while preserving local information in a compact form.
    
    \textbf{Resolution reduction as token reduction.}  
    The number of tokens $M$ is proportional to the input image resolution. For a ViT with patch size $P$, the number of tokens scales as $M \sim \frac{H}{P} \times \frac{W}{P}$. Particularly, the Qwen model we use applies an additional $2 \times 2$ pooling on the 2D token grid, reducing the number of tokens by a factor of 4. Thus, reducing the input resolution directly decreases the number of tokens.
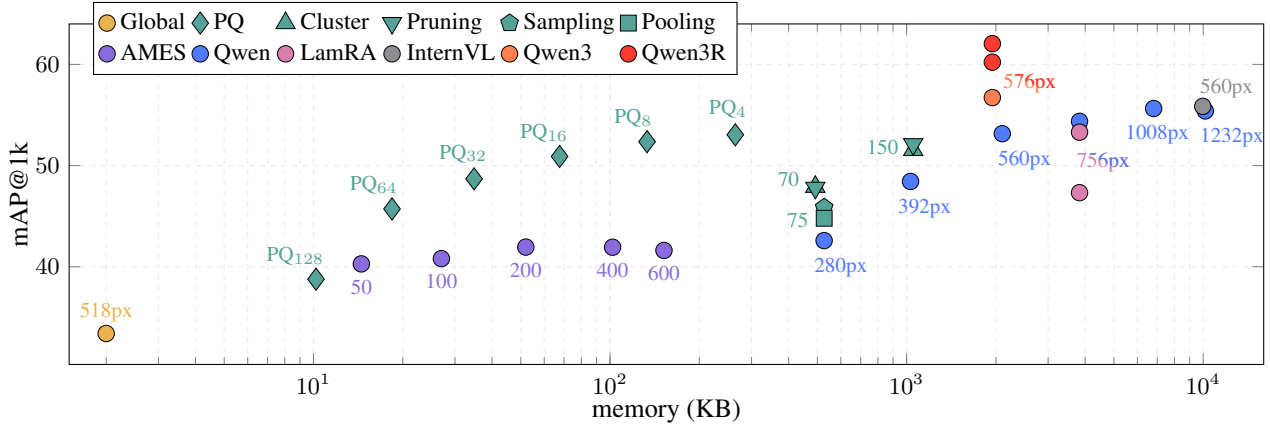
\begin{figure*}[t]
    \centering
    \pgfplotsset{every tick label/.append style={font=\small}}
\begin{tikzpicture}
\begin{axis}[%
  width=1\linewidth,
  height=0.35\linewidth,
  ylabel={mAP@1k},
  xlabel={memory (KB)},
  grid=both,
  grid style={color=lightgray!60, dash pattern=on 2pt off 2pt},
  xlabel style={yshift=5pt},
  ylabel style={yshift=-4pt},
  xmode=log,
  ymax=64,
  xmin=1.5,
  xmax=16000,
  legend columns=6, 
  legend style={
    anchor=south west, 
    at={(0.02,0.85)}, 
    cells={anchor=west}, 
    font =\small, 
    fill opacity=1, 
    inner sep=2pt
    },
  ]
  
        \addlegendimage{color=pecol, normmark}
        \addlegendentry{Global}
        
        \addlegendimage{pqmark, color=indexcol}
        \addlegendentry{PQ}
        \addlegendimage{clustmark, color=indexcol}
        \addlegendentry{Cluster}
        \addlegendimage{prunmark, color=indexcol}
        \addlegendentry{Pruning}
        \addlegendimage{samplemark, color=indexcol}
        \addlegendentry{Sampling}
        \addlegendimage{poolmark, color=indexcol}
        \addlegendentry{Pooling}
        
        \addlegendimage{color=amescol, normmark} 
        \addlegendentry{AMES}
        \addlegendimage{color=qwencol, normmark} 
        \addlegendentry{Qwen}
        \addlegendimage{color=lamracol, normmark} 
        \addlegendentry{LamRA}
        \addlegendimage{color=intercol, normmark} 
        \addlegendentry{InternVL}
        \addlegendimage{color=qwenthreecol, normmark} 
        \addlegendentry{Qwen3}
        \addlegendimage{color=qwenrerankcol, normmark} 
        \addlegendentry{Qwen3R}

        \addplot[color=pecol, normmark] coordinates {
                                (2, 33.40) 
        };

        \addplot[color=amescol, normmark] coordinates {
                                (152.0, 41.61) 
                                (102.0, 41.93) 
                                (52.0,  41.94) 
                                (27.0,  40.80) 
                                (14.5,  40.28) 
        };

        \addplot[color=qwencol, normmark] coordinates {
                                (10166.0, 55.40) 
                                (6806.0, 55.64) 
                                (3829.3, 54.37) 
                                (2102.0, 53.15) 
                                (1031.0, 48.43) 
                                (527.0,  42.58) 
        };

        \addplot[color=lamracol, normmark] coordinates {
                                (3829.3, 47.3204) 
                                (3829.3, 53.3) 
        };

        \addplot[color=intercol, normmark] coordinates {
                                (9968, 55.85) 
        };

        \addplot[color=qwenthreecol, normmark] coordinates {
                                (1946.0, 56.72) 
        };

        \addplot[color=qwenrerankcol, normmark] coordinates {
                                (1946.0, 60.22) 
                                (1946.0, 62.05) 
        };
        
        \addplot[color=indexcol, pqmark] coordinates {
                                (264.5,  53.04) 
                                (133.3,	 52.36) 
                                (67.6,   50.90) 
                                (34.8,   48.69) 
                                (18.4,   45.71) 
                                (10.2,   38.77) 
        };

        \addplot[color=indexcol, clustmark] coordinates {
                                (1052.3, 51.54) 
                                (492.3, 47.90) 
        };

        \addplot[color=indexcol, prunmark] coordinates {
                                (492.3, 47.80) 
                                (1052.3, 52.11) 
        };

        \addplot[color=indexcol, samplemark] coordinates {
                                (527.3, 45.86) 
        };

        \addplot[color=indexcol, poolmark] coordinates {
                                (527.3, 44.79) 
        };

        \node [above] at (axis cs:  250.5,53.8) {\footnotesize \textcolor{indexcol}{PQ$_{4}$}};
        \node [above] at (axis cs:  120.,52.9) {\footnotesize \textcolor{indexcol}{PQ$_{8}$}};
        \node [above] at (axis cs:  60.0,51.4) {\footnotesize \textcolor{indexcol}{PQ$_{16}$}};
        \node [above] at (axis cs:  32.0, 49.4) {\footnotesize \textcolor{indexcol}{PQ$_{32}$}};
        \node [above] at (axis cs:  16,46.3) {\footnotesize \textcolor{indexcol}{PQ$_{64}$}};
        \node [above] at (axis cs:  8.7,39.3) {\footnotesize \textcolor{indexcol}{PQ$_{128}$}};
        
        \node [above] at (axis cs:  830., 50.2) {\footnotesize \textcolor{indexcol}{150}};
        \node [above] at (axis cs:  430., 43.) {\footnotesize \textcolor{indexcol}{75}};
        \node [above] at (axis cs:  400., 47.) {\footnotesize \textcolor{indexcol}{70}};
        
        \node [above] at (axis cs:  600., 38.1) {\footnotesize \textcolor{qwencol}{280px}};
        \node [above] at (axis cs:  1150., 43.9) {\footnotesize \textcolor{qwencol}{392px}};
        \node [above] at (axis cs:  2500., 48.7) {\footnotesize \textcolor{qwencol}{560px}};
        \node [above] at (axis cs:  4600., 48.5) (T) {\footnotesize\textcolor{lamracol}{756px}};
        \begin{scope}    
            \clip (T.north west) -- (T.south east) -- (T.north east) -- cycle;
            \node at (T) {\footnotesize\color{qwencol} 756px};
        \end{scope}
        \node [above] at (axis cs:  7000., 51.3) {\footnotesize \textcolor{qwencol}{1008px}};
        \node [above] at (axis cs:  12500., 51.0) {\footnotesize \textcolor{qwencol}{1232px}};
        
        \node [above] at (axis cs:  12000., 55.8) {\footnotesize \textcolor{intercol}{560px}};
        \node [above] at (axis cs:  2., 33.7) {\footnotesize \textcolor{pecol}{518px}};
        
        \node [above] at (axis cs:  2600., 56.3) (T) {\footnotesize\textcolor{qwenthreecol}{576px}};
        \begin{scope}    
            \clip (T.north west) -- (T.south east) -- (T.north east) -- cycle;
            \node at (T) {\footnotesize\color{qwenrerankcol} 576px};
        \end{scope}
        
        \node [below] at (axis cs:  152., 41.01) {\footnotesize \textcolor{amescol}{600}};
        \node [below] at (axis cs:  102., 41.33) {\footnotesize \textcolor{amescol}{400}};
        \node [below] at (axis cs:  52., 41.34) {\footnotesize \textcolor{amescol}{200}};
        \node [below] at (axis cs:  27., 40.2) {\footnotesize \textcolor{amescol}{100}};
        \node [below] at (axis cs:  14.5, 39.68) {\footnotesize \textcolor{amescol}{50}};

    \end{axis}
\end{tikzpicture}
    \vspace{-8pt}
    \caption{\textbf{Performance \vs memory.} Memory footprint per image is approximated assuming a fixed resolution, where the longer image side is set to the target value and the shorter side is scaled to 3/4 of it, reflecting a typical aspect ratio.
    Five models for re-ranking, operating on top of {\color{pecol} global similarities} obtained from the PE model with linear adaptation.
    All {\color{indexcol} indexing strategies} are applied on the {\color{qwencol} Qwen} model with 560px image resolution. Text indicates: subspace dimension for PQ (\eg~PQ$_{16}$ has 16 subspace dimensions), pixels for different resolutions (\eg 560px), and number of tokens (\eg 70) for the rest of the indexing methods. {\color{amescol} AMES} used with descriptor sets of increasing cardinalities.
    {\color{lamracol} LamRA} and {\color{qwenrerankcol} Qwen3R} are shown both with their default prompts and also with our prompt, which performs better.
    \vspace{-5pt}
    \label{fig:map_vs_mem}}
\end{figure*}

\section{Experiments}
\label{sec:experiments}

\subsection{Evaluation setting}

\myparagraph{Datasets and metric.} We evaluate our method on several established datasets for instance-level image retrieval. Our main evaluation is conducted on \textit{\textbf{ILIAS}}~\cite{ilias2025}, a large-scale, multi-domain dataset designed for instance-level image retrieval. Due to its size, diversity, and challenging nature, ILIAS naturally reflects open-world settings. In addition, we evaluate on three other instance-level datasets, \ie \textit{\textbf{INSTRE}}~\cite{wj15}, \textit{\textbf{\rop}}~\cite{rit+18}, and \textit{\textbf{Product1M}}~\cite{zwd+21}.
Further details regarding the datasets are in the supplementary material. Performance is measured using mean Average Precision at rank $k$ (mAP@$k$), which computes the mean of the average precision across all queries, truncated at the top $k$ retrieved results. Unless specified otherwise, we evaluate mAP@$1k$.\vspace{15pt}

\myparagraph{Implementation details.}
As the default choice for the MLLM, we use the Qwen model~\cite{bai2025qwen2}, specifically the \texttt{Qwen2.5-VL-7B-Instruct} variant. In this configuration, the vision encoder is a 32-layer ViT with a patch size of 14. The Qwen2.5 family incorporates a token merger module that merges every four tokens before propagation to the LLM, resulting in an effective patch size of 28. An input image of resolution $560 \times 420$ produces $M = 300$ visual tokens, each of dimensionality $D = 3{,}584$. 
Hence, in the uncompressed setting, these tokens are stored in \texttt{float16} format, amounting to approximately 2MB of memory per image. We evaluate the indexing strategies using the following configurations: PQ with subspace dimensions $d\in\{4, 8, 16, 32, 64, 128\}$ learned on a set on 2M vectors extracted from LAION~\cite{schuhmann2022laion} images, input image resolutions $\{280, 392, 560, 756, 1008, 1232\}$, $k$-means clustering with $K \in \{150, 70\}$ clusters, and token pruning resulting in final token sets of $\{150, 70\}$ tokens. Following prior work~\cite{cas20,lsl+22,ski+24}, we use global-local similarity ensembling with fixed factor $\lambda=0.5$ for all datasets. We re-rank the top-1k neighbors based on global similarity. 
Unless stated otherwise, we use 756px for the no-compression setting and PQ$_{16}$ with 560px for the compression setting as the default configurations. The use of compression is denoted by C, \eg Qwen-C.\vspace{15pt}

\newblock

\myparagraph{Other methods.}
As a global descriptor we use the best-performing ones on ILIAS\footnote{\url{https://vrg.fel.cvut.cz/ilias/}}, \ie Perception Encoder (PE) with linear adaptation~\cite{ilias2025} and DINOv3~\cite{simeoni2025dinov3}. 
Additionally, we evaluate and compare with AMES~\cite{ski+24}, a state-of-the-art re-ranking method for instance-level image retrieval, which is trained on GLDv2~\cite{wac+20}, an instance-level dataset of landmarks. 
For a fair comparison, we train AMES with more powerful contemporary local descriptors extracted with DINOv3, using the public implementation provided by the authors.
We use the $fp$ variant that stores full-precision descriptors. AMES is evaluated with sets of $\{50, 100, 200, 400, 600\}$ local descriptors. 
We also benchmark the more recent MLLMs, InternVL3.5~\cite{wang2025internvl3_5} and Qwen3~\cite{yang2025qwen3}, used in a training-free way.
As MLLM-based models that are trained for re-ranking, we evaluate LamRA~\cite{lamra} (same variant as Qwen, provided by the authors) and Qwen3-Reranker (Qwen3R)~\cite{zhang2025qwen3}. 
Note that Qwen3R is not training-free as a re-ranking method, but its use remains training-free with respect to the instance-level nature of our retrieval task.
In the supplementary material, we apply the proposed indexing to Qwen3 and Qwen3R to demonstrate its universal applicability and effectiveness.

\subsection{Results}

\myparagraph{Performance \vs memory comparison.}
In \cref{fig:map_vs_mem}, we compare performance against memory footprint across different methods. 
Various re-ranking models improve retrieval performance on top of global descriptor baselines, at the expense of extra memory. 
Uncompressed Qwen variants across multiple input resolutions provide consistent gains, with the 1008px variant achieving over 22~points higher mAP compared to global, but with three orders of magnitude higher memory footprint.  The performance saturates for larger resolutions.
Although InternVL achieves slightly higher performance even at a smaller resolution of 560px, its memory cost remains significantly larger (by almost 50\%), due to the image tiling mechanism used for token extraction. 
Notably, with a more recent model, \ie Qwen3, performance is higher with a lower memory footprint, highlighting the value of training-free methods benefiting from MLLM advances in a straightforward way.
AMES, which is explicitly trained for this task on landmark images, yields smaller improvements but with a much lower memory footprint. 
Interestingly, the MLLM architecture, specifically the function $f$, closely resembles that of AMES, and is identical to RRT~\cite{tyo+21}, a re-ranking model preceding AMES. 
These models are designed to cross-compare sparse sets of local descriptors within and across images, analogous to how the MLLM performs cross-image reasoning, but with a dense and rich set of local descriptors. 
Still, Qwen benefits from its broad generalization capabilities, having been exposed to diverse visual domains, beyond landmarks, during training.

LamRA model performs comparably to Qwen when used with our prompt, indicating that general-purpose fine-tuning is not necessarily beneficial for instance-level retrieval. Nevertheless, results of Qwen3R show that its specialized architecture and fine-tuning can yield further improvements compared to the training-free variants.

Using indexing strategies on the representation of Qwen, \ie PQ in particular, allows Qwen to provide a much better trade-off than AMES. 
None of the token selection methods, namely pooling, sampling, clustering, or pruning, stands out in terms of a good trade-off. 
Interestingly, DivPrune, a previously proposed method for a similar task, \ie token pruning, is not better than our simple clustering-based baseline.

ELViS~\cite{suma2026elvis} is a concurrent work to ours, which achieves 52.0 mAP@1k, evaluated with 400 local descriptors from DINOv3, same as AMES. Its performance is much higher than AMES, and comparable to Qwen with PQ$_8$ for roughly the same memory. 
Its drawback is that it requires supervised training, but its advantage is speed, by re-ranking 1000 images in 0.1 seconds.

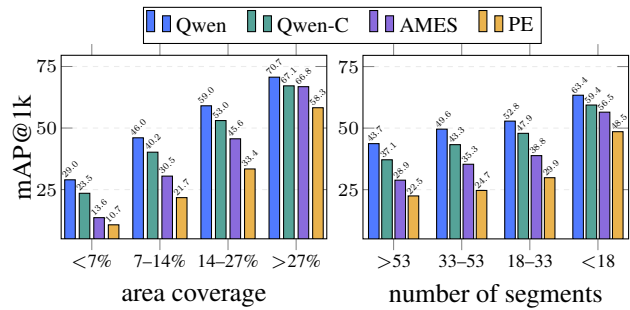
\begin{figure}[t]
    \vfill
    \centering
    \begin{tabular}{cc}
        \hspace{-10pt}
        \begin{subfigure}{0.48\textwidth}
            \begin{tikzpicture}
\begin{axis}[
    width=0.61\linewidth,
    height=0.48\linewidth,
    bar width=4pt, 
    ylabel={mAP@1k},
    xlabel={area coverage},
    ymin=5, ymax=76,
    ytick={25,50,75}, 
    ytick distance=5, 
    ybar=1.5,
    enlarge y limits={upper, value=0.05},
    xtick=data,
    symbolic x coords={  
        $<$7\%,
        7--14\%,
        14--27\%,
        $>$27\%,
    },
    ylabel style={yshift=-5pt},
    xlabel style={yshift=-1pt},
    xtick distance = 1,
    xticklabel style={
        font=\scriptsize,
        yshift=2pt,
        },
    yticklabel style={
        font=\scriptsize,
        xshift=2pt,
        },
    enlarge x limits=0.15, 
    nodes near coords,
    every node near coord/.append style={
        scale=.5, 
        anchor=south west, 
        xshift=-3.5pt,
        yshift=-3.5pt,
        color=black,
        rotate=45,
        font=\scriptsize, 
        /pgf/number format/.cd, 
        fixed zerofill, 
        precision=1,
    },
    legend style={
        at={(0.3,1.05)}, 
        anchor=south west, 
        legend columns=-1,  
        /tikz/every even column/.append style={column sep=0.2cm}, 
        font=\footnotesize,
        inner sep=1pt,  
        outer sep=1pt 
    },
    legend cell align={left}, 
    ymajorgrids=true,
    grid style={color=lightgray!60, dash pattern=on 2pt off 2pt},
]

\pgfplotstableread[col sep=comma]{data/performance_per_scale.csv}\loadeddata

\addplot+[ybar, color=black, fill=qwencol] table[x=domain,y=qwen] {\loadeddata};
\addplot+[ybar, color=black, fill=indexcol] table[x=domain,y=qwenc] {\loadeddata};
\addplot+[ybar, color=black, fill=amescol] table[x=domain,y=ames] {\loadeddata};
\addplot+[ybar, color=black, fill=pecol] table[x=domain,y=pe] {\loadeddata};

\legend{Qwen, Qwen-C,  AMES, PE}
    \end{axis}

\end{tikzpicture}
        \end{subfigure}
        &
        \hspace{-130pt}
        \begin{subfigure}{0.48\textwidth}
            \begin{tikzpicture}
\begin{axis}[
    width=0.61\linewidth,
    height=0.48\linewidth,
    bar width=4pt, 
    xlabel={number of segments},
    ymin=5, ymax=76,
    ytick={25,50,75}, 
    ytick distance=5, 
    ybar=1,
    enlarge y limits={upper, value=0.05},
    xtick=data,
    symbolic x coords={  
        $>$53,
        33--53,
        18--33,
        $<$18,
    },
    ylabel style={yshift=-5pt},
    xlabel style={yshift=1pt},
    xtick distance = 1,
    xticklabel style={
        font=\scriptsize,
        yshift=2pt,
        },
    yticklabel style={
        font=\scriptsize,
        xshift=2pt,
        },
    enlarge x limits=0.15, 
    nodes near coords,
    every node near coord/.append style={
        scale=.5, 
        anchor=south west, 
        xshift=-2.5pt,
        yshift=-4pt,
        color=black,
        rotate=45,
        font=\scriptsize, 
        /pgf/number format/.cd, 
        fixed zerofill, 
        precision=1,
    },
    legend style={
        at={(0.0,1.2)}, 
        anchor=north west, 
        legend columns=-1,  
        /tikz/every even column/.append style={column sep=0.2cm}, 
        font=\footnotesize,
        inner sep=1pt,  
        outer sep=1pt 
    },
    legend cell align={left}, 
    ymajorgrids=true,
    grid style={color=lightgray!60, dash pattern=on 2pt off 2pt},
]

\pgfplotstableread[col sep=comma]{data/performance_per_clutter.csv}\loadeddata

\addplot+[ybar, color=black, fill=qwencol] table[x=domain,y=qwen] {\loadeddata};
\addplot+[ybar, color=black, fill=indexcol] table[x=domain,y=qwenc] {\loadeddata};
\addplot+[ybar, color=black, fill=amescol] table[x=domain,y=ames] {\loadeddata};
\addplot+[ybar, color=black, fill=pecol] table[x=domain,y=pe] {\loadeddata};

\end{axis}

\end{tikzpicture}
            \vspace{-0.3pt}
        \end{subfigure}
    \end{tabular}
    \vspace{-10pt}
    \caption{\textbf{Performance comparison across different amounts of object area coverage and background clutter.} Positives across all queries are jointly ranked based on coverage or clutter and split into 4 equal-sized groups. mAP@1k averaged over the queries in each group. Comparison between Qwen with and without compression (Qwen-C), AMES, and PE.
    \vspace{-5pt}
    \label{fig:scale_clutter_groups}}
\end{figure}

\begin{figure*}[t]
    \vfill
    \centering
    \vspace{-4pt}
    \input{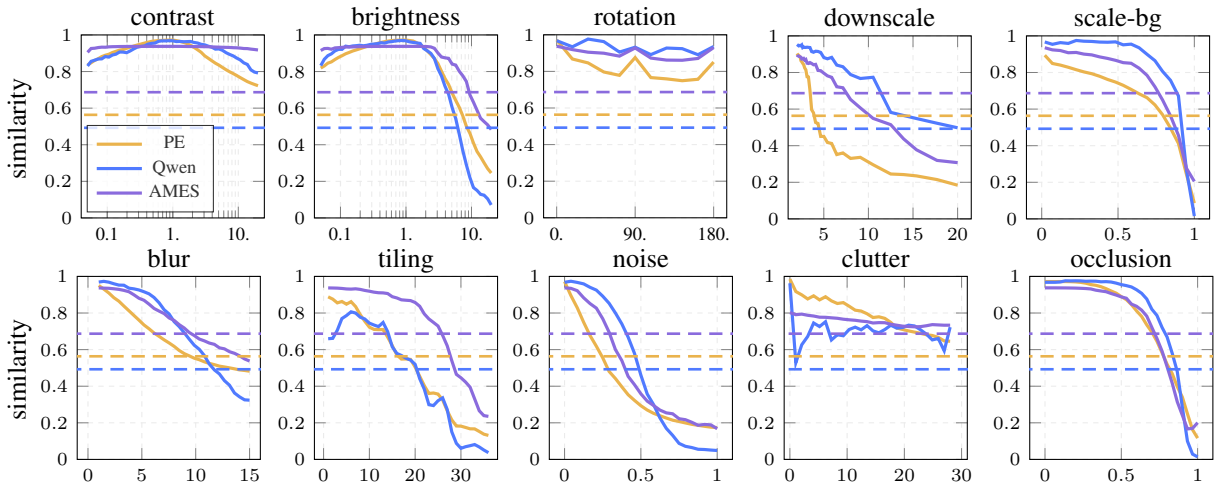}
    \vspace{-14pt}
    \caption{\textbf{Robustness analysis of MLLMs under controlled transformations.} Similarity scores for positive image pairs are shown across ten types of visual perturbations: contrast, brightness, rotation, scale, background scaling, blur, tiling, noise, clutter, and occlusion. 50 queries from ILIAS are used to generate positive pairs. Dashed lines indicate the average hard-negative similarity of the 5th percentile of the negatives in the re-ranking list, acting as a baseline. The crossing between the solid and dashed lines indicates the robustness point. Comparison between Qwen without compression, AMES, and PE. The horizontal axis represents the strength of each transformation.
    \vspace{-10pt}
    \label{fig:robustness}}
\end{figure*}

\myparagraph{Inference latency.}
Our method is designed as a re-ranking approach that explicitly controls execution time via the shortlist size. While lighter re-rankers, such as AMES, achieve lower latency, their performance saturates quickly as top-$k$ increases. \Cref{fig:time_map} reports performance as a function of re-ranking time.
For short re-ranking time, AMES provides a better trade-off by re-ranking more database images; however, our MLLM-based approach scales substantially better as time increases, especially with the more recent models, yielding significantly larger performance gains.
For fairness, AMES runtime is measured using a batch size that matches the GPU VRAM usage of Qwen-C for a single pair ($\approx$16 GB).

\myparagraph{Robustness to scale and clutter on ILIAS.} 
Following the original work~\cite{ilias2025}, we group query–positive pairs based on object area coverage and the degree of background clutter, and report average performance over queries that contain at least one positive example within each group. 
As shown in \cref{fig:scale_clutter_groups}, the MLLM-based re-ranking improves consistently across all cases. The improvement margin widens notably in challenging conditions -- specifically, when objects occupy a small fraction of the image or when scenes exhibit substantial clutter. This trend holds for both the uncompressed and compressed Qwen variants, where compression compromises the performance in a similar way across the whole range. 
While AMES, which relies on local descriptors, also consistently improves over global, the MLLM demonstrates superior robustness, particularly for small objects and heavily cluttered scenes.

\myparagraph{Robustness analysis.}
To better understand the behavior of the MLLMs, we perform a robustness analysis by artificially generating positive image pairs through controlled transformations. 
Each transformation has a continuous \emph{factor} that determines its strength -- the larger the factor, the stronger the applied distortion. 
Details of the transformations and their factor ranges are provided in the supplementary material. 
Additionally, to contextualize the absolute scores and enable model-specific relative comparisons, we establish a hard-negative baseline for each model.
For each query, we compute its similarity scores against all negative database images and extract the 5th percentile. The average of these 5th-percentile values across all queries serves as the indicative negative baseline for each model.
\cref{fig:robustness} illustrates the estimated similarity scores across ten types of transformations for each model, alongside each model's corresponding negative baseline, visualized as dashed lines.
Robustness is evaluated based on the factor value at which the model’s similarity for positive pairs drops below the negative baseline. 
Qwen exhibits strong robustness in several challenging cases, including scale variation, noise addition, Gaussian blur, and occlusion, consistently outperforming the other methods.
The only exceptions are \emph{brightness} and \emph{tiling} transformation, which simulate illumination changes and occlusion combined with background clutter, respectively, and where AMES demonstrates significantly better robustness. This indicates that drastic appearance changes pose a challenge for MLLMs, exposing a limitation that paves the way for future work. The rest of the cases are easier, where all models demonstrate robustness.

\begin{table}[t]
  \centering
  \newcolumntype{C}{>{\centering\arraybackslash}p{4.2em}}
\renewcommand{\arraystretch}{.9}
\setlength\tabcolsep{1pt}
\begin{tabular}{lCCCCCC}
\toprule
\textbf{Method} & \textbf{\ilias} & \textbf{INSTRE} & $\cR$\textbf{OP} & \textbf{\prodonem} \\ \midrule

\multicolumn{6}{c}{\textit{PE~\cite{bolya2025perception} with linear adaptation~\cite{ilias2025}}} \\ \midrule

~~\textbf{No re-rank}                 & 33.4 & 92.6 & 60.7 & 66.5  \\
~~\textbf{AMES}~\cite{ski+24}         & 41.9 & 91.9 & \textbf{79.5} & 66.8  \\
~~\textbf{LamRA}~\cite{lamra}         & 47.8 & 95.3 & 67.5 & 72.9 \\
~~\textbf{Qwen}~\cite{bai2025qwen2}   & 41.0 & 94.6 & 63.0 & 72.3  \\ \hline
\rowcolor{lightgray}
~~\textbf{LamRA}                      & 53.3 & \textbf{96.6} & 66.5 & 73.1  \\
\rowcolor{lightgray}
~~\textbf{Qwen}                       & \textbf{54.4} & 96.4 & 68.1 & \textbf{74.5}  \\
\rowcolor{lightgray}
~~\textbf{Qwen-C}                     & 50.9 & 95.2 & 65.8 & 73.6 \\

\midrule
\multicolumn{6}{c}{\textit{DINOv3~\cite{simeoni2025dinov3}}} \\
\midrule

~~\textbf{No re-rank}                 & 26.5 & 82.2 & 67.5 & 52.2   \\
~~\textbf{AMES}~\cite{ski+24}         & 34.2 & 85.1 & \textbf{82.1} & 54.1 \\
~~\textbf{LamRA}~\cite{lamra}       & 39.4 & 89.1 & 67.2 & 62.7 \\
~~\textbf{Qwen}~\cite{bai2025qwen2} & 32.8 & 89.1 & 64.5 & 62.4 \\
\hline
\rowcolor{lightgray}
~~\textbf{LamRA}                      & 41.2 & 90.5 & 67.9 & 62.2  \\
\rowcolor{lightgray}
~~\textbf{Qwen}                       & \textbf{41.7} & \textbf{90.7} & 71.1 & \textbf{64.4}  \\
\rowcolor{lightgray}
~~\textbf{Qwen-C}                     & 38.6 & 89.7 & 68.8 & 63.8 \\
\bottomrule
\end{tabular}
  \vspace{-8pt}
  \caption{
    \textbf{Performance (mAP@1k) comparison on multiple datasets} using two different global descriptors. 
    We present four instance-level to show the different behavior between training-free Qwen and LaMRA, which is fine-tuned but for more generic retrieval tasks. 
    AMES is reported using 400 local descriptors (102kB), which achieves the best performance on average.
    {\colorbox{lightgray}{Gray}} indicates the use of our task-specific prompt; otherwise, a generic prompt is used. 
    \vspace{-12pt}
    \label{tab:sota}}
\end{table}

Furthermore, to better understand the model's behavior, we test the effect of geometric consistency between the patches of the two images on Qwen’s decoder, by shuffling positional encodings. 
Specifically, we shuffle position IDs after extracting visual tokens, just before applying function $f$. Shuffling only the query reduces performance on ILIAS to 52.6 (-1.8), while shuffling both the query and database further drops it to 51.6 (-2.8). The relatively small degradation indicates that Qwen primarily depends on appearance cues, with geometric consistency playing a minor role.

\begin{figure*}[t]
    \centering
    \scalebox{1.0}{
        \scriptsize
\begin{tabular}{@{\hspace{-2pt}}c@{\hspace{1pt}}c@{\hspace{12pt}}c@{\hspace{1pt}}c@{\hspace{12pt}}c@{\hspace{1pt}}c@{\hspace{12pt}}c@{\hspace{1pt}}c@{\hspace{12pt}}c@{\hspace{1pt}}c@{\hspace{-2pt}}}

\multicolumn{10}{c}{\textbf{global \vs Qwen~~~---~~~Qwen outperforms~~~---~~~global $\rightarrow$ Qwen}} \\[0pt]
\multicolumn{2}{c}{$991 \rightarrow 4$} &
\multicolumn{2}{c}{$924 \rightarrow 11$} &
\multicolumn{2}{c}{$967 \rightarrow 87$} &
\multicolumn{2}{c}{$901 \rightarrow 0$} &
\multicolumn{2}{c}{$890 \rightarrow 0$} \\[0pt]
\includegraphics[height=1.4cm,width=1.4cm]{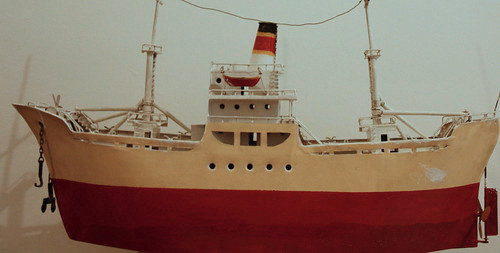} &
\includegraphics[height=1.4cm,width=1.4cm]{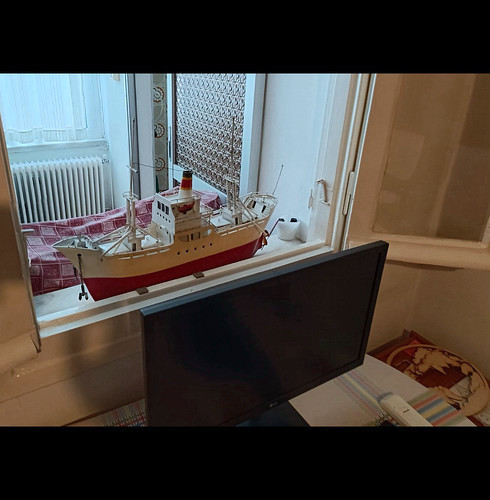} &
\includegraphics[height=1.4cm,width=1.4cm]{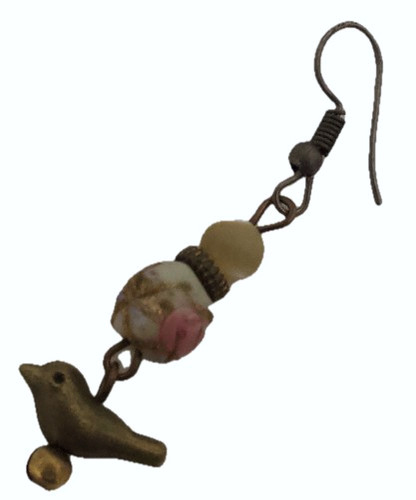} &
\includegraphics[height=1.4cm,width=1.4cm]{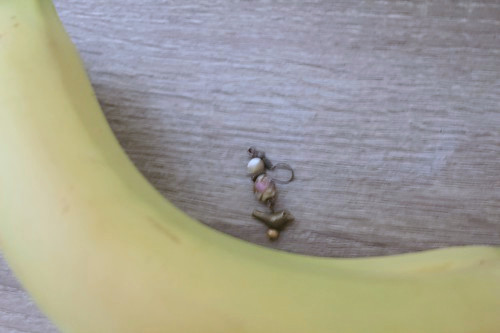} &
\includegraphics[height=1.4cm,width=1.4cm]{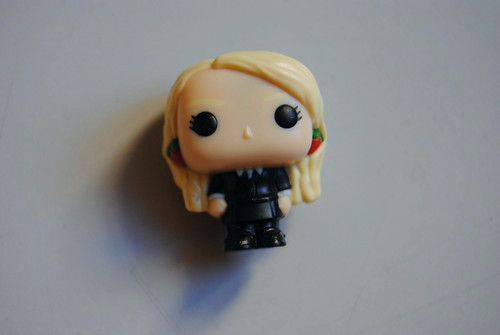} &
\includegraphics[height=1.4cm,width=1.4cm]{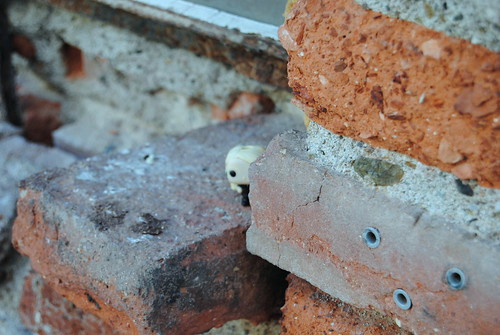} &
\includegraphics[height=1.4cm,width=1.4cm]{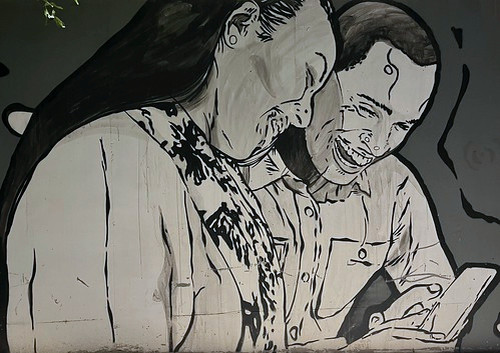} &
\includegraphics[height=1.4cm,width=1.4cm]{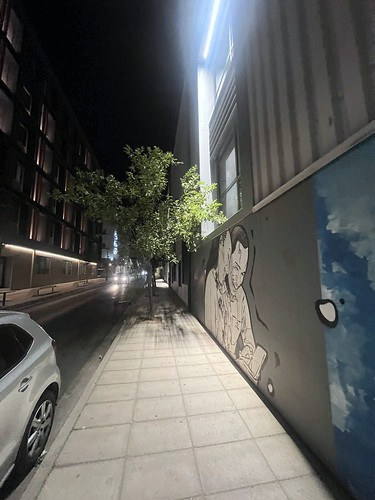} &
\includegraphics[height=1.4cm,width=1.4cm]{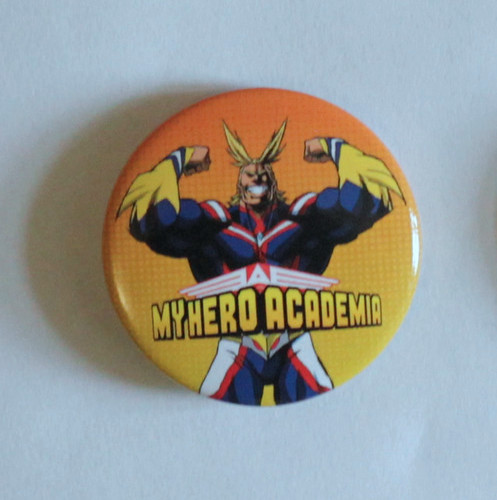} &
\includegraphics[height=1.4cm,width=1.4cm]{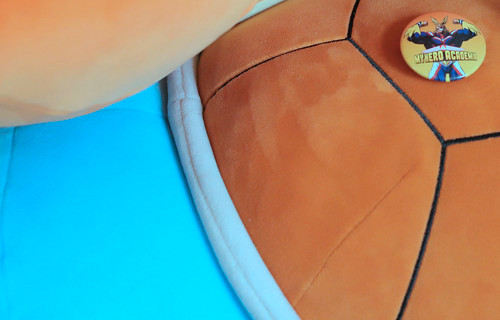} \\[0pt]

\multicolumn{10}{c}{\textbf{global \vs Qwen~~~---~~~global outperforms~~~---~~~Qwen $\rightarrow$ global}} \\[0pt]
\multicolumn{2}{c}{$963 \rightarrow 140$} &
\multicolumn{2}{c}{$764 \rightarrow 85$} &
\multicolumn{2}{c}{$189 \rightarrow 12$} &
\multicolumn{2}{c}{$200 \rightarrow 4$} &
\multicolumn{2}{c}{$122 \rightarrow 1$} \\[0pt]
\includegraphics[height=1.4cm,width=1.4cm]{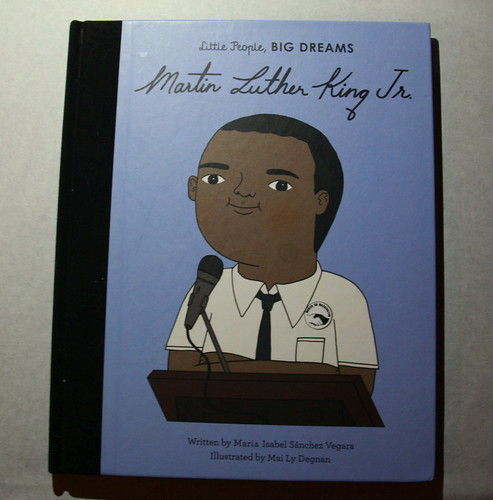} &
\includegraphics[height=1.4cm,width=1.4cm]{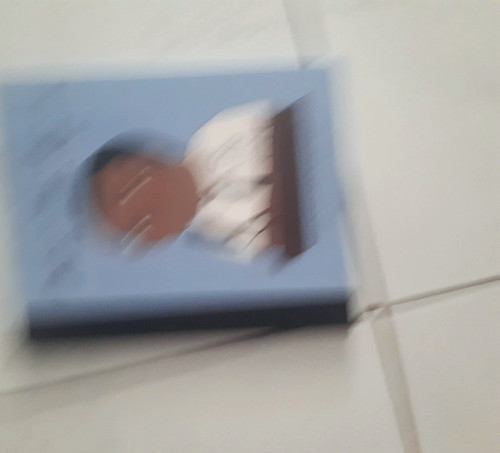} &
\includegraphics[height=1.4cm,width=1.4cm]{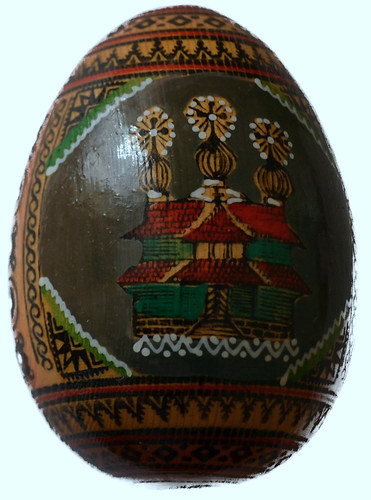} &
\includegraphics[height=1.4cm,width=1.4cm]{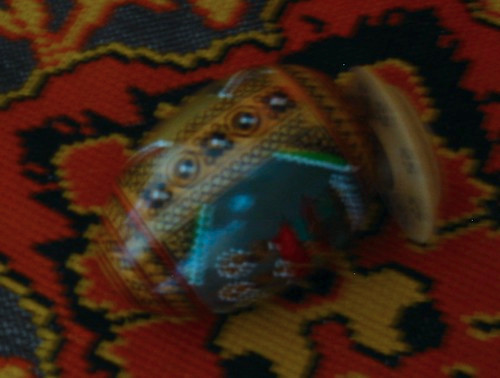} &
\includegraphics[height=1.4cm,width=1.4cm]{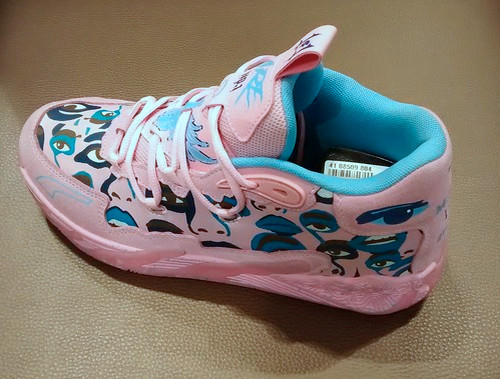} &
\includegraphics[height=1.4cm,width=1.4cm]{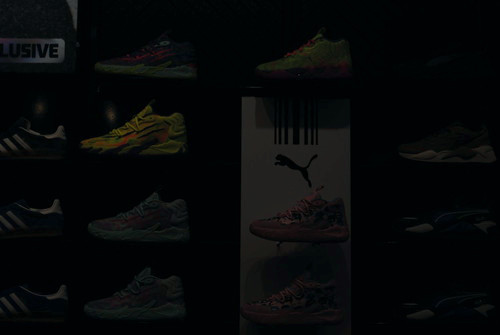} &
\includegraphics[height=1.4cm,width=1.4cm]{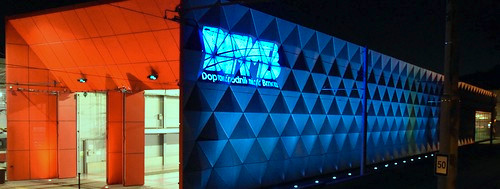} &
\includegraphics[height=1.4cm,width=1.4cm]{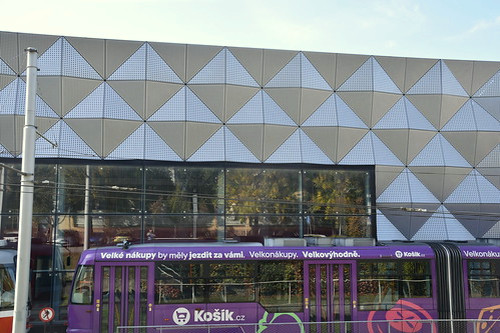} &
\includegraphics[height=1.4cm,width=1.4cm]{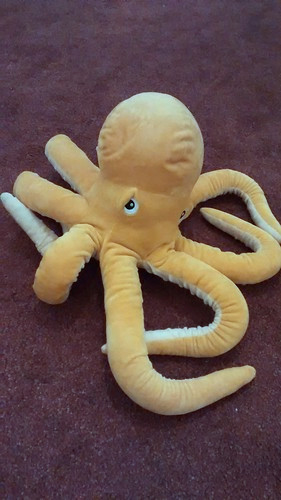} &
\includegraphics[height=1.4cm,width=1.4cm]{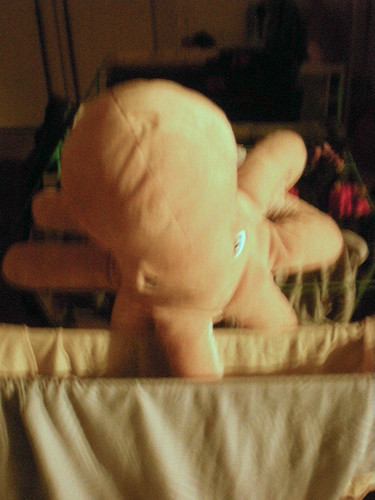} \\[0pt]

\multicolumn{10}{c}{\textbf{AMES \vs Qwen~~~---~~~Qwen outperforms~~~---~~~AMES $\rightarrow$ Qwen}} \\[0pt]
\multicolumn{2}{c}{$859 \rightarrow 0$} &
\multicolumn{2}{c}{$873 \rightarrow 3$} &
\multicolumn{2}{c}{$874 \rightarrow 15$} &
\multicolumn{2}{c}{$759 \rightarrow 26$} &
\multicolumn{2}{c}{$721 \rightarrow 1$} \\[0pt] 
\includegraphics[height=1.4cm,width=1.4cm]{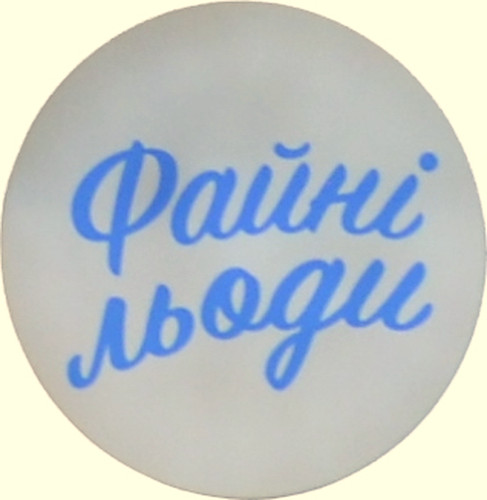} &
\includegraphics[height=1.4cm,width=1.4cm]{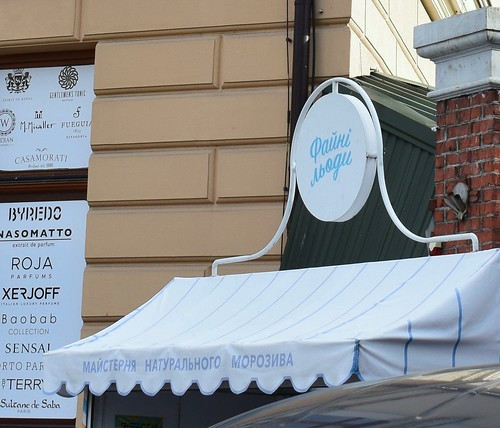} &
\includegraphics[height=1.4cm,width=1.4cm]{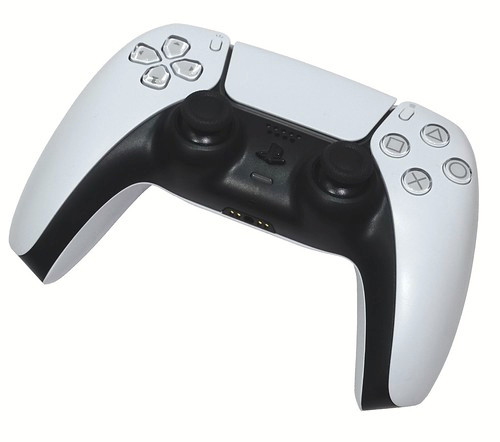} &
\includegraphics[height=1.4cm,width=1.4cm]{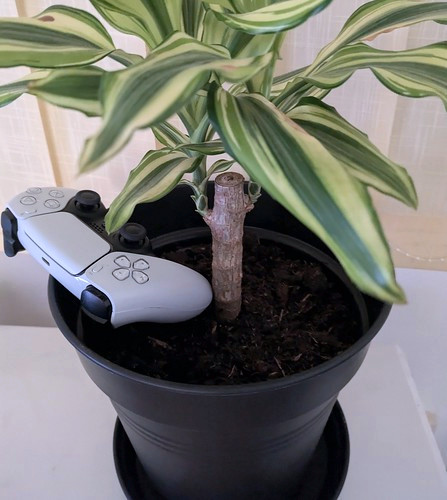} &
\includegraphics[height=1.4cm,width=1.4cm]{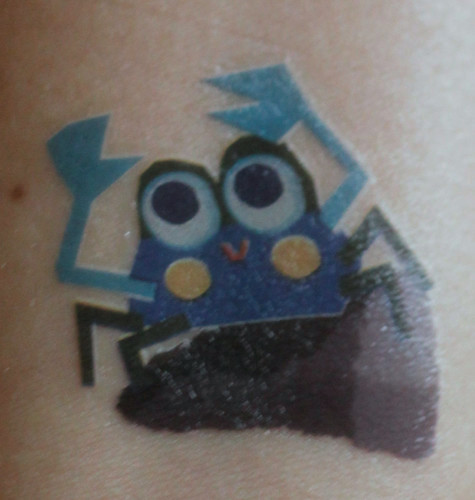} &
\includegraphics[height=1.4cm,width=1.4cm]{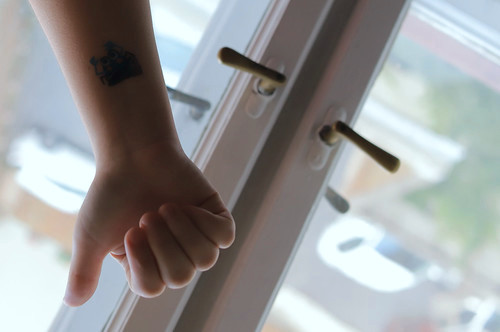} &
\includegraphics[height=1.4cm,width=1.4cm]{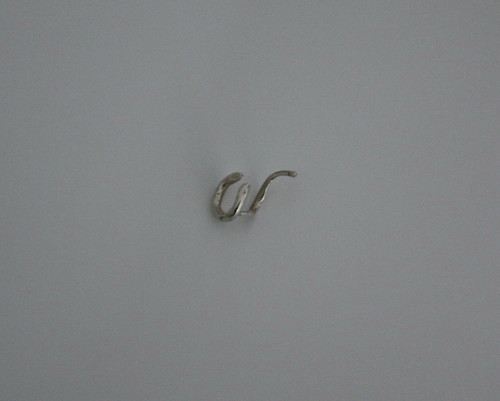} &
\includegraphics[height=1.4cm,width=1.4cm]{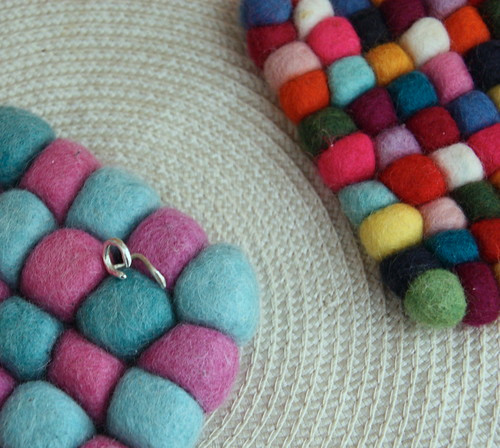} &
\includegraphics[height=1.4cm,width=1.4cm]{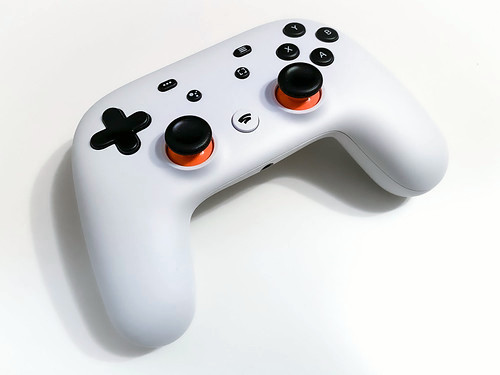} &
\includegraphics[height=1.4cm,width=1.4cm]{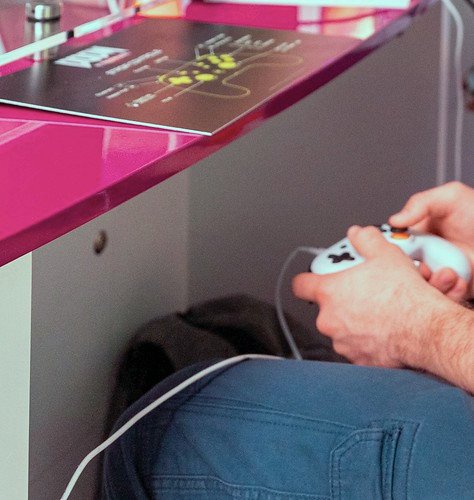} \\[0pt]

\multicolumn{10}{c}{\textbf{AMES \vs Qwen~~~---~~~AMES outperforms~~~---~~~Qwen $\rightarrow$ AMES}} \\[0pt]
\multicolumn{2}{c}{$523 \rightarrow 0$} &
\multicolumn{2}{c}{$536 \rightarrow 16$} &
\multicolumn{2}{c}{$616 \rightarrow 113$} &
\multicolumn{2}{c}{$365 \rightarrow 0$} &
\multicolumn{2}{c}{$219 \rightarrow 16$} \\[0pt]
\includegraphics[height=1.4cm,width=1.4cm]{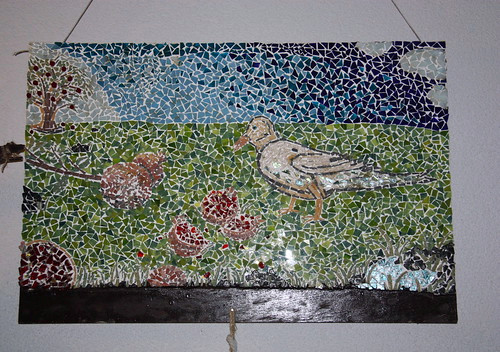} &
\includegraphics[height=1.4cm,width=1.4cm]{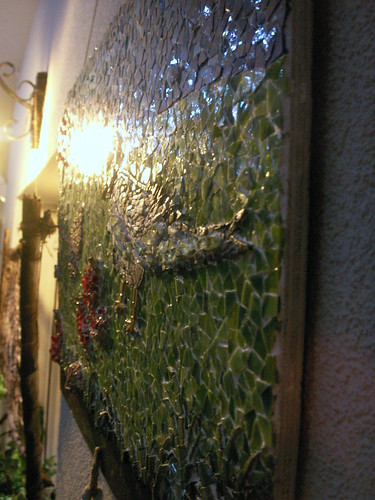} &
\includegraphics[height=1.4cm,width=1.4cm]{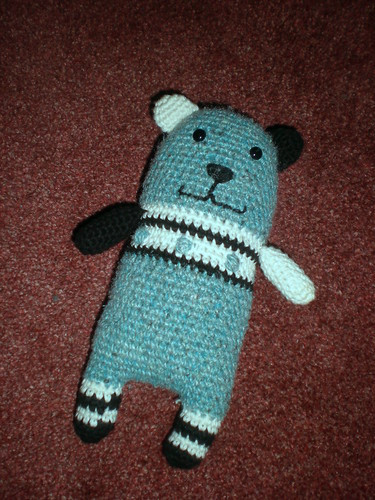} &
\includegraphics[height=1.4cm,width=1.4cm]{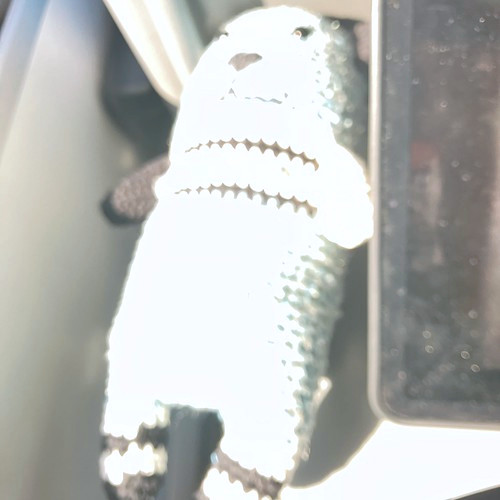} &
\includegraphics[height=1.4cm,width=1.4cm]{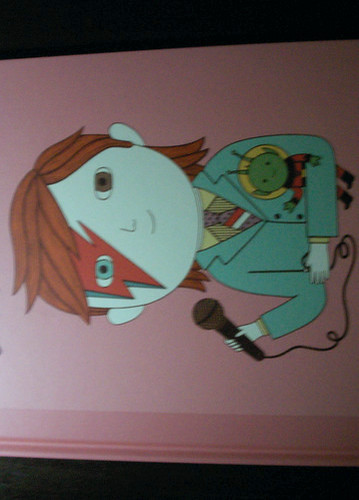} &
\includegraphics[height=1.4cm,width=1.4cm]{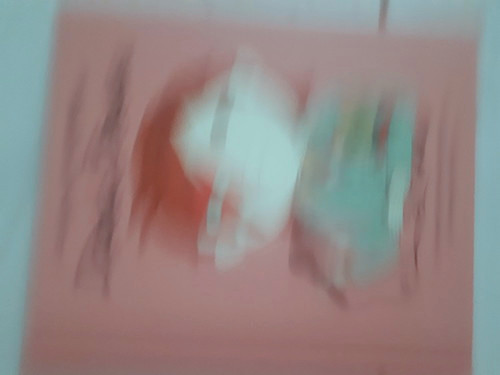} &
\includegraphics[height=1.4cm,width=1.4cm]{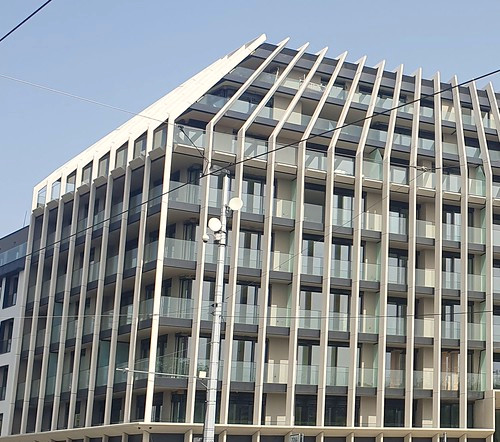} &
\includegraphics[height=1.4cm,width=1.4cm]{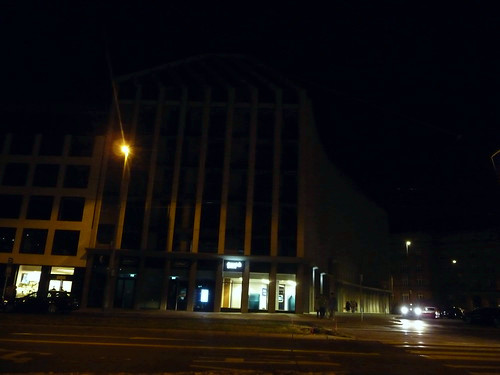} &
\includegraphics[height=1.4cm,width=1.4cm]{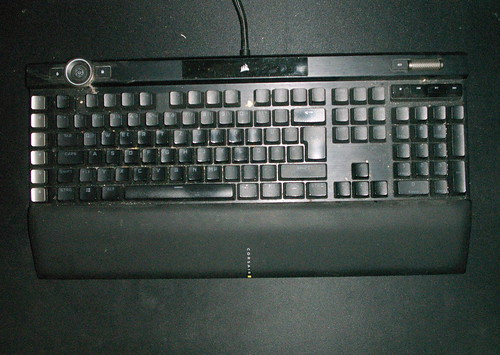} &
\includegraphics[height=1.4cm,width=1.4cm]{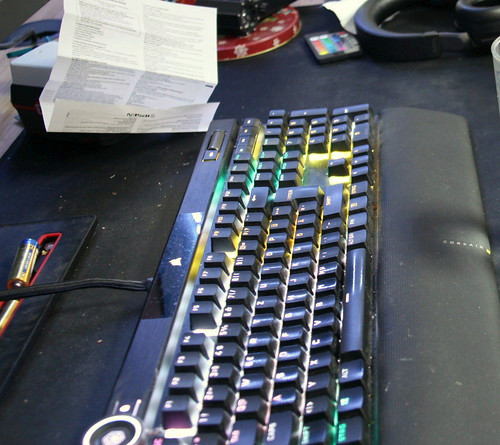} \\

\end{tabular}

    }
    \vspace{-8pt}
    \caption{\textbf{Qualitative examples where one method benefits the most compared to another.} 
    We compare global (PE) and AMES \vs Qwen by showing pairs of query and positive image. $\rightarrow$ indicates the number of negative images ranked before the positive for two models, and it goes from the weaker to the stronger model for each pair.
    \vspace{-10pt}
    \label{fig:llm_vs_global_vs_ames}}
\end{figure*}

\myparagraph{Performance on multiple datasets.}
In \cref{tab:sota}, we compare various re-ranking methods across multiple datasets. 
AMES achieves the best performance only on the landmarks dataset, \ie its training domain, while, apart from ILIAS, it provides minimal or no improvements over global due to the domain gap.
LamRA, when used with its original prompt and despite being trained for retrieval, consistently underperforms compared to the training-free Qwen, prompted appropriately. 
With our revised prompts, LamRA exhibits a substantial performance increase; however, even under this setting, Qwen generally outperforms LamRA. 
This suggests that although LamRA's training on a diverse set of queries and domains may compromise its performance relative to the training-free variant in some scenarios, it compromises its effectiveness on the most fine-grained task, such as instance-level retrieval. 
A similar trend is observed when DINOv3 is used for computing global similarities, indicating that the conclusions hold regardless of the model used for the selection of the top-$k$ candidates.

\myparagraph{Qualitative analysis.}
Delving deeper into the model comparisons, we conduct a qualitative comparison between global ranking with PE, MLLM-based re-ranking with Qwen, and AMES re-ranking. 
Figure~\ref{fig:llm_vs_global_vs_ames} presents examples where one method outperforms the others, measured by the number of negative images ranked before the positive one.

\emph{Global vs. Qwen:} 
Qwen significantly improves upon the global ranking, particularly in scenarios involving objects at varying scales, small objects, cluttered scenes, and partial occlusions. This suggests that Qwen possesses strong spatial reasoning, fine-grained recognition capabilities, and the ability to perform partial matching. 
However, there are instances where Qwen degrades performance, especially in cases affected by motion blur or significant appearance changes due to factors such as illumination. This aligns with our observations from the robustness analysis.

\emph{AMES vs. Qwen:}
Qwen consistently outperforms AMES in handling small objects, occlusions, and scenes with severe visual clutter, reinforcing the observations from the previous comparison. 
Conversely, AMES demonstrates superior performance in scenarios involving illumination changes, motion blur, and large viewpoint variations; similarly, weaknesses are exposed in our robustness analysis.

\section{Conclusion}
\label{sec:conclusions}

We present a training-free approach for instance-level image retrieval that leverages MLLMs as universal similarity estimators. By prompting them with task-oriented instructions, we repurpose them into effective image-to-image comparators capable of detecting object instances and translating their output to similarity estimation without any additional training. 
To enable scalability, we investigate indexing strategies that significantly reduce memory cost with marginal loss in performance. Digging deeper, we reveal that MLLMs, particularly Qwen, exhibit strong robustness to common visual transformations such as scale variation, occlusion, and clutter; however, they are vulnerable to drastic appearance changes, such as illumination changes and motion blur. 
Overall, our findings highlight the untapped potential of MLLMs in instance-level image retrieval, being able to scale to databases with millions of images. 


\begin{center}
\Large \textbf{Supplementary materials}
\vspace{30pt}
\end{center}

\setcounter{page}{1}
\renewcommand{\thesection}{\Alph{section}}
\renewcommand{\thefigure}{\Alph{figure}}
\renewcommand{\thetable}{\Alph{table}}

\setcounter{section}{0}
\setcounter{table}{0}
\setcounter{figure}{0}

\section{Additional results}
\label{sec:add_results}

\begin{figure}[t]
    \vspace{0pt}
    \centering
    \pgfplotsset{every tick label/.append style={font=\small}}
\begin{tikzpicture}
\begin{axis}[%
  width=1.03\linewidth,
  height=0.75\linewidth,
  ylabel={mAP@1k},
  xlabel={memory (KB)},
  grid=both,
  grid style={color=lightgray!60, dash pattern=on 2pt off 2pt},
  xlabel style={yshift=0pt},
  ylabel style={yshift=-2pt},
  xmode=log,
  ymax=59,
  legend columns=4, 
  legend style={
    anchor=south, 
    at={(0.5,0.93)}, 
    cells={anchor=west}, 
    font =\small, 
    fill opacity=1, 
    inner sep=2pt
    },
  ]

        \addlegendimage{color=applemustard, pqline} 
        \addlegendentry{392px}
        \addlegendimage{color=appleteal, pqline} 
        \addlegendentry{560px}
        \addlegendimage{color=applepurple, pqline} 
        \addlegendentry{756px}
        \addlegendimage{color=applered, pqline} 
        \addlegendentry{1008px}
        
        \addplot[color=applemustard, pqline] coordinates {
                                (515.5,	48.43) 
                                (130.6,	48.43) 
                                (66.3,	48.11) 
                                (34.2,	46.78) 
                                (18.1,	45.12) 
                                (10.0,	41.62) 
                                (6.0,	34.9)  
        };
        
        \addplot[color=appleteal, pqline] coordinates {
                                (1052.0, 53.16) 
                                (264.5,  53.04) 
                                (133.3,	 52.36) 
                                (67.6,   50.90) 
                                (34.8,   48.69) 
                                (18.4,   45.71) 
                                (10.2,   38.77) 
        };
        
        \addplot[color=applepurple, pqline] coordinates {
                                (1914., 54.2)  
                                (480.4,	54.2)  
                                (241.2,	53.62) 
                                (121.6,	52.21) 
                                (61.8,	50.35) 
                                (31.9,  47.15) 
                                (17.0,	39.3)  
        };

        \addplot[color=applered, pqline] coordinates {
                                (3403,	55.5)  
                                (852.5,	55.2)  
                                (427.3,	55.0)  
                                (214.6,	53.43) 
                                (108.3,	51.87) 
                                (55.2,	48.73) 
                                (28.6,	38.9)  
        };

    \end{axis}
\end{tikzpicture}
    \vspace{-8pt}
    \caption{\textbf{Impact of compression for different resolutions.} mAP@1k of Qwen on ILIAS for seven PQ compressions (\ie PQ \{1, 4, 8, 16, 32, 64, 128\} from right to left) and four image resolutions (\ie \{392px, 560px, 756px, 1008px\} in different colors).
    \vspace{0pt}
    \label{fig:res_pq}}
\end{figure}
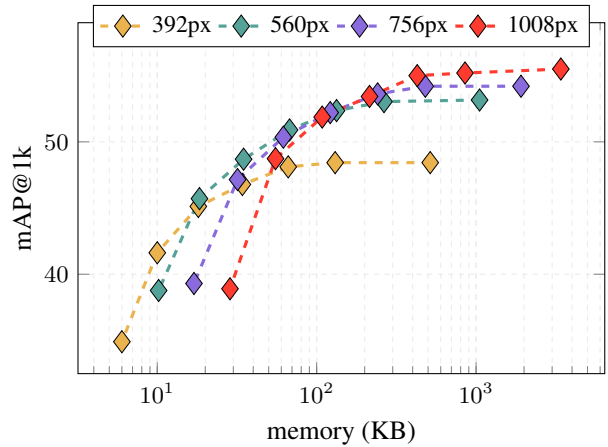

\myparagraph{Impact of compression for different resolutions.}
We investigate how input image resolution affects performance when PQ is applied with varying compression levels. \cref{fig:res_pq} shows the results for seven PQ compression rates and four input image resolutions. 
For a larger memory footprint, higher-resolution images consistently outperform lower-resolution ones when compared at equivalent memory requirements.
However, this trend does not hold for a smaller memory footprint. In particular, PQ\textsubscript{128} shows substantial performance degradation across all resolutions. Consequently, within the 10KB to 100KB memory range, a resolution of 560px yields the best performance; whereas, in the extremely low-memory regime, \ie $<$10KB, a resolution of 392px performs best. 
We conclude that aggressive PQ compression severely compromises performance. Beyond a certain compression level, it becomes more effective to reduce memory usage by lowering the input resolution rather than increasing PQ compression further.

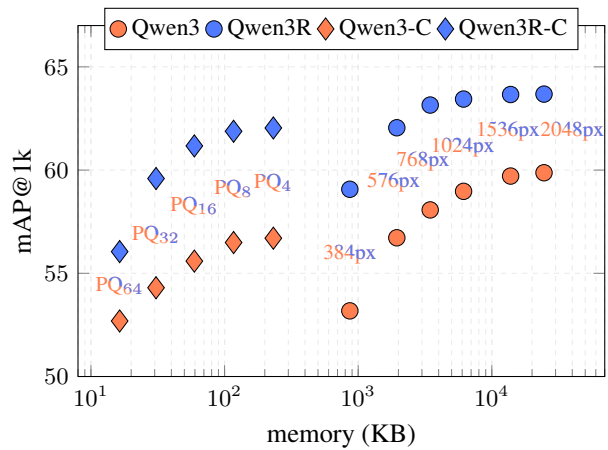
\begin{figure}[t]
    \vspace{0pt}
    \centering
    \pgfplotsset{every tick label/.append style={font=\small}}
\begin{tikzpicture}
\begin{axis}[%
  width=1.03\linewidth,
  height=0.75\linewidth,
  ylabel={mAP@1k},
  xlabel={memory (KB)},
  grid=both,
  grid style={color=lightgray!60, dash pattern=on 2pt off 2pt},
  xlabel style={yshift=0pt},
  ylabel style={yshift=-2pt},
  xmode=log,
  ymin=50,
  ymax=67,
  xmin=8,
  xmax=70000,
  legend columns=4, 
  legend style={
    anchor=south, 
    at={(0.5,0.93)}, 
    cells={anchor=west}, 
    font=\small, 
    fill opacity=1, 
    inner sep=2pt
    },
  ]

        \addlegendimage{color=qwenthreecol, normmark} 
        \addlegendentry{Qwen3}
        \addlegendimage{color=qwencol, normmark} 
        \addlegendentry{Qwen3R}
        \addlegendimage{color=qwenthreecol, pqmark} 
        \addlegendentry{Qwen3-C}
        \addlegendimage{color=qwencol, pqmark} 
        \addlegendentry{Qwen3R-C}
        
        \addplot[color=qwenthreecol, normmark] coordinates {
                                (24578.0, 59.87) 
                                (13826.0, 59.71) 
                                (6146.0, 58.97) 
                                (3458.0, 58.07) 
                                (1946.0, 56.72) 
                                (866.0,  53.18) 
        };

        \addplot[color=qwencol, normmark] coordinates {
                                (24578.0, 63.68) 
                                (13826.0, 63.66) 
                                (6146.0, 63.44) 
                                (3458.0, 63.15) 
                                (1946.0, 62.05) 
                                (866.0,  59.07) 
        };
        
        \addplot[color=qwenthreecol, pqmark] coordinates {
                                (231.7, 56.70) 
                                (116.8, 56.49) 
                                (59.4,  55.59) 
                                (30.7,  54.30) 
                                (16.4,  52.69) 
        };

        \addplot[color=qwencol, pqmark] coordinates {
                                (231.7, 62.04) 
                                (116.8, 61.88) 
                                (59.4,  61.17) 
                                (30.7,  59.59) 
                                (16.4,  56.05) 
        };

        \node [above] at (axis cs:231.7,58.5) (RGa) {\footnotesize \textcolor{qwenthreecol}{PQ$_{4}$}};
        \begin{scope}
            \clip (RGa.north west) -- (RGa.south east) -- (RGa.north east) -- cycle;
            \node at (RGa) {\footnotesize\color{qwencol}{PQ$_{4}$}};
        \end{scope}

        \node [above] at (axis cs:116.8,58.2) (RGb) {\footnotesize \textcolor{qwenthreecol}{PQ$_{8}$}};
        \begin{scope}
            \clip (RGb.north west) -- (RGb.south east) -- (RGb.north east) -- cycle;
            \node at (RGb) {\footnotesize\color{qwencol}{PQ$_{8}$}};
        \end{scope}

        \node [above] at (axis cs:59.4,57.4) (RGc) {\footnotesize \textcolor{qwenthreecol}{PQ$_{16}$}};
        \begin{scope}
            \clip (RGc.north west) -- (RGc.south east) -- (RGc.north east) -- cycle;
            \node at (RGc) {\footnotesize\color{qwencol}{PQ$_{16}$}};
        \end{scope}

        \node [above] at (axis cs:30.7,56.0) (RGd) {\footnotesize \textcolor{qwenthreecol}{PQ$_{32}$}};
        \begin{scope}
            \clip (RGd.north west) -- (RGd.south east) -- (RGd.north east) -- cycle;
            \node at (RGd) {\footnotesize\color{qwencol}{PQ$_{32}$}};
        \end{scope}

        \node [above] at (axis cs:16.4,53.5) (RGe) {\footnotesize \textcolor{qwenthreecol}{PQ$_{64}$}};
        \begin{scope}
            \clip (RGe.north west) -- (RGe.south east) -- (RGe.north east) -- cycle;
            \node at (RGe) {\footnotesize\color{qwencol}{PQ$_{64}$}};
        \end{scope}

        \node [above] at (axis cs:40000,61.0) (QTa) {\footnotesize \textcolor{qwenthreecol}{2048px}};
        \begin{scope}
            \clip (QTa.north west) -- (QTa.south east) -- (QTa.north east) -- cycle;
            \node at (QTa) {\footnotesize\color{qwencol}{2048px}};
        \end{scope}
        
        \node [above] at (axis cs:13200,61.0) (QTa) {\footnotesize \textcolor{qwenthreecol}{1536px}};
        \begin{scope}
            \clip (QTa.north west) -- (QTa.south east) -- (QTa.north east) -- cycle;
            \node at (QTa) {\footnotesize\color{qwencol}{1536px}};
        \end{scope}
        
        \node [above] at (axis cs:6050.0,60.2) (QTa) {\footnotesize \textcolor{qwenthreecol}{1024px}};
        \begin{scope}
            \clip (QTa.north west) -- (QTa.south east) -- (QTa.north east) -- cycle;
            \node at (QTa) {\footnotesize\color{qwencol}{1024px}};
        \end{scope}

        \node [above] at (axis cs:3058.0,59.5) (QTb) {\footnotesize \textcolor{qwenthreecol}{768px}};
        \begin{scope}
            \clip (QTb.north west) -- (QTb.south east) -- (QTb.north east) -- cycle;
            \node at (QTb) {\footnotesize\color{qwencol}{768px}};
        \end{scope}

        \node [above] at (axis cs:1839.0,58.5) (QTc) {\footnotesize \textcolor{qwenthreecol}{576px}};
        \begin{scope}
            \clip (QTc.north west) -- (QTc.south east) -- (QTc.north east) -- cycle;
            \node at (QTc) {\footnotesize\color{qwencol}{576px}};
        \end{scope}

        \node [above] at (axis cs:866.0,55.0) (QTd) {\footnotesize \textcolor{qwenthreecol}{384px}};
        \begin{scope}
            \clip (QTd.north west) -- (QTd.south east) -- (QTd.north east) -- cycle;
            \node at (QTd) {\footnotesize\color{qwencol}{384px}};
        \end{scope}

    \end{axis}
\end{tikzpicture}
    \vspace{-8pt}
    \caption{\textbf{Impact of compression and resolution on a better MLLM.} mAP@1k comparison of Qwen3R and Qwen3 on ILIAS. Compression is applied via PQ with 560px image resolution.
    \vspace{-5pt}
    \label{fig:qwen3}}
\end{figure}

\myparagraph{Impact of compression and resolution on better MLLMs.}
To assess the effectiveness of our compression strategy and the transferability of our findings to more recent MLLM architectures, we evaluate Qwen3R and Qwen3 under different PQ compression levels and image resolutions. We choose image resolutions that are divisible by the employed ViT's patch size.
\cref{fig:qwen3} shows the results on ILIAS. Qwen3R consistently outperforms Qwen3, highlighting the effectiveness of task-specific model adjustments and large-scale fine-tuning. Compared with the results from the main paper, both models consistently outperform their predecessor, Qwen, across all memory footprints.

\begin{figure*}[t]
  \centering
  \vspace{-5pt}
  \scalebox{0.95}{
    \pgfplotsset{every tick label/.append style={font=\scriptsize}}

\newcommand{\addFilling}[4]{
    \addplot [
        fill=gray, fill opacity=0.1, draw=none,
    ] coordinates {
        ({#1}, 100)
        ({#2}, 100)
        ({#2}, -1)
        ({#1}, -1)
    } -- cycle;
}

\begin{tikzpicture}
\begin{axis}[
    width=1.05\linewidth,
    height=0.282\linewidth,
    ymin=15, ymax=100,
    ytick={10,30,50,70,90}, 
    ylabel={mAP@1k},
    ylabel style={yshift=-5pt},
    yticklabel style={font=\scriptsize,xshift=2pt},
    xtick=data,
    xticklabel style={rotate=45, anchor=east, font=\scriptsize, yshift=-2pt, xshift=2pt, name=xlabel\ticknum},
    symbolic x coords=\xcoords,
    symbolic x coords={
        art       - paper model   (22),
        art       - sculpture     (59),
        art       - digital       (11),
        art       - craft         (10),
        art       - painting      (58),
        art       - textile       (8),
        test,
        landmark  - architecture  (32),
        landmark  - sign          (20),
        landmark  - public art    (110),
        test,
        toy       - boardgame     (10),
        toy       - action figure (38),
        toy       - trading       (29),
        toy       - other         (13),
        toy       - stuffed       (36),
        toy       - sport         (6),
        toy       - playset       (15),
        toy       - vehicle       (8),
        test,
        fashion   - other         (2),
        fashion   - outwear       (38),
        fashion   - accessory     (46),
        fashion   - tattoo        (9),
        fashion   - footwear      (14),
        fashion   - jewelry       (34),
        test,
        household - textile       (5),
        household - decor         (19),
        household - accessory     (14),
        household - light         (13),
        household - container     (18),
        household - other         (5),
        household - tableware     (25),
        test,
        technology- other         (3),
        technology- gaming        (8),
        technology- appliances    (21),
        technology- automation    (30),
        technology- multimedia    (18),
        technology- gadget        (7),
        technology- peripheral    (10),
        test,
        media     - stamp         (12),
        media     - other         (4),
        media     - book          (48),
        media     - sticker       (22),
        media     - recording     (7),
        test,
        product   - hygiene       (10),
        product   - writing tool  (10),
        product   - perfume       (8),
        product   - food          (21),
        product   - drink         (31),
        product   - other         (3),
    },
    xticklabels={
         paper model,
         digital,
         craft,
         painting,
         sculpture,
         textile,
         architecture,
         public art,
         sign,
         stuffed,
         playset,
         action ,
         boardgame,
         other,
         trading,
         vehicle,
         sport,
         footwear,
         outwear,
         tattoo,
         other,
         accessory,
         jewelry,
         light,
         container,
         accessory,
         decor,
         textile,
         other,
         tableware,
         appliances,
         automation,
         multimedia,
         other,
         gadget,
         gaming,
         peripheral,
         other,
         book,
         recording,
         stamp,
         sticker,
         perfume,
         writing tool,
         other,
         hygiene,
         food,
         drink,
    },
    grid=major,
    grid style={color=lightgray!60, dash pattern=on 2pt off 2pt},
    enlarge x limits={abs=0.2cm},
    legend style={
        at={(0.5,0.805)}, 
        anchor=south, 
        legend columns=-1,  
        /tikz/every even column/.append style={column sep=0.2cm}, 
        font=\footnotesize,
        inner sep=1pt,  
        outer sep=1pt 
    },
    legend cell align={left},
    name=dist,
]

\pgfplotstableread[col sep=comma]{data/performance_per_subdomain.csv}\loadeddata

\addplot+[
    only marks, 
    mark=*, 
    color=qwencol, 
    mark options={fill=qwencol, draw=black}, 
] table [
    x=subdomain, 
    y=qwen,
    col sep=comma
] {\loadeddata};
\addlegendentry{Qwen}

\addplot+[
    only marks, 
    mark=*, 
    color=lamracol, 
    mark options={fill=lamracol, draw=black}, 
] table [
    x=subdomain, 
    y=lamra,
    col sep=comma
] {\loadeddata};
\addlegendentry{LamRA}

\addplot+[
    only marks, 
    mark=*, 
    color=indexcol, 
    mark options={fill=indexcol, draw=black}, 
] table [
    x=subdomain, 
    y=qwenc,
    col sep=comma
] {\loadeddata};
\addlegendentry{Qwen-C}

\addplot+[
    only marks, 
    mark=*, 
    color=amescol, 
    mark options={fill=amescol, draw=black}, 
] table [
    x=subdomain, 
    y=ames,
    col sep=comma
] {\loadeddata};
\addlegendentry{AMES}

\addplot+[
    only marks, 
    mark=*, 
    color=pecol,
    mark options={fill=pecol, draw=black},
] table [
    x=subdomain, 
    y=global,
    col sep=comma
] {\loadeddata};
\addlegendentry{PE}

\addplot [
    fill=gray, fill opacity=0.1,draw=none,
] coordinates {
    ({art       - paper model (22)}, 100)
    ({art         - textile      (8)}, 100)
    ({art         - textile      (8)}, -1)
    ({art       - paper model   (22)}, -1)
} -- cycle;

\addplot [
    fill=gray, fill opacity=0.1,  draw=none,
] coordinates {
    ({landmark    - architecture (32)}, 100)
    ({landmark    - public art    (110)}, 100)
    ({landmark    - public art    (110)}, -1)
    ({landmark    - architecture (32)}, -1)
} -- cycle;

\addplot [
    fill=gray, fill opacity=0.15,draw=none,
] coordinates {
    ({toy       - boardgame     (10)}, 100)
    ({toy       - vehicle       (8)}, 100)
    ({toy       - vehicle       (8)}, -1)
    ({toy       - boardgame     (10)}, -1)
} -- cycle;

\addplot [
    fill=gray, fill opacity=0.1, draw=none,
] coordinates {
    ({fashion   - other         (2)}, 100)
    ({fashion     - jewelry      (34)}, 100)
    ({fashion     - jewelry      (34)}, -1)
    ({fashion   - other         (2)}, -1)
} -- cycle;

\addplot [
    fill=gray, fill opacity=0.1,draw=none,
] coordinates {
    ({household - textile       (5)}, 100)
    ({household - tableware         (25)}, 100)
    ({household - tableware         (25)}, -1)
    ({household - textile       (5)}, -1)
} -- cycle;

\addplot [
    fill=gray, fill opacity=0.1,draw=none,
] coordinates {
    ({technology- other         (3)}, 100)
    ({technology- peripheral        (10)}, 100)
    ({technology- peripheral        (10)}, -1)
    ({technology- other         (3)}, -1)
} -- cycle;

\addplot [
    fill=gray, fill opacity=0.1,draw=none,
] coordinates {
    ({media     - stamp         (12)}, 100)
    ({media     - recording     (7)}, 100)
    ({media     - recording     (7)}, -1)
    ({media     - stamp         (12)}, -1)
} -- cycle;

\addplot [
    fill=gray, fill opacity=0.1,draw=none,
] coordinates {
    ({product   - hygiene       (10)}, 100)
    ({product   - other         (3)}, 100)
    ({product   - other         (3)}, -1)
    ({product   - hygiene       (10)}, -1)
} -- cycle;

\end{axis}

\foreach \X/\value in {0.96/art, 2.64/landmark, 4.60/toys, 7.02/fashion, 9.30/household, 11.72/technology,  13.83/media,  15.8/product} {
    \node[anchor=north, font=\footnotesize] at (\X,3.70cm) {\scalebox{1.}{\value}};
}
\end{tikzpicture}
  }
  \vspace{-17pt}
  \caption{\textbf{Performance comparison per category.} mAP@1k averaged over objects in the same mid-level taxonomy category of ILIAS, grouped by their primary-level category size, with sorting within each group by Qwen performance. Comparison between Qwen with and without compression (Qwen-C), AMES, LamRA, and PE.
  \label{fig:subdomains}
  \vspace{-10pt}
  }
\end{figure*}
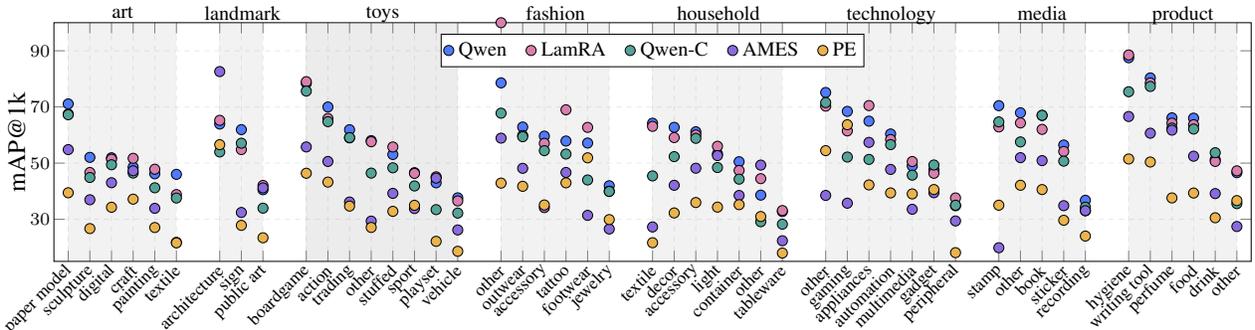

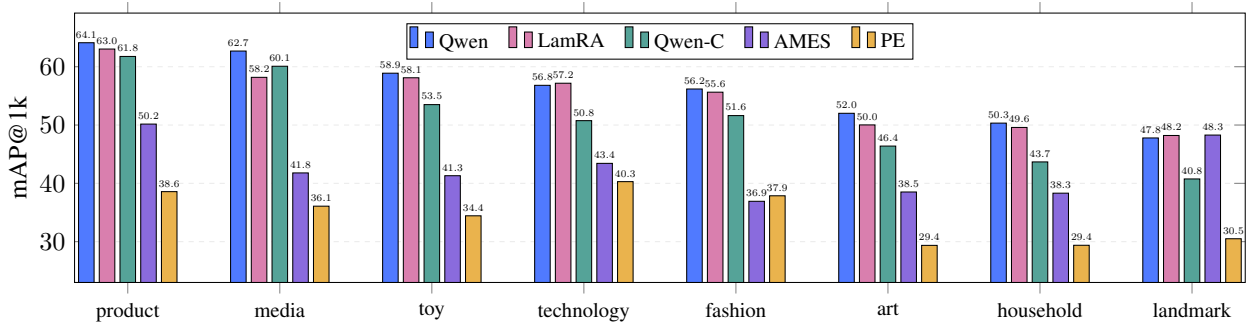
\begin{figure*}[!ht]
    \centering
    \scalebox{0.98}{
       \begin{tikzpicture}
\begin{axis}[
    width=1\linewidth,
    height=0.3\linewidth,
    bar width=6pt, 
    ylabel={mAP@1k},
    ymin=23, ymax=67,
    ytick={10,20,30,40,50,60}, 
    ytick distance=5, 
    ybar=2,
    enlarge y limits={upper, value=0.05},
    xtick=data,
    symbolic x coords={  
        product,
        media,
        toy,
        technology,
        fashion,
        art,
        household,
        landmark,
    },
    ylabel style={yshift=-2pt},
    xtick distance = 0.4,
    xticklabel style={
        font=\footnotesize},
    enlarge x limits=0.05, 
    nodes near coords,
    every node near coord/.append style={
        scale=0.56, 
        color=black,        
        font=\scriptsize, 
        /pgf/number format/.cd, 
        fixed zerofill, 
        precision=1,
    },
    legend style={
        at={(0.5,0.97)}, 
        anchor=north, 
        legend columns=-1,  
        /tikz/every even column/.append style={column sep=0.2cm}, 
        font=\footnotesize,
        inner sep=1pt,  
        outer sep=1pt 
    },
    legend cell align={left}, 
    ymajorgrids=true,
    grid style={color=lightgray!60, dash pattern=on 2pt off 2pt},
]

\pgfplotstableread[col sep=comma]{data/performance_per_domain.csv}\loadeddata

\addplot+[ybar, color=black, fill=qwencol] table[x=domain,y=qwen] {\loadeddata};
\addplot+[ybar, color=black, fill=lamracol] table[x=domain,y=lamra] {\loadeddata};
\addplot+[ybar, color=black, fill=indexcol] table[x=domain,y=qwenc] {\loadeddata};
\addplot+[ybar, color=black, fill=amescol] table[x=domain,y=ames] {\loadeddata};
\addplot+[ybar, color=black, fill=pecol] table[x=domain,y=pe] {\loadeddata};

\legend{Qwen, LamRA, Qwen-C, AMES, PE}
\end{axis}
\end{tikzpicture}
    }
    \vspace{-12pt}
    \caption{\textbf{Performance comparison per domain.} mAP@1k  averaged over objects in the same coarser taxonomy level of ILIAS, sorted by Qwen performance. Comparison between Qwen without and with compression (Qwen-C), AMES, LamRA, and PE.
    \label{fig:performance_barplot}
    \vspace{-10pt}
    }
\end{figure*}

\myparagraph{Performance comparison over different semantic categories and domains.}
In \cref{fig:subdomains}, we compare methods by averaging query performance across the mid-level categories defined in the ILIAS taxonomy. 
AMES outperforms the LLM-based approaches only in the landmark domain, particularly in the architecture category, which aligns with its training domain, indicating a strong domain bias.  
Although AMES generally improves upon global across most categories, there are some exceptions, \eg jewelry, footwear, or textiles, presumably due to their semantic distance from its training domain. On the contrary, Qwen-based re-ranking consistently improves performance across all categories, showcasing the value of large-scale and diverse domain training. Interestingly, LamRA only surpasses Qwen on a few categories, showcasing once more that its generic training compromises performance on the specific task of instance-level retrieval, which is the focus of this work.

Additionally, \cref{fig:performance_barplot} displays the aggregated performance on the ILIAS domains according to the coarser taxonomy level. Qwen consistently outperforms all other models in most domains, demonstrating robust generalization in different object types.
As in the subdomains, AMES achieves the best performance on landmarks, while LamRA slightly outperforms Qwen on technology. 

\begin{table}[t]
  \centering
  \hspace{0pt}
  \scalebox{1.}{
    \newcolumntype{C}{>{\centering\arraybackslash}p{3em}}
\begin{tabular}{lCCC}
\toprule
\textbf{Model} & \textbf{392px} & \textbf{560px} & \textbf{756px} \\ \midrule
~~\textbf{No-rerank}    & 33.4 & 33.4 & 33.4  \\
~~\textbf{Qwen-3B}      & 19.7 & 24.3 & 25.3  \\
~~\textbf{Qwen-7B}      & 48.5 & 53.2 & 54.4  \\
~~\textbf{Qwen-32B}     & 48.1 & 51.3 & 52.0  \\
~~\textbf{Qwen-72B}     & 54.2 & 54.5 & OOM  \\
\bottomrule
\end{tabular}
  }
  \vspace{-5pt}
  \caption{
    \textbf{Impact of MLLM size.} mAP@1k of four Qwen variants with three image resolutions on ILIAS. No re-ranking is provided for comparison. OOM stands for out-of-memory.
    \vspace{-12pt}
    \label{tab:model_size}}
\end{table}

\myparagraph{Impact of MLLM size.}
\cref{tab:model_size} reports the results of Qwen in various sizes with different image resolutions. 
The largest models achieve the best performance. In contrast, the smallest variant performs notably worse than the no re-ranking baseline, suggesting that the fine-grained nature of the task requires sufficient model capacity. Nevertheless, model size alone does not fully explain performance trends; scaling from 7B to 32B does not bring a proportional boost. This is consistent for all image resolutions.

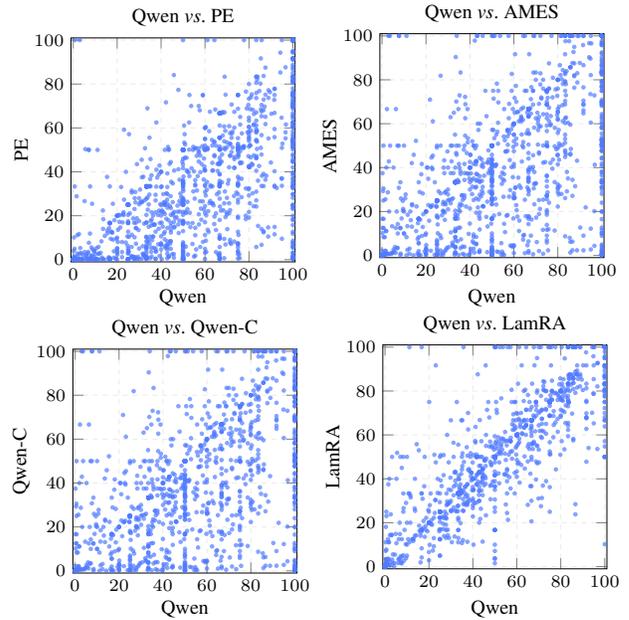
\begin{figure}[t]
  \centering
  \vspace{-5pt}
  \scalebox{0.94}{
    \begin{tabular}{@{\zsp}c@{\zsp}c@{\zsp}}

\begin{tikzpicture}
    \footnotesize
    \begin{axis}[
        width=0.57\linewidth,
        height=0.57\linewidth,
        xmin=-1,
        xmax=101,
        ymin=-1,
        ymax=101,
        title={\footnotesize Qwen \vs PE},
        grid=both,
        grid style={color=lightgray!60, dash pattern=on 2pt off 2pt},
        xtick={0, 20, 40, 60, 80, 100},
        ytick={0, 20, 40, 60, 80, 100},
        title style = {yshift = -4pt},
        xlabel = {Qwen},
        ylabel = {PE},
        xlabel style={yshift=2pt},
        ylabel style={yshift=-3pt},
        yticklabel style={font=\scriptsize},
        xticklabel style=
    {/pgf/number format/1000 sep=,anchor=north,font=\scriptsize},
        legend pos=north west,
        legend style={cells={anchor=east}, font=\scriptsize, fill opacity=0.7, row sep=-1pt},
    ]
    \addplot[only marks, mark=*, opacity=0.7, mark size=0.7, color=appleblue] 
    table[x=qwen, y=global] {./data/pairwise.csv};
    \end{axis}
\end{tikzpicture}

&

\begin{tikzpicture}
    \footnotesize
    \begin{axis}[
        width=0.57\linewidth,
        height=0.57\linewidth,
        xmin=-1,
        xmax=101,
        ymin=-1,
        ymax=101,
        title={\footnotesize Qwen \vs AMES},
        grid=both,
        grid style={color=lightgray!60, dash pattern=on 2pt off 2pt},
        xtick={0, 20, 40, 60, 80, 100},
        ytick={0, 20, 40, 60, 80, 100},
        title style = {yshift = -4pt},
        xlabel = {Qwen},
        ylabel = {AMES},
        xlabel style={yshift=0.3pt},
        ylabel style={yshift=-3pt},
        yticklabel style={font=\scriptsize},
        xticklabel style=
    {/pgf/number format/1000 sep=,anchor=north,font=\scriptsize},
        legend pos=north west,
        legend style={cells={anchor=east}, font=\scriptsize, fill opacity=0.7, row sep=-1pt},
    ]
    \addplot[only marks, mark=*, opacity=0.7, mark size=0.7, color=appleblue] 
    table[x=qwen, y=ames] {./data/pairwise.csv};
    \end{axis}
\end{tikzpicture}

\vspace{-5pt}
\\

\begin{tikzpicture}
    \footnotesize
    \begin{axis}[
        width=0.57\linewidth,
        height=0.57\linewidth,
        xmin=-1,
        xmax=101,
        ymin=-1,
        ymax=101,
        title={\footnotesize Qwen \vs Qwen-C},
        grid=both,
        grid style={color=lightgray!60, dash pattern=on 2pt off 2pt},
        xtick={0, 20, 40, 60, 80, 100},
        ytick={0, 20, 40, 60, 80, 100},
        title style = {yshift = -4pt},
        xlabel = {Qwen},
        ylabel = {Qwen-C},
        xlabel style={yshift=2pt},
        ylabel style={yshift=-3pt},
        yticklabel style={font=\scriptsize},
        xticklabel style=
    {/pgf/number format/1000 sep=,anchor=north,font=\scriptsize},
        legend pos=north west,
        legend style={cells={anchor=east}, font=\scriptsize, fill opacity=0.7, row sep=-1pt},
    ]
    \addplot[only marks, mark=*, opacity=0.7, mark size=0.7, color=appleblue] 
    table[x=qwen, y=ames] {./data/pairwise.csv};
    \end{axis}
\end{tikzpicture}

&

\begin{tikzpicture}
    \footnotesize
    \begin{axis}[
        width=0.57\linewidth,
        height=0.57\linewidth,
        xmin=-1,
        xmax=101,
        ymin=-1,
        ymax=101,
        title={\footnotesize Qwen \vs LamRA},
        grid=both,
        grid style={color=lightgray!60, dash pattern=on 2pt off 2pt},
        xtick={0, 20, 40, 60, 80, 100},
        ytick={0, 20, 40, 60, 80, 100},
        title style = {yshift = -4pt},
        xlabel = {Qwen},
        ylabel = {LamRA},
        xlabel style={yshift=0.3pt},
        ylabel style={yshift=-3pt},
        yticklabel style={font=\scriptsize},
        xticklabel style=
    {/pgf/number format/1000 sep=,anchor=north,font=\scriptsize},
        legend pos=north west,
        legend style={cells={anchor=east}, font=\scriptsize, fill opacity=0.7, row sep=-1pt},
    ]
    \addplot[only marks, mark=*, opacity=0.7, mark size=0.7, color=appleblue] 
    table[x=qwen, y=lamra] {./data/pairwise.csv};
    \end{axis}
\end{tikzpicture}

\end{tabular}
  }
  \vspace{-13pt}
  \caption{\textbf{In-depth performance comparison.} AP per query comparison between Qwen and PE, AMES, Qwen-C, and LamRA. Each point corresponds to one query.
  \vspace{-15pt}
  \label{fig:pairwise}}
\end{figure}

\myparagraph{In-depth performance comparison.}
\cref{fig:pairwise} presents scatterplots of AP per query, comparing Qwen with four other approaches. 
The superiority of Qwen over PE and AMES is evident, as the majority of points lie in the bottom-right region, indicating a higher AP for Qwen. Nonetheless, for PE, performance in several queries is degraded after re-ranking. Also, there are several queries where larger boosts are demonstrated with AMES. Similar observations can be inferred from the comparison with Qwen-C. Qwen and LamRA exhibit quite correlated performance, as reflected by the strong concentration along the diagonal.

\begin{figure}[t]
    \vspace{3pt}
    \centering
    \pgfplotsset{every tick label/.append style={font=\small}}
\begin{tikzpicture}
\begin{axis}[%
  width=1\linewidth,
  height=0.7\linewidth,
  ylabel={mAP@1k},
  xlabel={top-$k$},
  grid=both,
  grid style={color=lightgray!60, dash pattern=on 2pt off 2pt},
  xmode=log,
  ylabel near ticks, xlabel near ticks, 
  log ticks with fixed point,
  ymax=60,
  legend columns=4, 
  legend style={
    anchor=south, 
    at={(0.5,0.93)}, 
    cells={anchor=west}, 
    font =\small, 
    fill opacity=1, 
    inner sep=2pt
    },
  ]

        \addlegendimage{color=qwencol, markline} 
        \addlegendentry{Qwen}
        \addlegendimage{color=appleteal, markline} 
        \addlegendentry{Qwen-C}
        \addlegendimage{color=applepurple, markline} 
        \addlegendentry{AMES}
        
        \addplot[color=qwencol, markline] coordinates {
                                (10,     37.20)
                                (50,     44.62)
                                (100,	 47.60)
                                (200,    50.03)
                                (400,    52.37)
                                (1000,   54.75)
                                (5000,   57.10)
        };
        
        \addplot[color=appleteal, markline] coordinates {
                                (10,     38.49)
                                (50,     43.76)
                                (100,	 45.91)
                                (200,    47.75)
                                (400,    49.33)
                                (1000,   50.94)
                                (5000,   52.67)
        };

        \addplot[color=amescol, markline] coordinates {
                                (10,     36.08)
                                (50,     38.49)
                                (100,	 39.38)
                                (200,    40.08)
                                (400,    40.75)
                                (1000,   41.61)
                                (5000,   42.37)
        };

    \end{axis}
\end{tikzpicture}
    \vspace{-7pt}
    \caption{\textbf{Impact of re-ranking.} mAP@1k of three models on ILIAS for increasing shortlist sizes, up to 5k images per query.
    \vspace{0pt}
    \label{fig:topk}}
\end{figure}
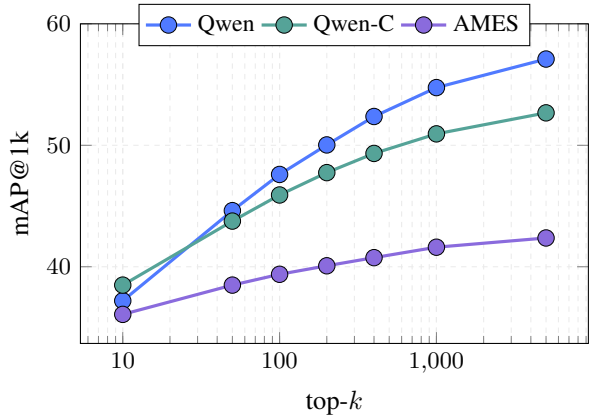

\myparagraph{Increasing the number of re-ranked images.} \cref{fig:topk} demonstrates re-ranking for different top-$k$, going as low as 10 and up to 5k shortlist images. 
Increasing $k$ generally leads to improved performance for both Qwen models. Notably, we observe a performance crossover: while Qwen-C yields higher performance when re-ranking fewer candidates ($k<50$), the standard Qwen model benefits significantly more from a larger candidate list, continuing to improve up to $k=5000$. In contrast, AMES saturates early, suggesting that it struggles to effectively distinguish increasing numbers of hard negatives.

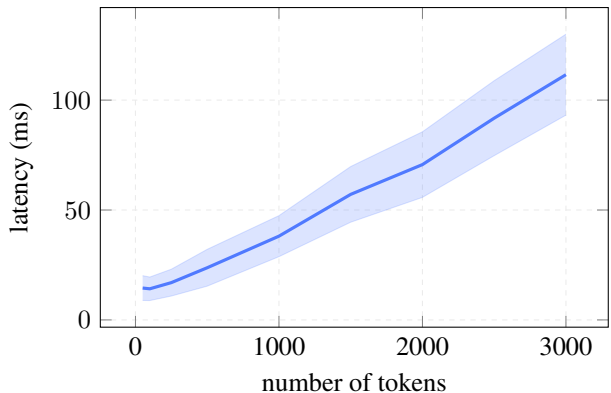
\begin{figure}[t]
    \vspace{8pt}
    \centering
    \begin{tikzpicture}
  \begin{axis}[
    width=1\linewidth,
    height=0.7\linewidth,
    xlabel={number of tokens},
    ylabel={latency (ms)},
    grid=both,
    grid style={color=lightgray!60, dash pattern=on 2pt off 2pt},
    xtick = {0,1000,2000,3000},
    xticklabels = {0,1000,2000,3000},
    error bars/.cd]

    \addplot+[mark=none, solid, line width=1.2pt, color=appleblue] coordinates {
    (50,14.49358)
    (100,14.16047)
    (250,16.96803)
    (500,23.73661)
    (1000,38.10897)
    (1500,57.11678)
    (2000,70.66612)
    (2500,91.78160)
    (3000,111.56978)
    };
    
    \addplot+[name path=lower, mark=none, solid, opacity=0.2, color=appleblue] coordinates {
    (50,14.49358-5.64055)
    (100,14.16047-5.34777)
    (250,16.96803-6.10026)
    (500,23.73661-8.37694)
    (1000,38.10897-9.33424)
    (1500,57.11678-12.66262)
    (2000,70.66612-14.92227)
    (2500,91.78160-17.05742)
    (3000,111.56978-18.41331)
    };
    
    \addplot+[name path=upper, mark=none, solid, opacity=0.2, color=appleblue] coordinates {
    (50,14.49358+5.64055)
    (100,14.16047+5.34777)
    (250,16.96803+6.10026)
    (500,23.73661+8.37694)
    (1000,38.10897+9.33424)
    (1500,57.11678+12.66262)
    (2000,70.66612+14.92227)
    (2500,91.78160+17.05742)
    (3000,111.56978+18.41331)
    };
    
    \addplot[fill=appleblue, fill opacity=0.2] fill between[of=upper and lower];
  \end{axis}
\end{tikzpicture}
    \vspace{-2pt}
    \caption{\textbf{Latency \vs prompt length.} The plot shows the latency of the LLM as a function of the number of prompt tokens. The central line represents the mean latency, while the upper and lower bounds indicate the variance.  
    \vspace{-5pt}
    \label{fig:timings}}
\end{figure}

\myparagraph{Latency vs. prompt length.}
We analyze how the length of the prompt affects the processing time, measuring the latency of the LLM as a function of the input prompt length. Only the LLM component is considered in this measurement, as the representations extracted by the vision encoder are considered pre-computed. As shown in \cref{fig:timings}, latency increases proportionally with the number of prompt tokens, indicating that longer textual inputs directly contribute to a higher computational overhead. In our case, the total prompt length is approximately 720 and 1200 tokens for image resolutions of 560px and 756px, respectively.

\begin{figure*}[t]
    \centering
    \scalebox{1.05}{
        \scriptsize
\begin{tabular}{@{\hspace{-2pt}}c@{\hspace{1pt}}c@{\hspace{12pt}}c@{\hspace{1pt}}c@{\hspace{12pt}}c@{\hspace{1pt}}c@{\hspace{12pt}}c@{\hspace{1pt}}c@{\hspace{12pt}}c@{\hspace{1pt}}c@{\hspace{-2pt}}}

\multicolumn{10}{c}{\textbf{global \vs Qwen~~~---~~~Qwen outperforms~~~---~~~global $\rightarrow$ Qwen}} \\[3pt]
\multicolumn{2}{c}{$988 \rightarrow 0$} &
\multicolumn{2}{c}{$987 \rightarrow 0$} &
\multicolumn{2}{c}{$977 \rightarrow 14$} &
\multicolumn{2}{c}{$947 \rightarrow 8$} &
\multicolumn{2}{c}{$941 \rightarrow 4$} \\[3pt]
\includegraphics[height=1.65cm,width=1.5cm]{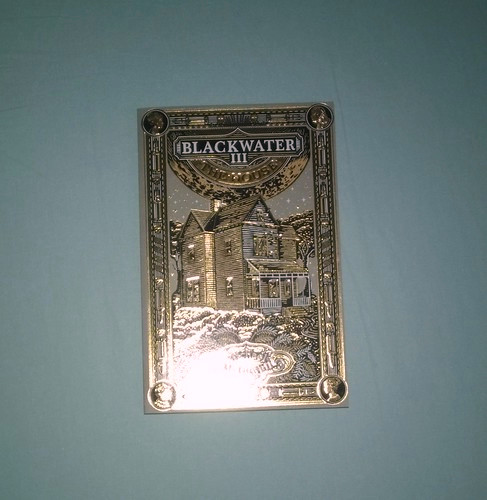} &
\includegraphics[height=1.65cm,width=1.5cm]{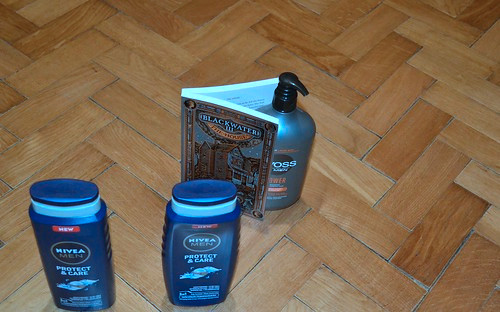} &
\includegraphics[height=1.65cm,width=1.5cm]{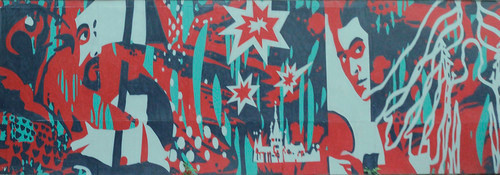} &
\includegraphics[height=1.65cm,width=1.5cm]{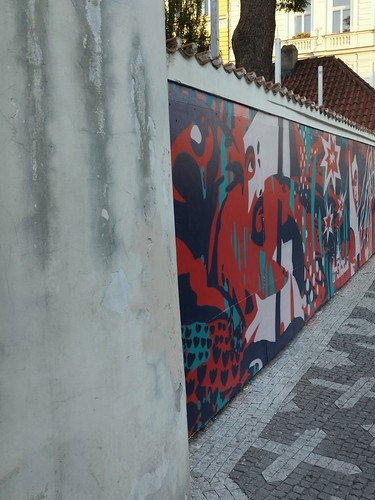} &
\includegraphics[height=1.65cm,width=1.5cm]{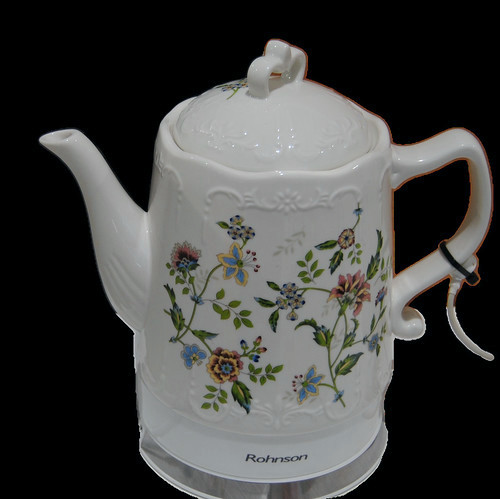} &
\includegraphics[height=1.65cm,width=1.5cm]{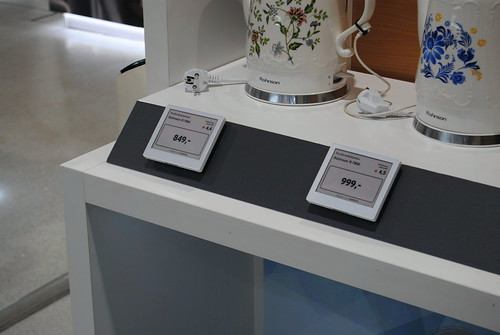} &
\includegraphics[height=1.65cm,width=1.5cm]{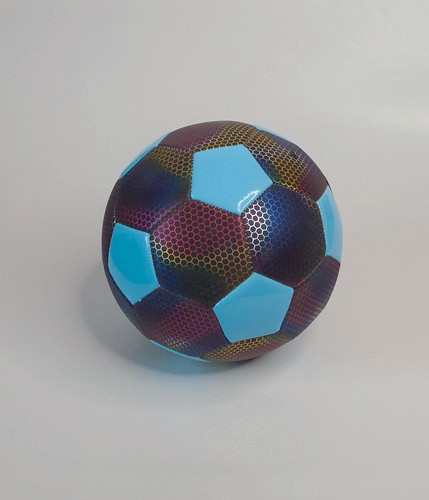} &
\includegraphics[height=1.65cm,width=1.5cm]{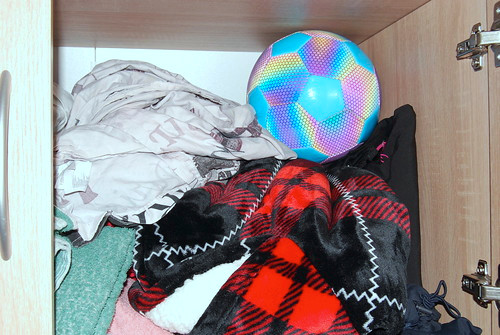} &
\includegraphics[height=1.65cm,width=1.5cm]{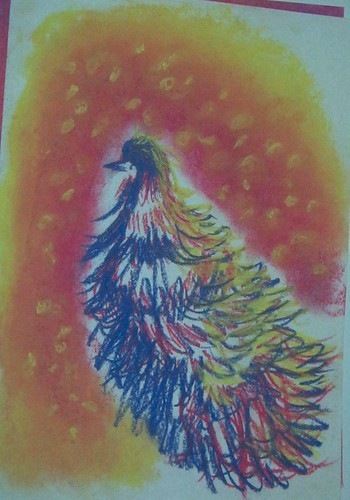} &
\includegraphics[height=1.65cm,width=1.5cm]{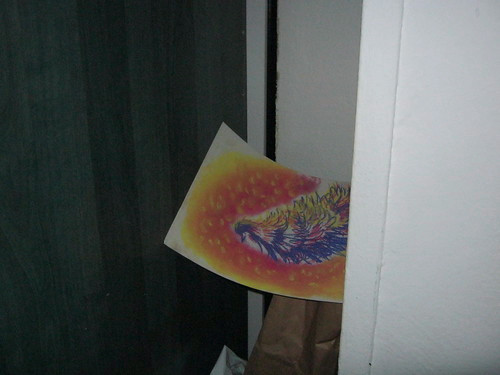} \\[3pt]
\multicolumn{2}{c}{$917 \rightarrow 3$} &
\multicolumn{2}{c}{$983 \rightarrow 0$} &
\multicolumn{2}{c}{$979 \rightarrow 4$} &
\multicolumn{2}{c}{$963 \rightarrow 0$} &
\multicolumn{2}{c}{$962 \rightarrow 6$} \\[3pt]
\includegraphics[height=1.65cm,width=1.5cm]{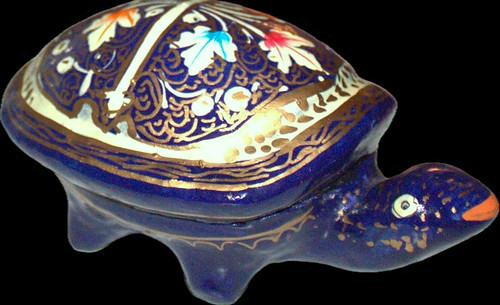} &
\includegraphics[height=1.65cm,width=1.5cm]{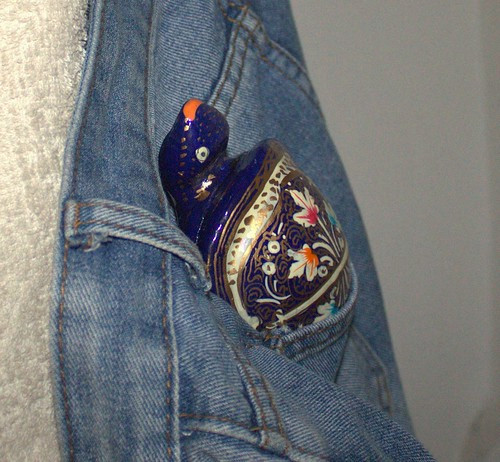} &
\includegraphics[height=1.65cm,width=1.5cm]{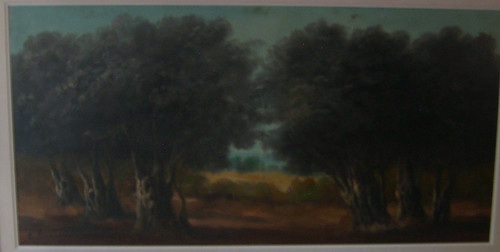} &
\includegraphics[height=1.65cm,width=1.5cm]{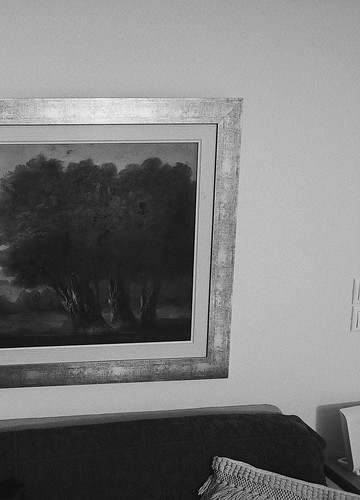} &
\includegraphics[height=1.65cm,width=1.5cm]{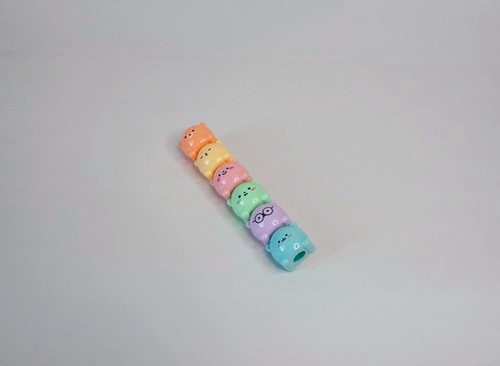} &
\includegraphics[height=1.65cm,width=1.5cm]{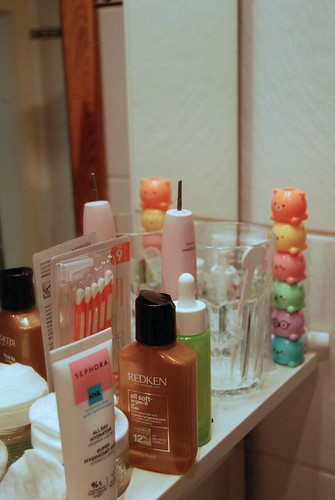} &
\includegraphics[height=1.65cm,width=1.5cm]{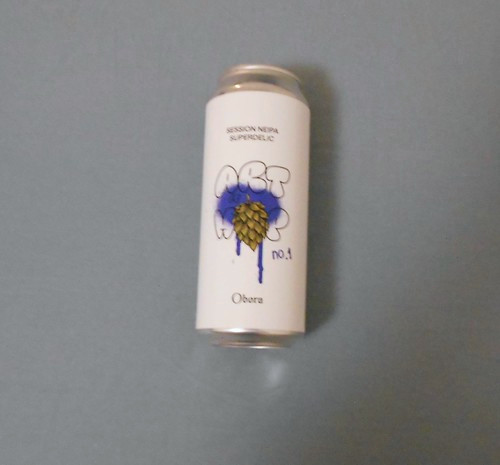} &
\includegraphics[height=1.65cm,width=1.5cm]{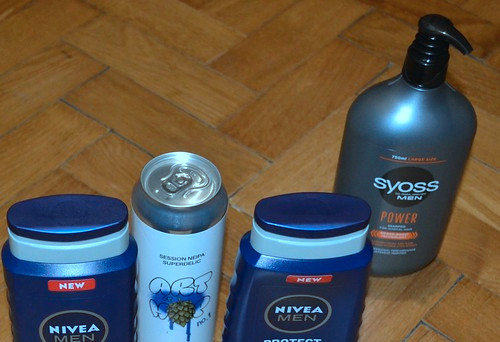} &
\includegraphics[height=1.65cm,width=1.5cm]{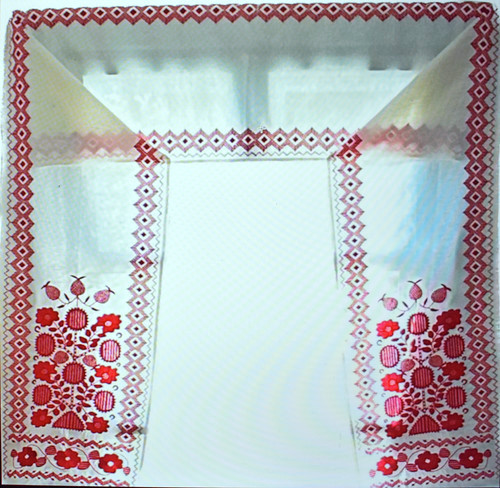} &
\includegraphics[height=1.65cm,width=1.5cm]{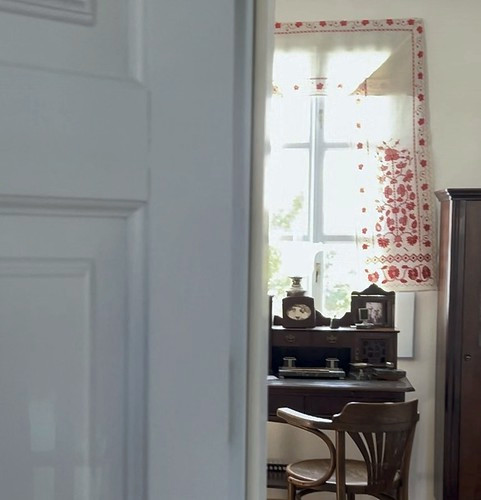} \\[3pt]

\multicolumn{10}{c}{\textbf{global \vs Qwen~~~---~~~global outperforms~~~---~~~Qwen $\rightarrow$ global}} \\[3pt]
\multicolumn{2}{c}{$743 \rightarrow 16$} &
\multicolumn{2}{c}{$808 \rightarrow 4$} &
\multicolumn{2}{c}{$802 \rightarrow 188$} &
\multicolumn{2}{c}{$310 \rightarrow 100$} &
\multicolumn{2}{c}{$218 \rightarrow 4$} \\[3pt]
\includegraphics[height=1.65cm,width=1.5cm]{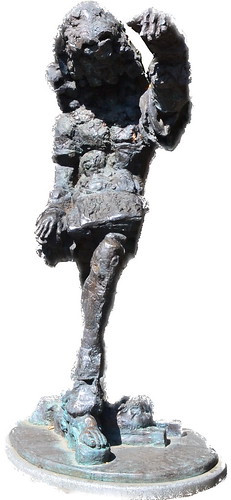} &
\includegraphics[height=1.65cm,width=1.5cm]{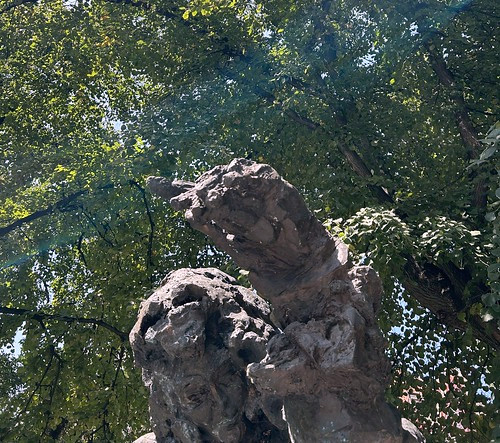} &
\includegraphics[height=1.65cm,width=1.5cm]{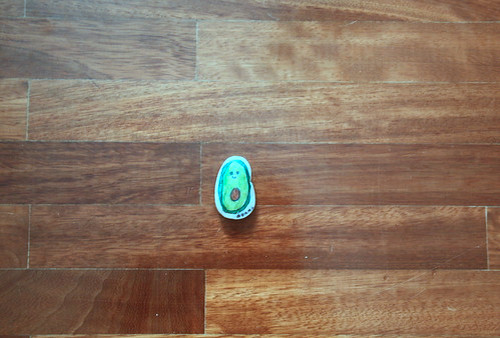} &
\includegraphics[height=1.65cm,width=1.5cm]{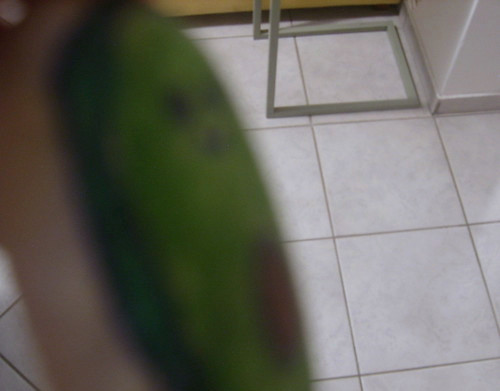} &
\includegraphics[height=1.65cm,width=1.5cm]{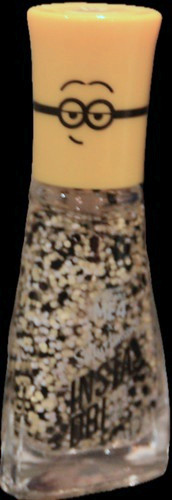} &
\includegraphics[height=1.65cm,width=1.5cm]{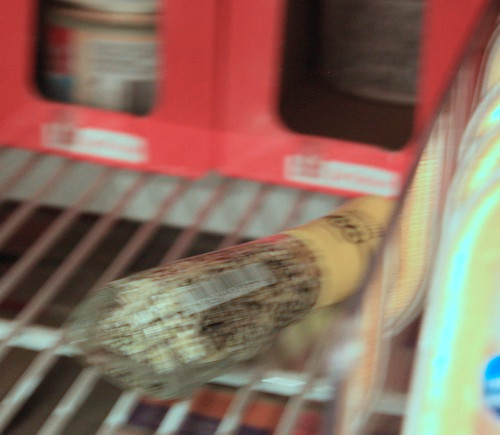} &
\includegraphics[height=1.65cm,width=1.5cm]{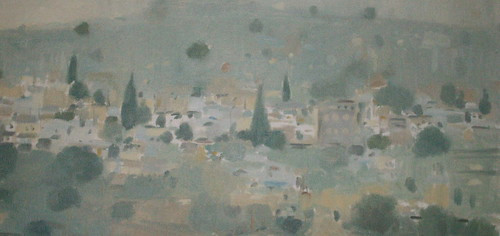} &
\includegraphics[height=1.65cm,width=1.5cm]{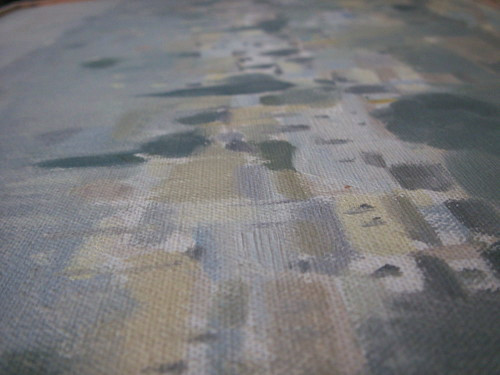} &
\includegraphics[height=1.65cm,width=1.5cm]{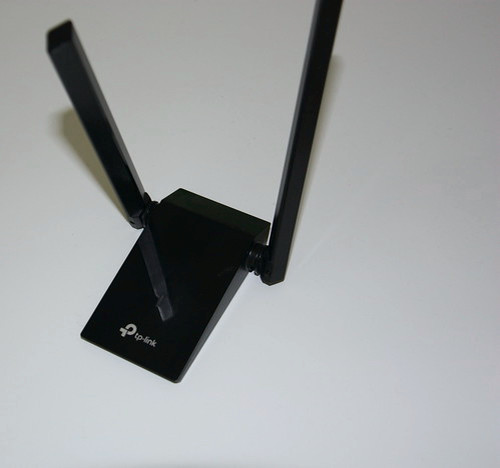} &
\includegraphics[height=1.65cm,width=1.5cm]{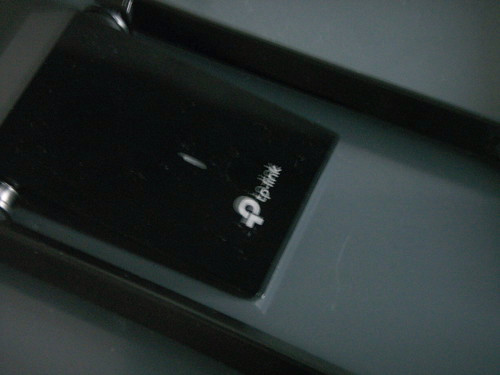} \\[3pt]
\multicolumn{2}{c}{$250 \rightarrow 0$} &
\multicolumn{2}{c}{$560 \rightarrow 43$} &
\multicolumn{2}{c}{$638 \rightarrow 107$} &
\multicolumn{2}{c}{$124 \rightarrow 7$} &
\multicolumn{2}{c}{$258 \rightarrow 0$} \\[3pt]
\includegraphics[height=1.65cm,width=1.5cm]{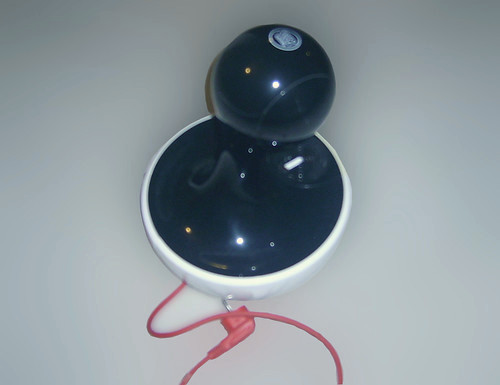} &
\includegraphics[height=1.65cm,width=1.5cm]{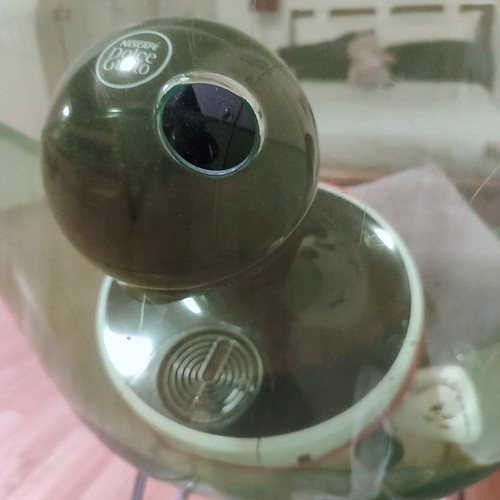} &
\includegraphics[height=1.65cm,width=1.5cm]{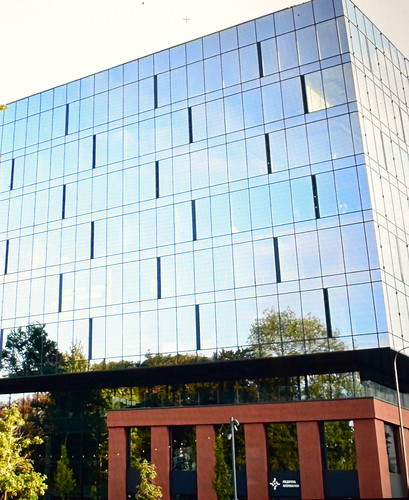} &
\includegraphics[height=1.65cm,width=1.5cm]{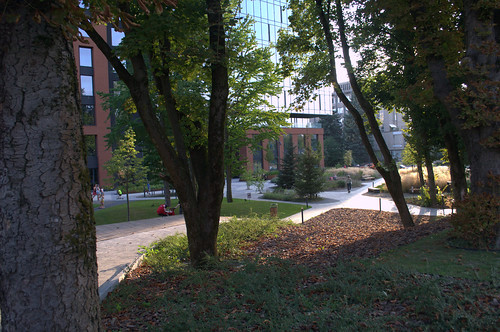} &
\includegraphics[height=1.65cm,width=1.5cm]{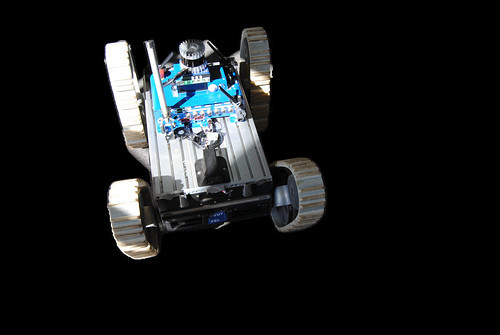} &
\includegraphics[height=1.65cm,width=1.5cm]{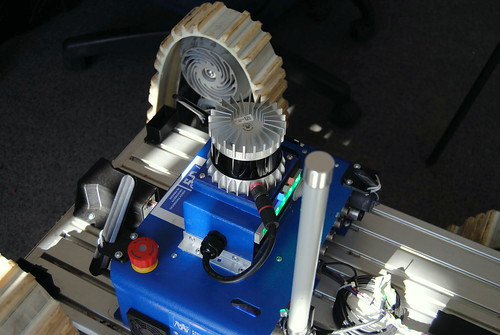} &
\includegraphics[height=1.65cm,width=1.5cm]{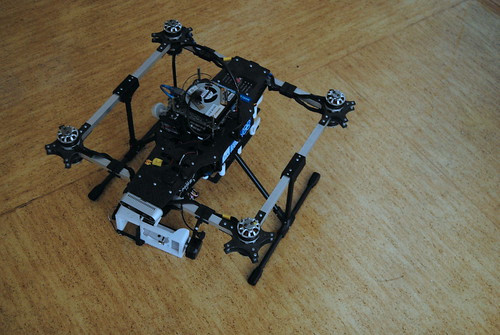} &
\includegraphics[height=1.65cm,width=1.5cm]{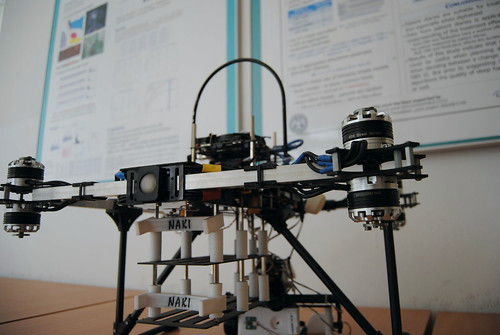} &
\includegraphics[height=1.65cm,width=1.5cm]{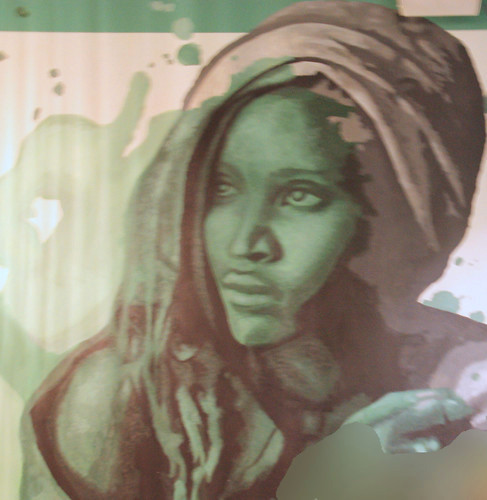} &
\includegraphics[height=1.65cm,width=1.5cm]{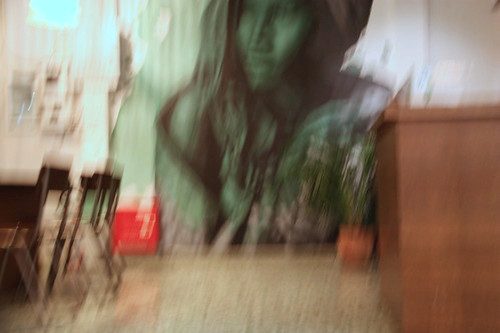} \\[3pt]

\multicolumn{10}{c}{\textbf{AMES \vs Qwen~~~---~~~Qwen outperforms~~~---~~~AMES $\rightarrow$ Qwen}} \\[3pt]
\multicolumn{2}{c}{$963 \rightarrow 0$} &
\multicolumn{2}{c}{$886 \rightarrow 0$} &
\multicolumn{2}{c}{$926 \rightarrow 0$} &
\multicolumn{2}{c}{$747 \rightarrow 1$} &
\multicolumn{2}{c}{$797 \rightarrow 0$} \\[3pt] 
\includegraphics[height=1.65cm,width=1.5cm]{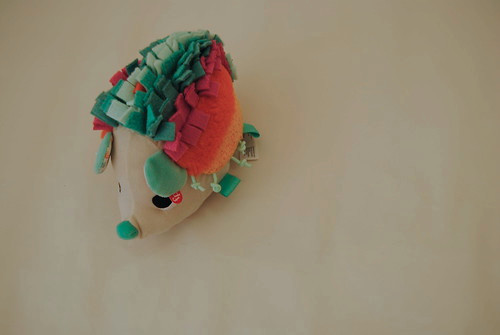} &
\includegraphics[height=1.65cm,width=1.5cm]{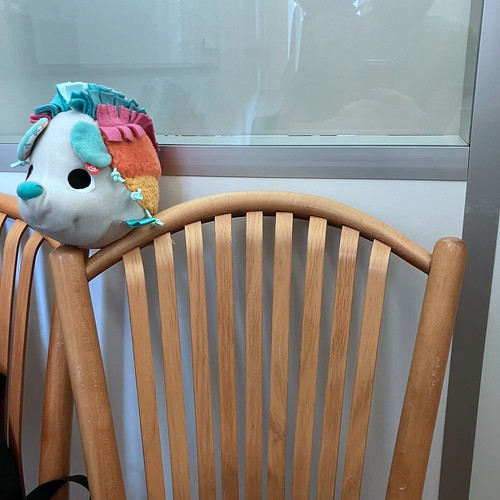} &
\includegraphics[height=1.65cm,width=1.5cm]{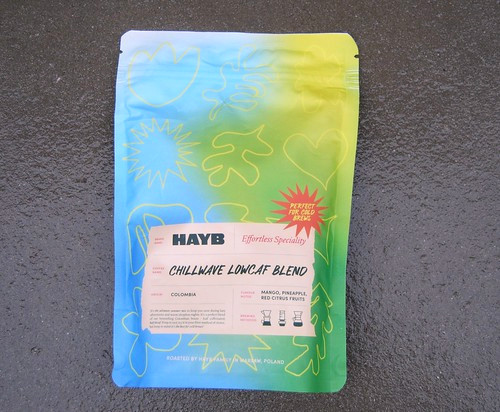} &
\includegraphics[height=1.65cm,width=1.5cm]{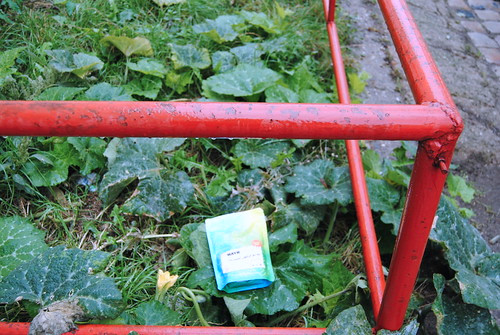} &
\includegraphics[height=1.65cm,width=1.5cm]{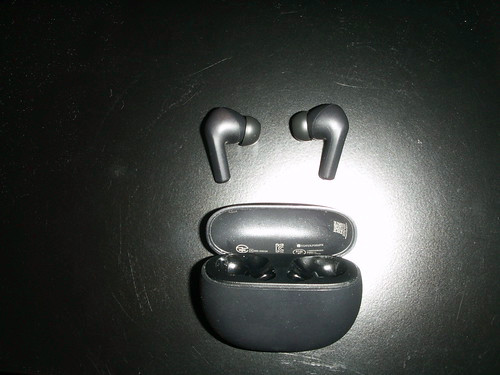} &
\includegraphics[height=1.65cm,width=1.5cm]{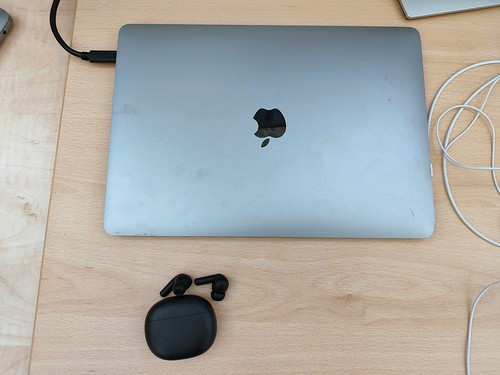} &
\includegraphics[height=1.65cm,width=1.5cm]{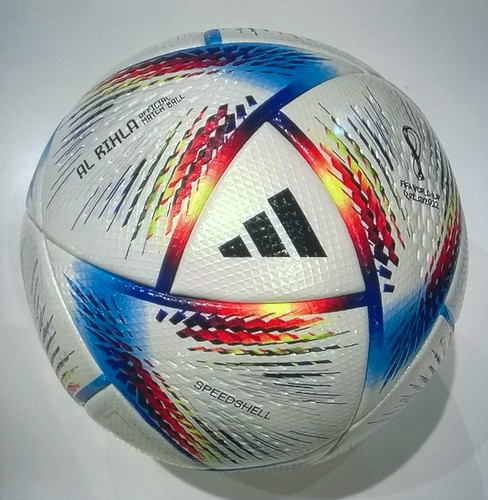} &
\includegraphics[height=1.65cm,width=1.5cm]{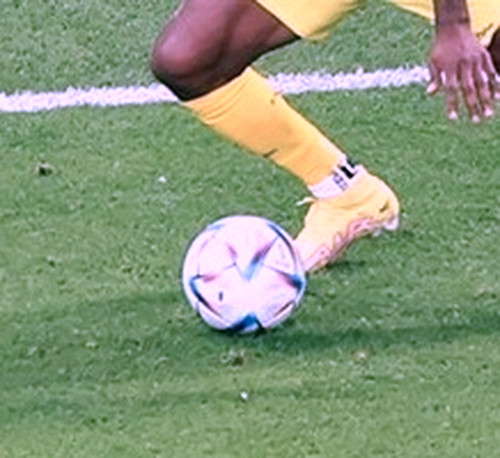} &
\includegraphics[height=1.65cm,width=1.5cm]{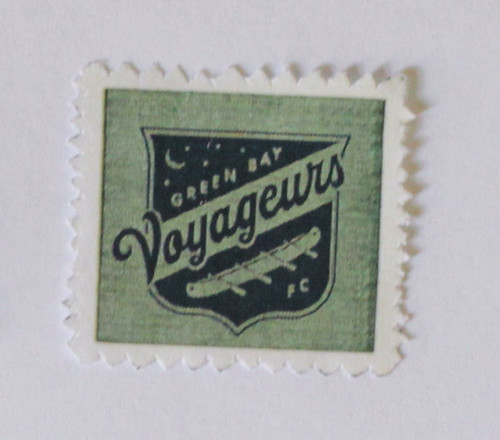} &
\includegraphics[height=1.65cm,width=1.5cm]{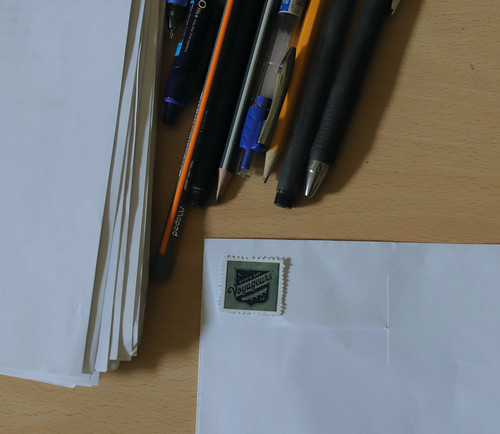} \\[3pt]
\multicolumn{2}{c}{$741 \rightarrow 4$} &
\multicolumn{2}{c}{$755 \rightarrow 0$} &
\multicolumn{2}{c}{$766 \rightarrow 1$} &
\multicolumn{2}{c}{$753 \rightarrow 25$} &
\multicolumn{2}{c}{$699 \rightarrow 0$} \\[3pt]
\includegraphics[height=1.65cm,width=1.5cm]{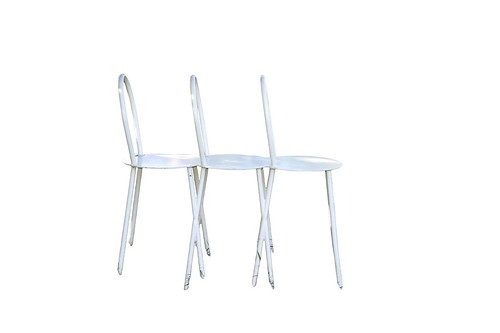} &
\includegraphics[height=1.65cm,width=1.5cm]{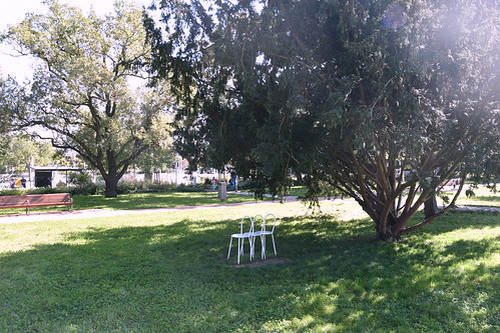} &
\includegraphics[height=1.65cm,width=1.5cm]{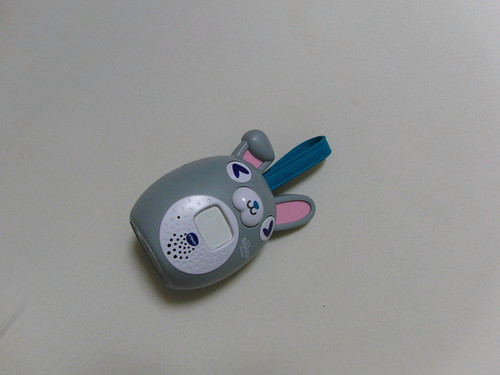} &
\includegraphics[height=1.65cm,width=1.5cm]{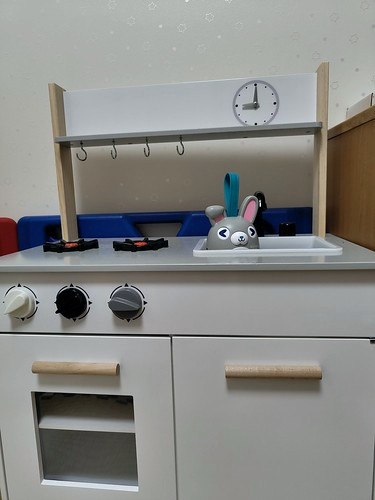} &
\includegraphics[height=1.65cm,width=1.5cm]{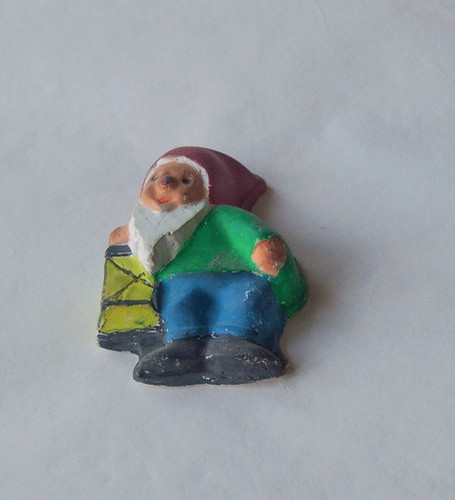} &
\includegraphics[height=1.65cm,width=1.5cm]{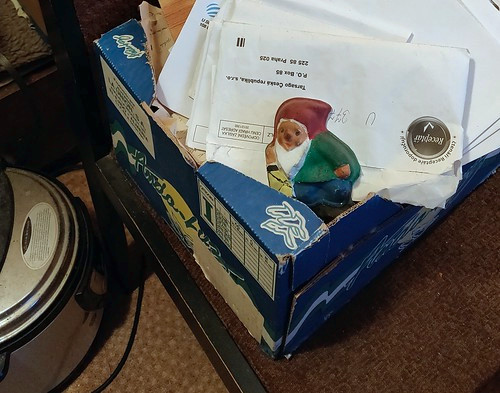} &
\includegraphics[height=1.65cm,width=1.5cm]{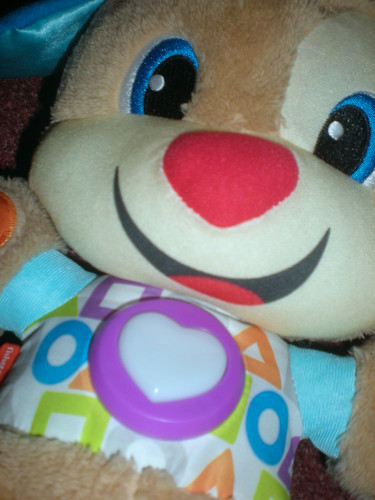} &
\includegraphics[height=1.65cm,width=1.5cm]{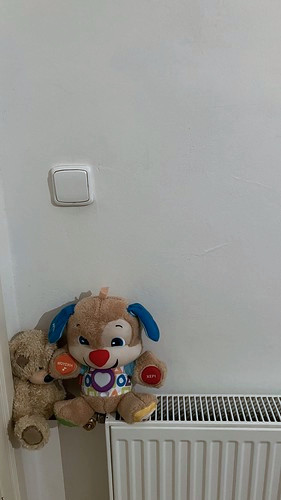} &
\includegraphics[height=1.65cm,width=1.5cm]{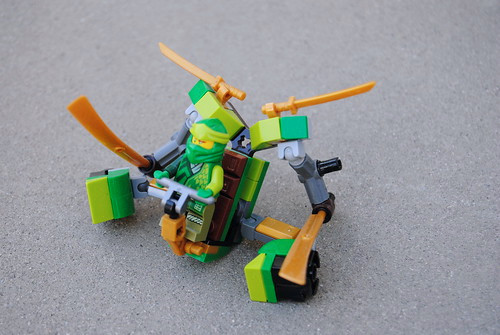} &
\includegraphics[height=1.65cm,width=1.5cm]{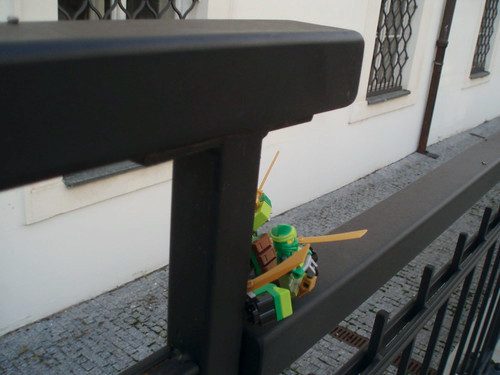} \\[3pt]

\multicolumn{10}{c}{\textbf{AMES \vs Qwen~~~---~~~AMES outperforms~~~---~~~Qwen $\rightarrow$ AMES}} \\[3pt]
\multicolumn{2}{c}{$484 \rightarrow 42$} &
\multicolumn{2}{c}{$24 \rightarrow 0$} &
\multicolumn{2}{c}{$297 \rightarrow 0$} &
\multicolumn{2}{c}{$456 \rightarrow 58$} &
\multicolumn{2}{c}{$227 \rightarrow 3$} \\[3pt]
\includegraphics[height=1.65cm,width=1.5cm]{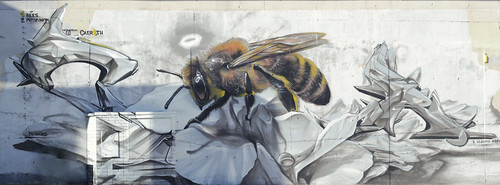} &
\includegraphics[height=1.65cm,width=1.5cm]{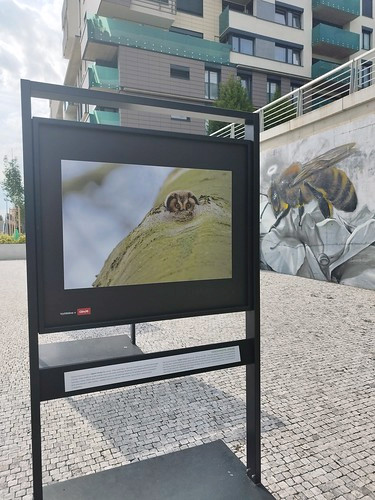} &
\includegraphics[height=1.65cm,width=1.5cm]{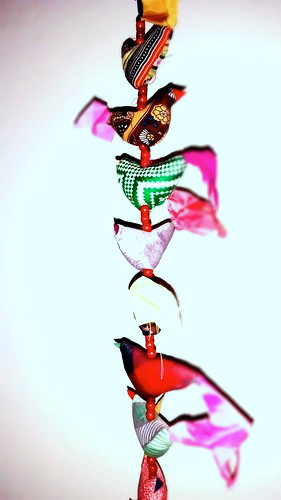} &
\includegraphics[height=1.65cm,width=1.5cm]{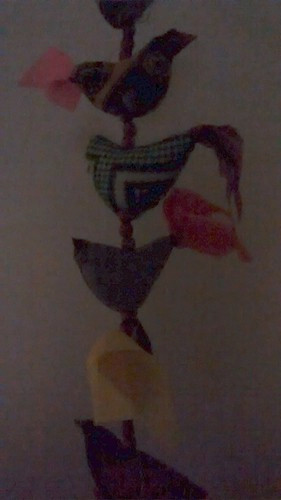} &
\includegraphics[height=1.65cm,width=1.5cm]{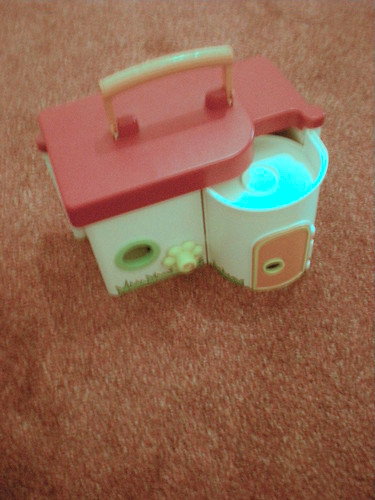} &
\includegraphics[height=1.65cm,width=1.5cm]{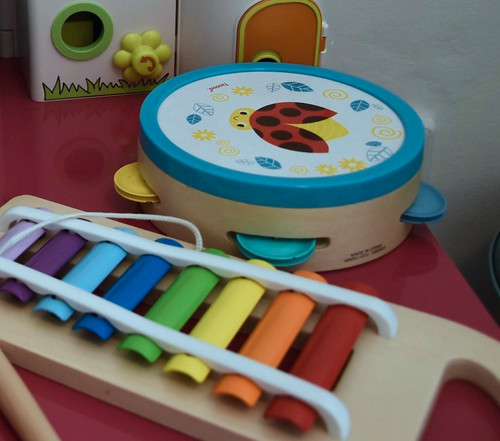} &
\includegraphics[height=1.65cm,width=1.5cm]{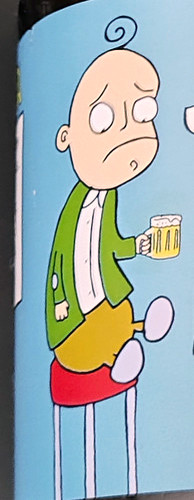} &
\includegraphics[height=1.65cm,width=1.5cm]{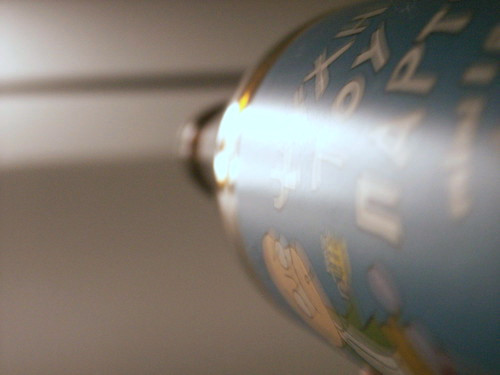} &
\includegraphics[height=1.65cm,width=1.5cm]{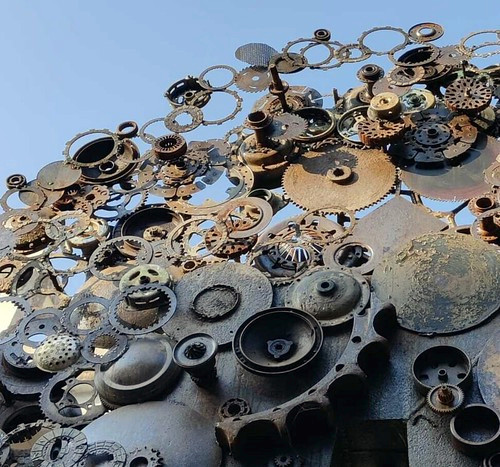} &
\includegraphics[height=1.65cm,width=1.5cm]{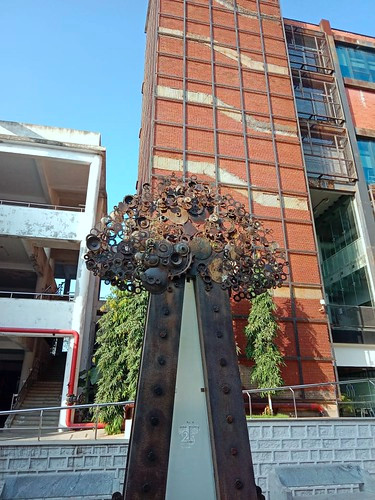} \\
\multicolumn{2}{c}{$453 \rightarrow 93$} &
\multicolumn{2}{c}{$287 \rightarrow 69$} &
\multicolumn{2}{c}{$210 \rightarrow 14$} &
\multicolumn{2}{c}{$434 \rightarrow 245$} &
\multicolumn{2}{c}{$633 \rightarrow 447$} \\[3pt]
\includegraphics[height=1.65cm,width=1.5cm]{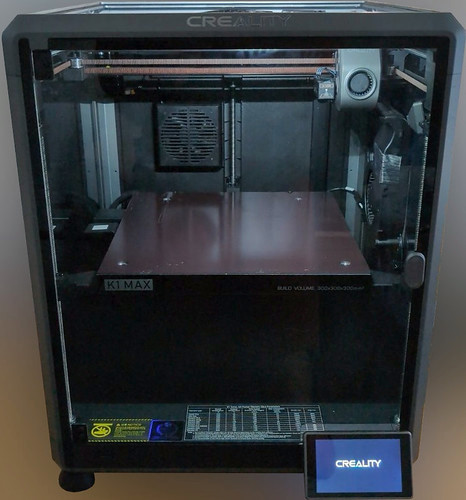} &
\includegraphics[height=1.65cm,width=1.5cm]{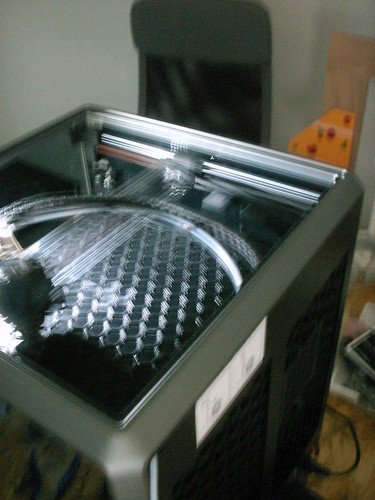} &
\includegraphics[height=1.65cm,width=1.5cm]{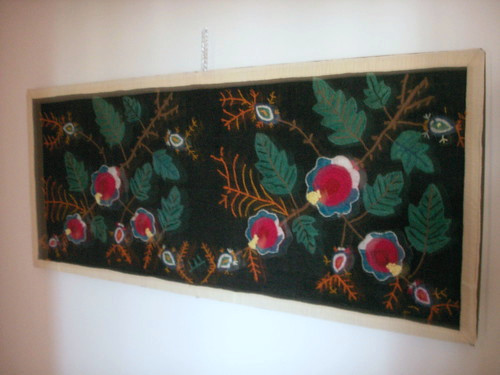} &
\includegraphics[height=1.65cm,width=1.5cm]{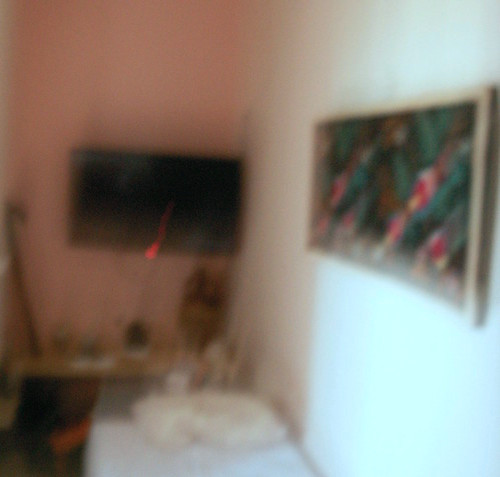} &
\includegraphics[height=1.65cm,width=1.5cm]{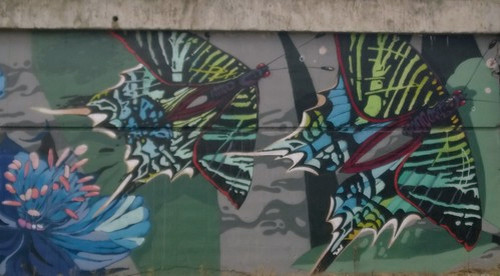} &
\includegraphics[height=1.65cm,width=1.5cm]{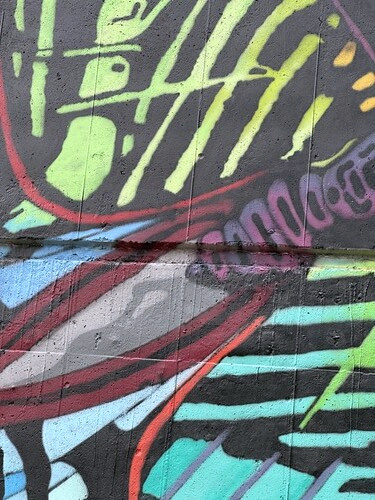} &
\includegraphics[height=1.65cm,width=1.5cm]{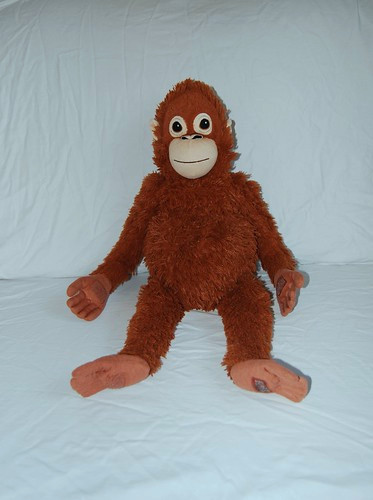} &
\includegraphics[height=1.65cm,width=1.5cm]{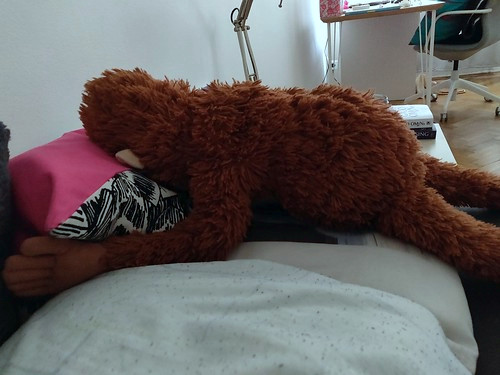} &
\includegraphics[height=1.65cm,width=1.5cm]{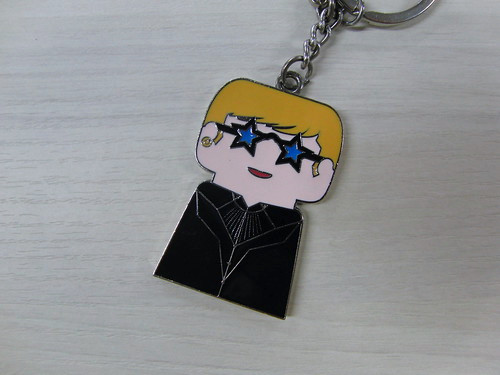} &
\includegraphics[height=1.65cm,width=1.5cm]{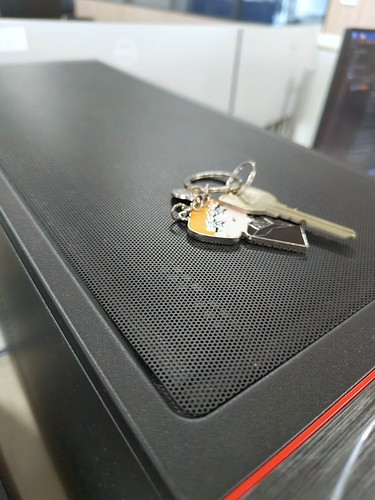} \\

\end{tabular}
    }
    \vspace{0pt}
    \caption{\textbf{More qualitative examples where one method benefits the most compared to another.} 
    We compare global (PE) and AMES \vs Qwen by showing pairs of query and positive image. $\rightarrow$ indicates the number of negative images ranked before the positive for two models, and it goes from the weaker to the stronger model for each pair.
    \label{fig:llm_vs_global_vs_ames_suppl}}
\end{figure*}

\myparagraph{Additional qualitative results.} In~\cref{fig:llm_vs_global_vs_ames_suppl}, we provide additional qualitative examples. We draw similar conclusions to the ones from the main paper.

\section{Dataset details}
\label{sec:dataset_details}
Below is the information regarding the datasets we used in our evaluation:

\textit{\textbf{ILIAS}}~\cite{ilias2025} is a large-scale, multi-domain dataset designed for instance-level image retrieval. 
It consists of 1,000 object instances spanning various domains, based on which 1,232 queries and a database of 4,715 positives have been collected and combined with 100 million distractors sourced from YFCC100M. 
We adopt ILIAS as the primary testbed in our evaluation due to its large scale, diversity, and challenging nature.

\textit{\textbf{INSTRE}}~\cite{wj15} is another instance-level multi-domain dataset. It consists of 200 objects and 1,250 single- and multi-object queries, and a database of 27.3k images.

\textit{\textbf{\rop}}~\cite{rit+18} is the combination of two instance-level datasets of the landmark domain, \ie \roxf~\cite{pci+07} and \rpar~\cite{pci+08} containing 70 queries each from 11 landmarks, and databases of 5k and 6k, respectively. They are extended with one million distractors. We measure performance based on mAP on the \textit{Medium} and \textit{Hard} settings.

\textit{\textbf{Product1M}}~\cite{zwd+21} is an instance-level dataset for product retrieval. It consists of 6.2k queries and 38.7k database images. In our evaluation, we randomly sample 1k queries.

\section{Prompts}
\label{sec:prompts}
\begin{figure*}[!ht]
    \centering
    \begin{tabular}{c}
        \begin{subfigure}{0.45\textwidth}
            \noindent
            \begingroup
            \setlength{\fboxsep}{8pt}%
            \ttfamily\footnotesize 
            \colorbox{gray!15}{%
                \parbox{0.9\linewidth}{%
                You are given two images: a query and a candidate.  Determine whether the candidate is similar to the query image.
                
                Output strictly a single digit: 
                \begin{itemize}
                    \item 0 = the object instance does not appear. 
                    \item 1 = the object instance appears in the candidate. 
                \end{itemize}
                Do not output anything else.
                }}
            \endgroup
            \caption{\texttt{generic} prompt}
        \end{subfigure}
            
        \\
        
        \begin{subfigure}{0.45\textwidth}
            \noindent
            \begingroup
            \setlength{\fboxsep}{8pt}%
            \ttfamily\footnotesize 
            \colorbox{gray!15}{%
                \parbox{0.9\linewidth}{%
                You are given two images: a query and a candidate. Determine whether the exact same object instance from the query image is present in the candidate image. 
                \begin{itemize}
                    \item The instance must be the same, not just a similar object. 
                    \item The instance may appear at a different scale, partially occluded, or among other objects. 
                \end{itemize}
                
                Output strictly a single digit: 
                \begin{itemize}
                    \item 0 = the object instance does not appear. 
                    \item 1 = the object instance appears in the candidate. 
                \end{itemize}
                Do not output anything else.
                }}
            \endgroup
            \caption{\texttt{object} prompt}
        \end{subfigure}
\end{tabular}
\begin{tabular}{c}
    \begin{subfigure}{0.45\textwidth}
        \noindent
        \begingroup
        \setlength{\fboxsep}{8pt}%
        \ttfamily\footnotesize 
        \colorbox{gray!15}{%
            \parbox{0.9\linewidth}{%
            You are given two images: a query and a candidate. Determine whether the exact same landmark, building, or architectural detail from the query image is present in the candidate image.
            \begin{itemize}
                \item The instance must be the same, not just a similar-looking building or structure.
                \item The query image may show the entire landmark or just a specific, cropped part of it (like a doorway, statue, or window).
                \item The instance in the candidate image may appear at a different scale, from a different viewpoint/angle, under different lighting, or be partially occluded.
            \end{itemize}
            Output strictly a single digit: 
            \begin{itemize}
                \item 0 = the object instance does not appear. 
                \item 1 = the object instance appears in the candidate. 
            \end{itemize}
            Do not output anything else.
            }}
        \endgroup
        \caption{\texttt{landmark} prompt}
    \end{subfigure}
\end{tabular}

    \vspace{-7pt}
    \caption{\textbf{Prompts for similarity estimation.} We use one generic and three task-specific prompts to evaluate the benchmarked MLLMs. We use \texttt{object} prompt for ILIAS, INSTRE, and Product1M, and the \texttt{landmark} prompt for \rop.
    \vspace{-7pt}
    \label{fig:prompts}
    }
\end{figure*}

We provide a set of prompts that are tailored to the nature of the task and the domain of the dataset. All prompts are illustrated in \cref{fig:prompts}. Specifically, we include: (i) A generic prompt, applicable to arbitrary images and object types, and represent the generic case. Variants of such prompts are typically employed for generic retrieval tasks~\cite{lamra}. (ii) An object prompt, suited for datasets containing diverse objects. We use this prompt for ILIAS, INSTRE, and Product1M. Regarding the latter, we also try prompts tailored for the product domain, but with no or insignificant improvements. (iii) A landmark prompt, designed for scenes involving buildings, monuments, or architectural elements, used in \rop.

We also provide an analysis of how MLLMs are sensitive to prompting. \cref{tab:prompt_sensitivity} shows that prompt sensitivity matters primarily at the semantic level. Task-specific prompts consistently perform best for the corresponding dataset, while unrelated domain-specific prompts perform poorly. At the same time, once the prompt is aligned with the task, its exact wording has only a limited effect: the task-specific prompt achieves 54.4 mAP, while five paraphrased variants remain close with a variance of \(\sigma^2=0.585\), indicating that the MLLM is largely insensitive to prompt phrasing.
\cref{tab:qwen3R-prompting} further shows that applying our task-specific prompt to Qwen3R consistently outperforms its default generic prompt, which comes with the Qwen3-Reranker model family, demonstrating the benefit of explicit task-aware prompting even for Qwen3R.

\begin{table}[t]
  \centering
  \hspace{0pt}
  \scalebox{1.}{
    \newcolumntype{C}{>{\centering\arraybackslash}p{3.0em}}

\begin{tabular}{lCCCC}
\toprule
\textbf{prompt} & \textbf{ILIAS} & \textbf{INSTRE} & \textbf{ROP} & \textbf{Prod1M} \\
\midrule
~~\texttt{generic}    & 42.0 & 94.6 & 63.0 & 72.3 \\
~~\texttt{object}     & \textbf{53.3} & \textbf{96.4} & 65.7 & \textbf{74.5} \\
~~\texttt{landmark}   & 40.0 & 90.7 & \textbf{68.1} & 66.9 \\
\bottomrule
\end{tabular}
  }
  \vspace{0pt}
  \caption{\textbf{Prompt Sensitivity.} Retrieval performance (mAP) across datasets for different prompt types.
  \label{tab:prompt_sensitivity}}
  \vspace{0pt}
\end{table}

\section{Transform details}
\begin{figure*}
    \centering
    \begin{tabular}{c@{\ssp}c@{\ssp}c@{\ssp}c@{\ssp}c@{\ssp}c@{\ssp}c@{\ssp}c@{\ssp}c@{\ssp}c@{\ssp}c@{\ssp}c@{\ssp}}
     \multicolumn{11}{l}{\textbf{contrast}} \\
     \includegraphics[width=1.4cm]{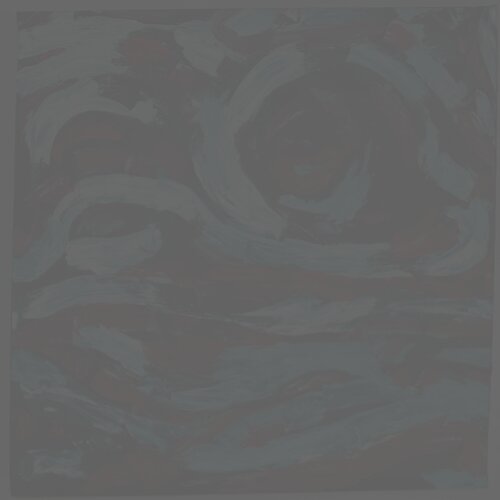} &
     \includegraphics[width=1.4cm]{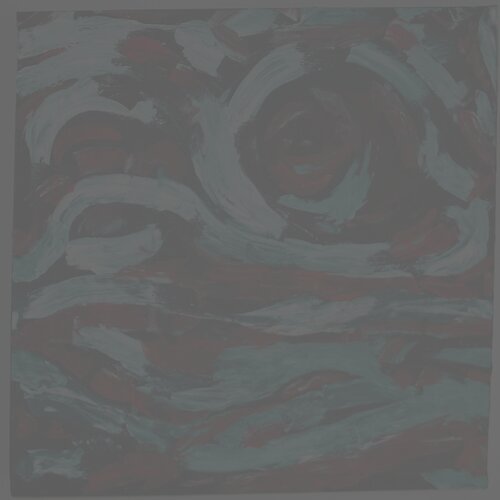} &
     \includegraphics[width=1.4cm]{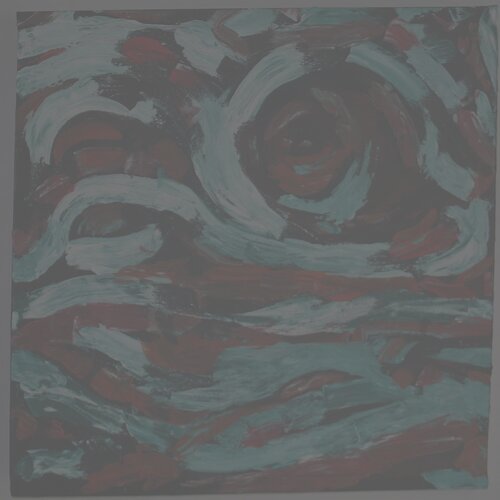} &
     \includegraphics[width=1.4cm]{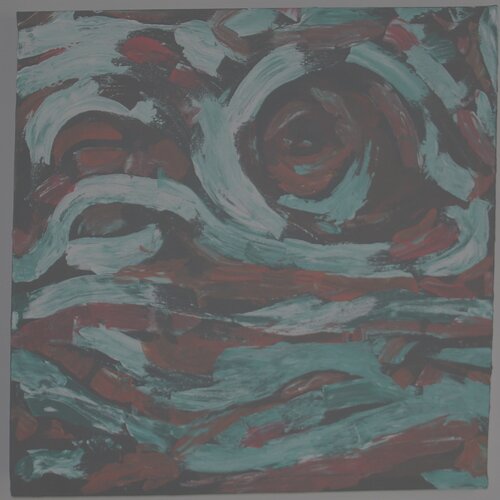} &
     \includegraphics[width=1.4cm]{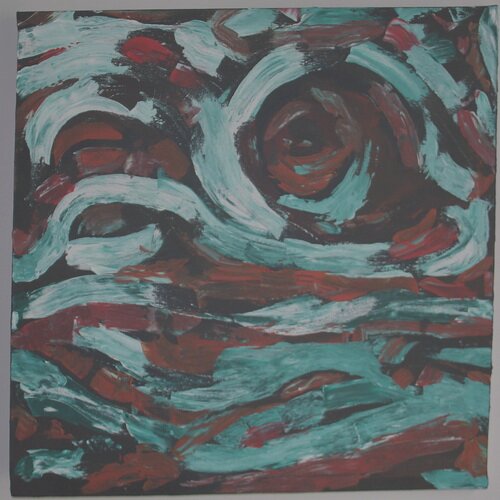} &
     \includegraphics[width=1.4cm]{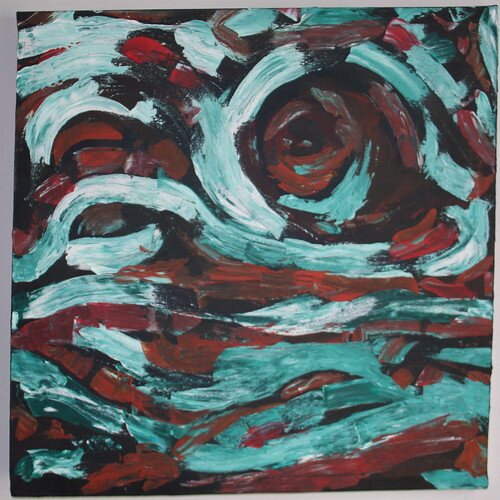} &
     \includegraphics[width=1.4cm]{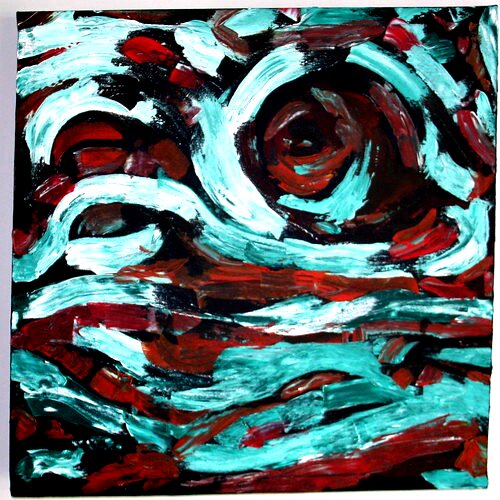} &
     \includegraphics[width=1.4cm]{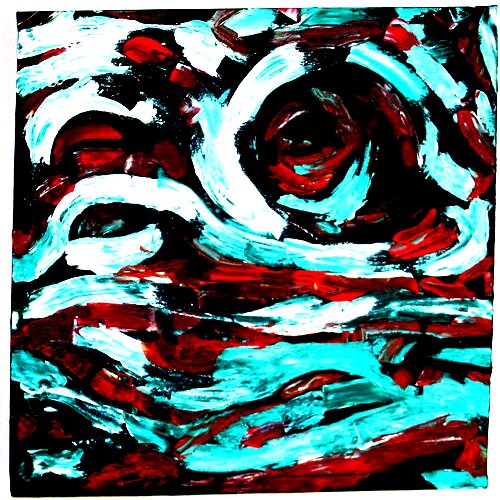} &
     \includegraphics[width=1.4cm]{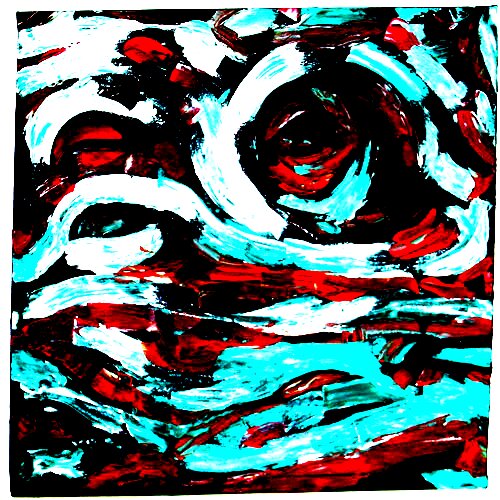} &
     \includegraphics[width=1.4cm]{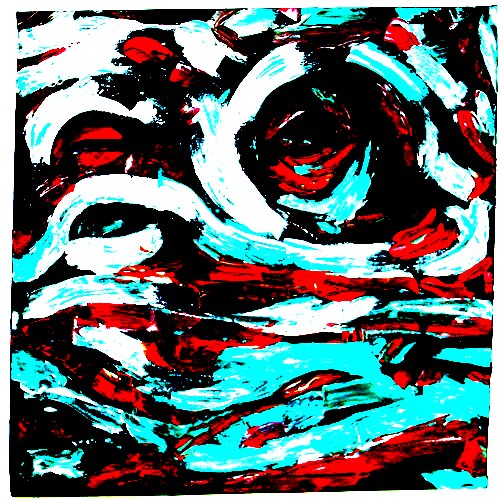} &
     \includegraphics[width=1.4cm]{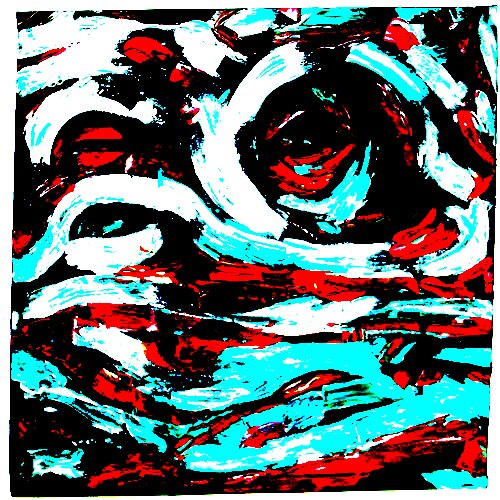}
     \vspace{3pt} \\
     \multicolumn{11}{l}{\textbf{brightness}} \\
     \includegraphics[width=1.4cm]{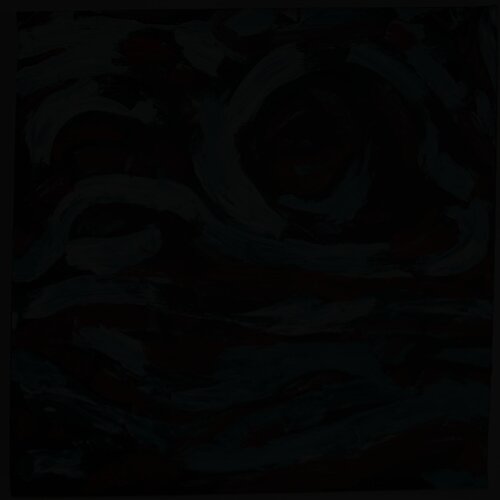} &
     \includegraphics[width=1.4cm]{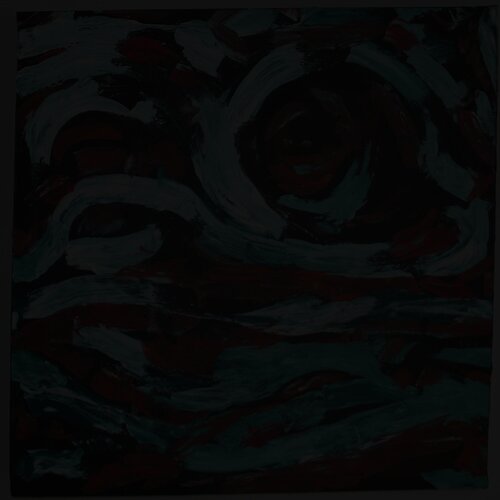} &
     \includegraphics[width=1.4cm]{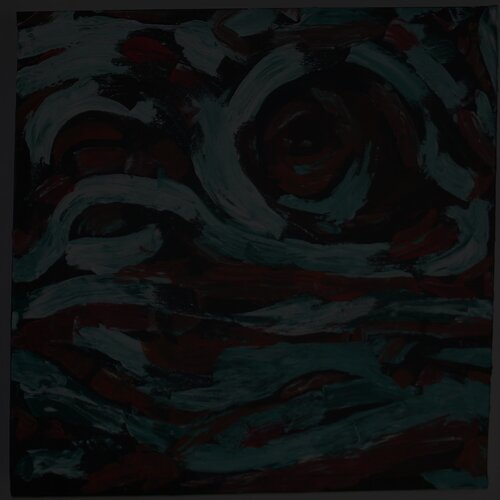} &
     \includegraphics[width=1.4cm]{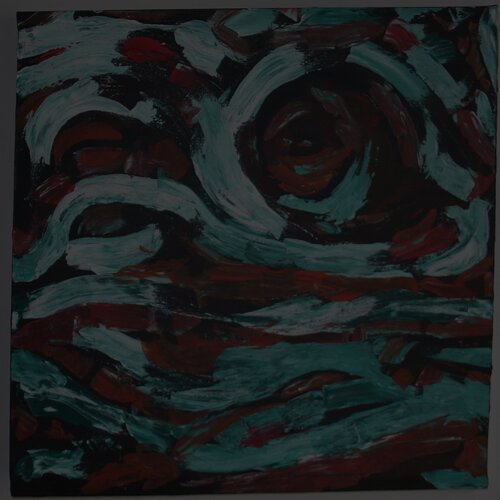} &
     \includegraphics[width=1.4cm]{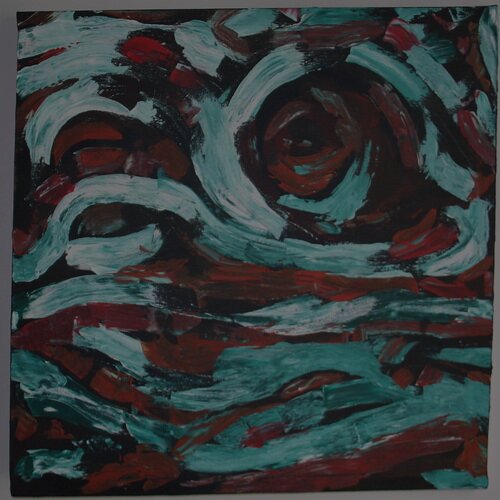} &
     \includegraphics[width=1.4cm]{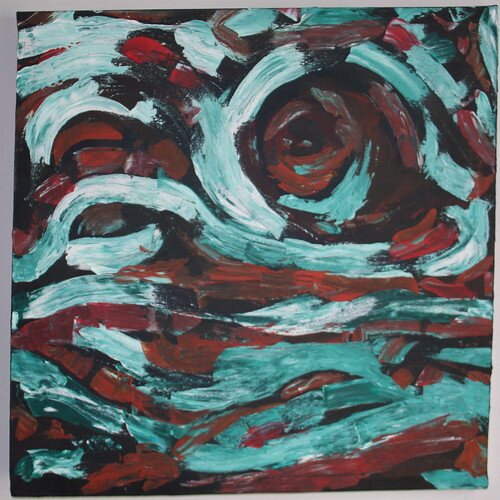} &
     \includegraphics[width=1.4cm]{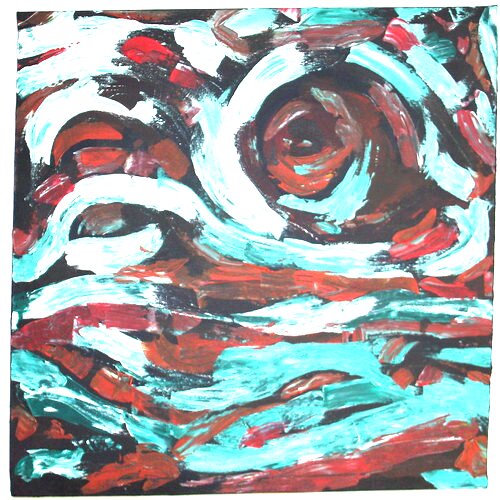} &
     \includegraphics[width=1.4cm]{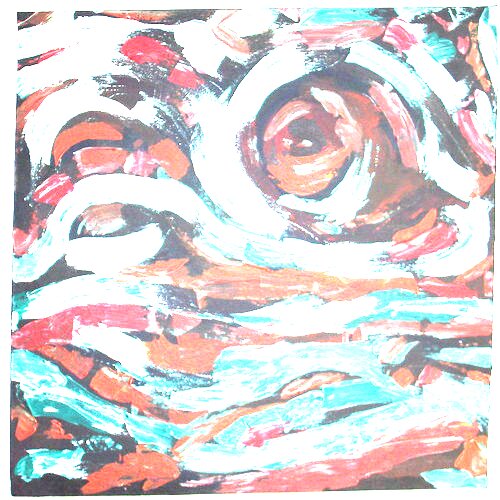} &
     \includegraphics[width=1.4cm]{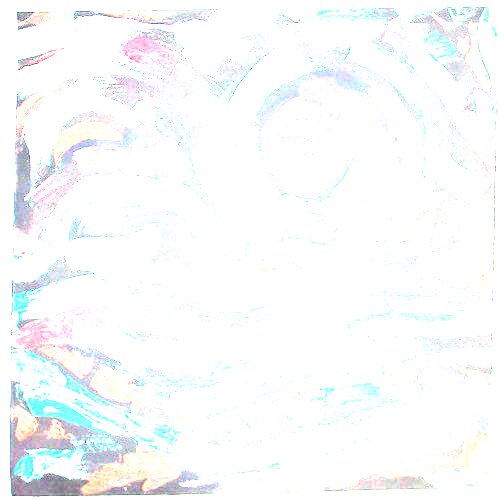} &
     \includegraphics[width=1.4cm]{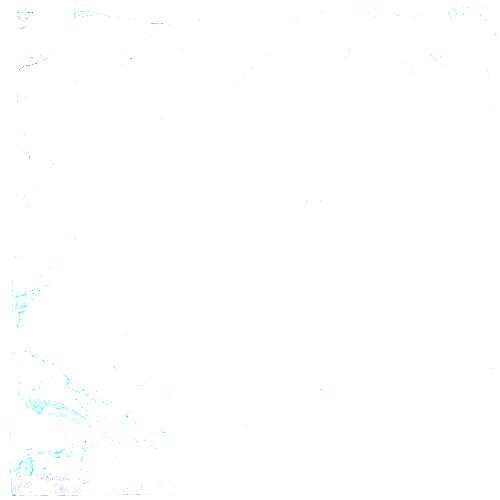} &
     \includegraphics[width=1.4cm]{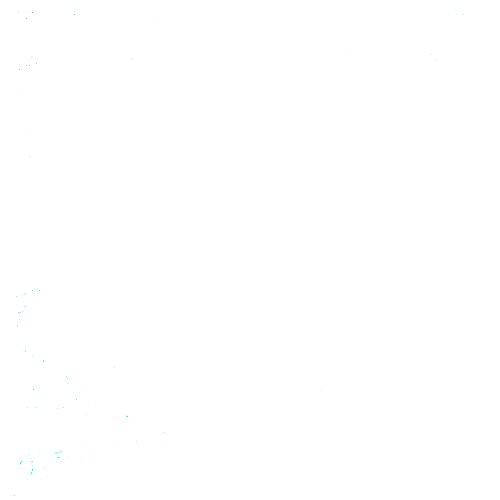}
     \vspace{3pt} \\
     \multicolumn{11}{l}{\textbf{rotation}} \\
     \includegraphics[width=1.4cm]{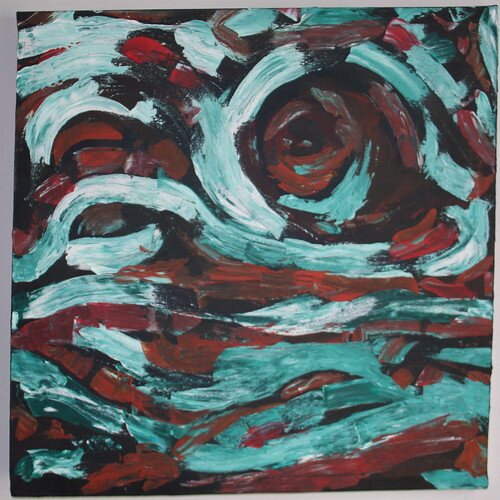} &
     \includegraphics[width=1.4cm]{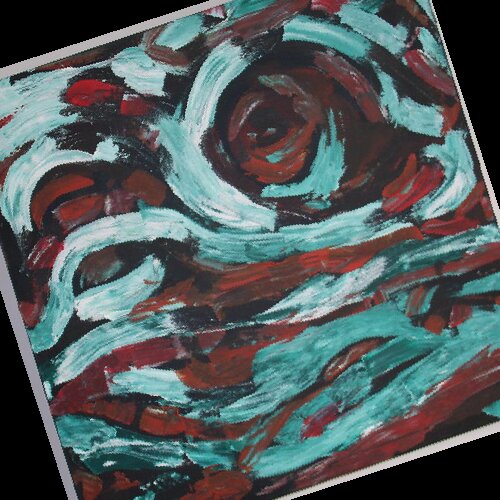} &
     \includegraphics[width=1.4cm]{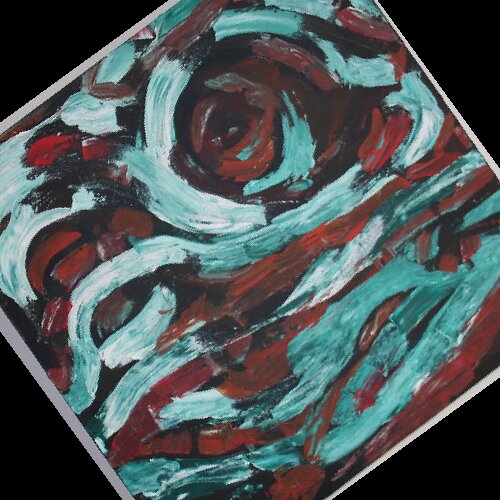} &
     \includegraphics[width=1.4cm]{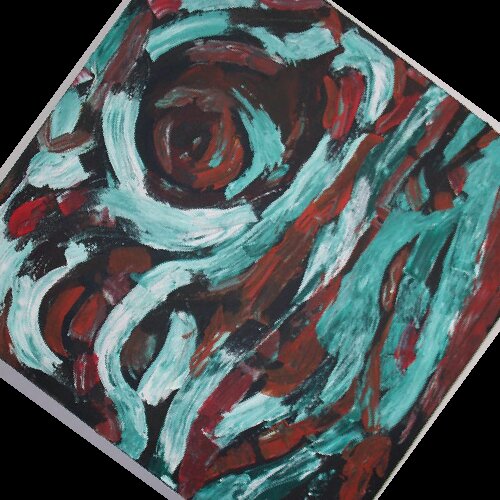} &
     \includegraphics[width=1.4cm]{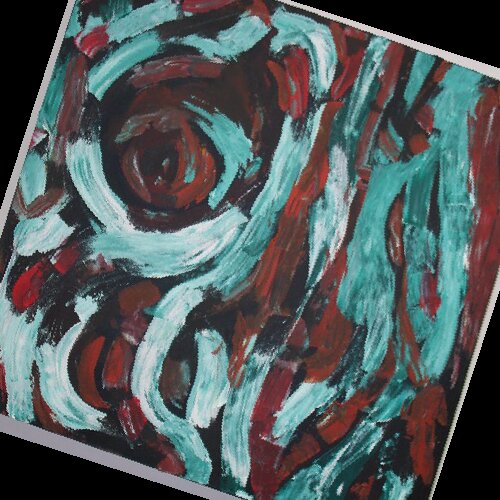} &
     \includegraphics[width=1.4cm]{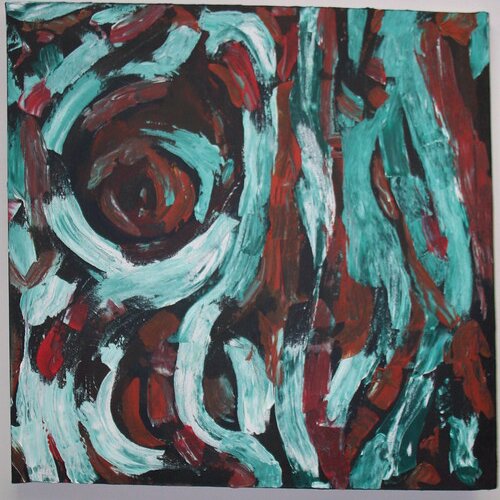} &
     \includegraphics[width=1.4cm]{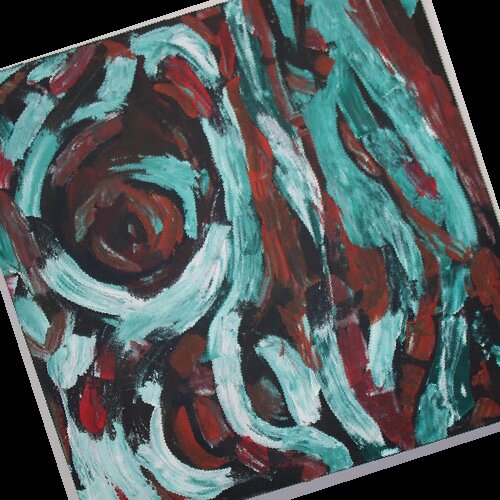} &
     \includegraphics[width=1.4cm]{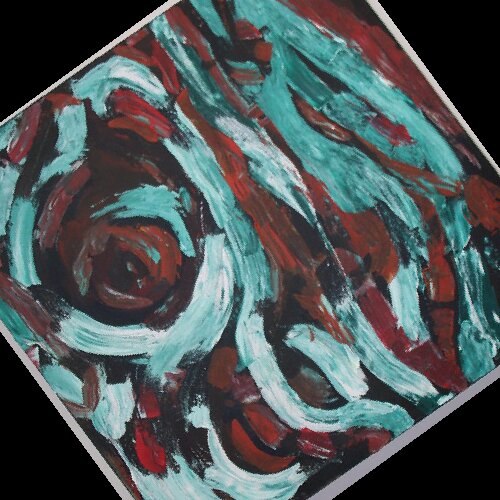} &
     \includegraphics[width=1.4cm]{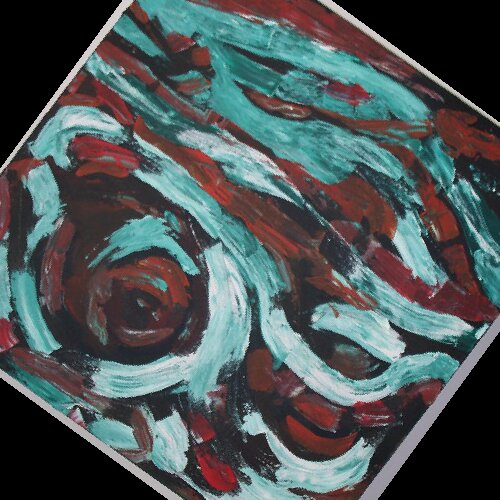} &
     \includegraphics[width=1.4cm]{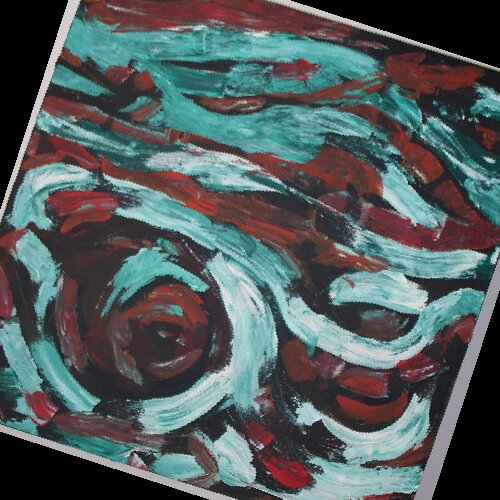} &
     \includegraphics[width=1.4cm]{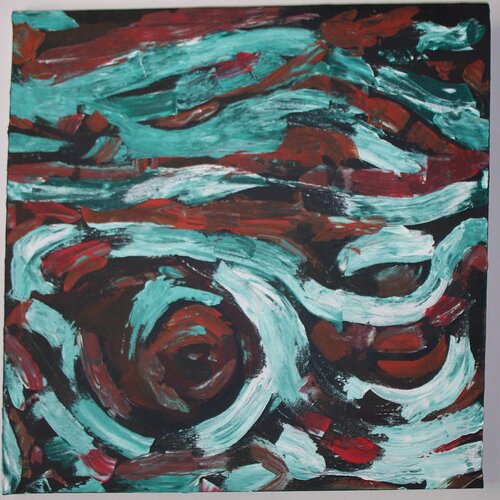}
     \vspace{3pt} \\
     \multicolumn{11}{l}{\textbf{downscale}} \\
     \includegraphics[width=1.4cm]{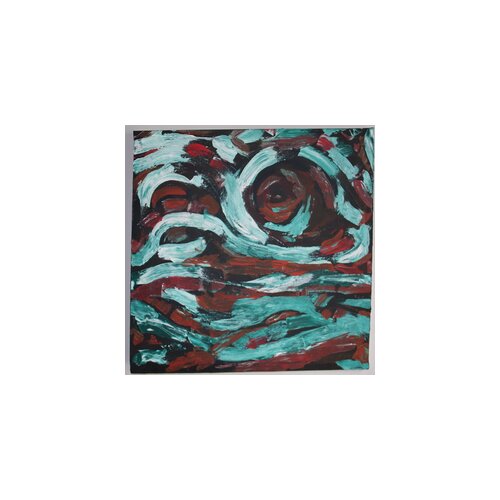} &
     \includegraphics[width=1.4cm]{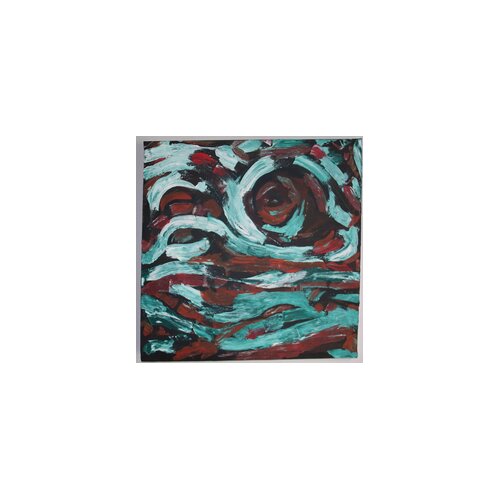} &
     \includegraphics[width=1.4cm]{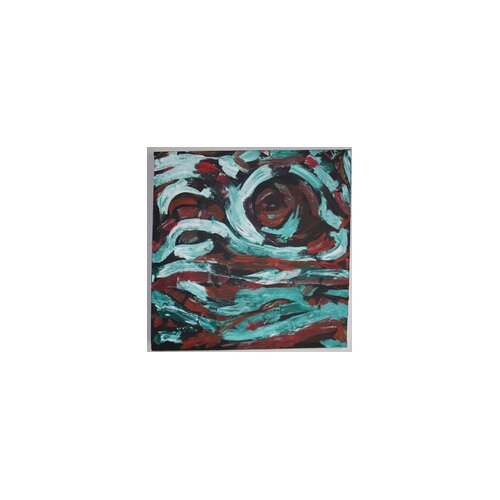} &
     \includegraphics[width=1.4cm]{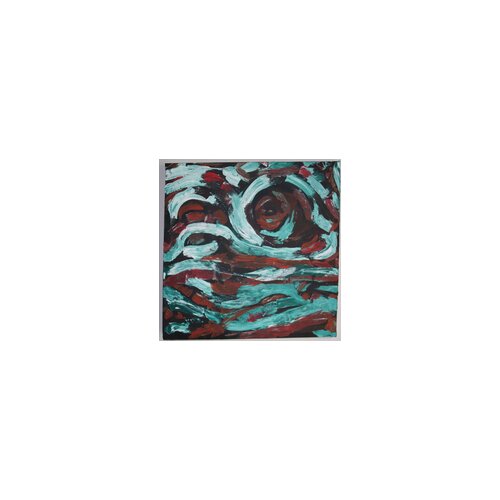} &
     \includegraphics[width=1.4cm]{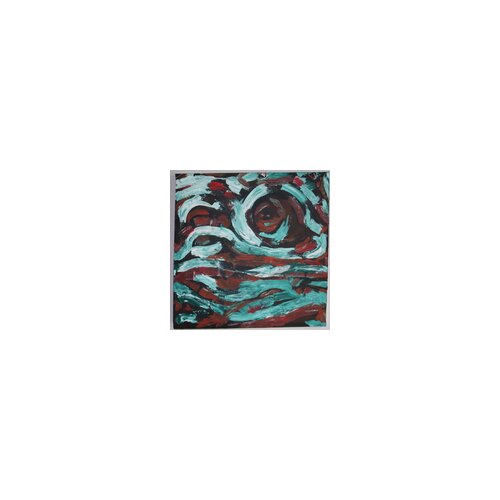} &
     \includegraphics[width=1.4cm]{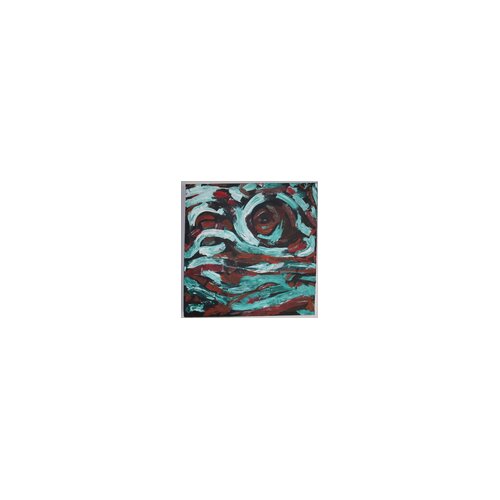} &
     \includegraphics[width=1.4cm]{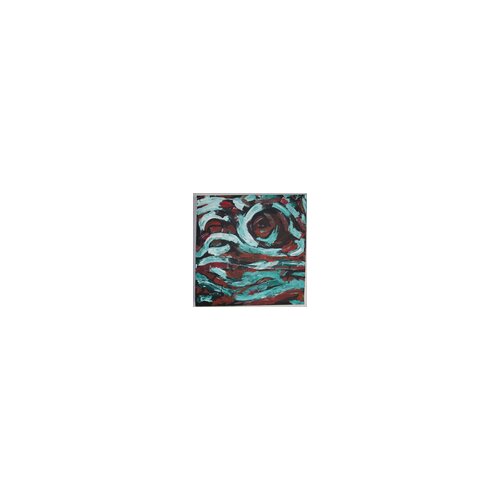} &
     \includegraphics[width=1.4cm]{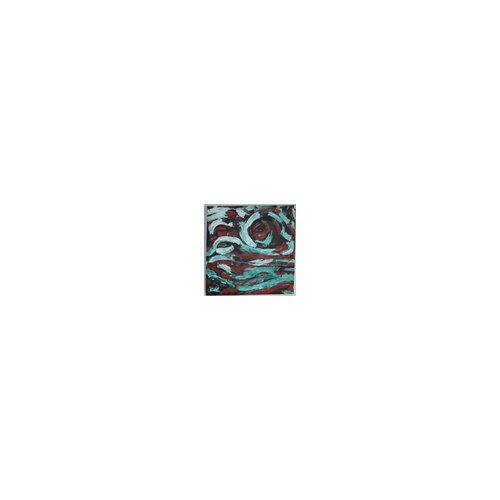} &
     \includegraphics[width=1.4cm]{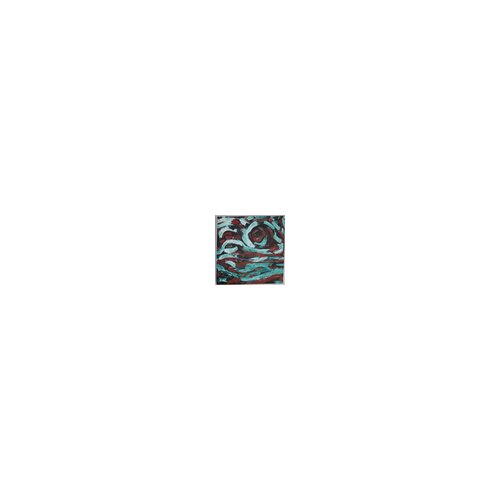} &
     \includegraphics[width=1.4cm]{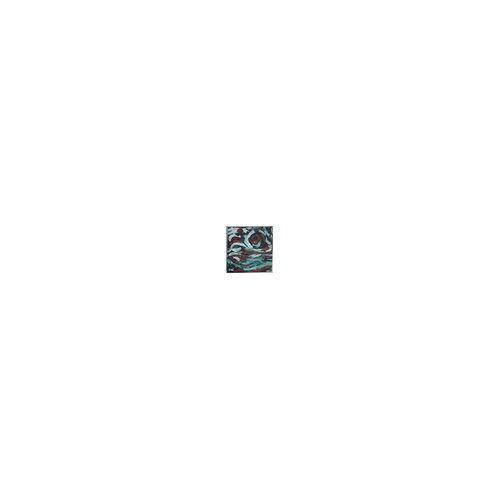} &
     \includegraphics[width=1.4cm]{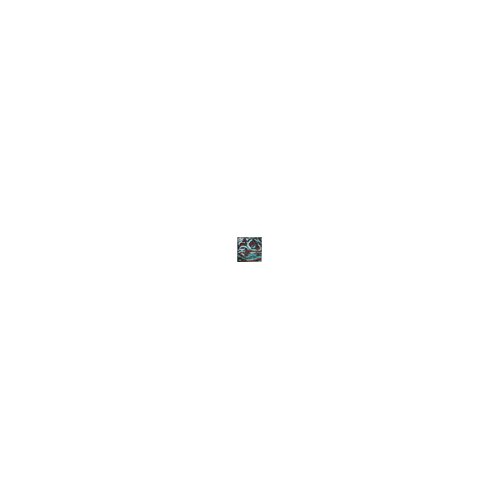}
     \vspace{3pt} \\
     \multicolumn{11}{l}{\textbf{scale-bg}} \\
     \includegraphics[width=1.4cm]{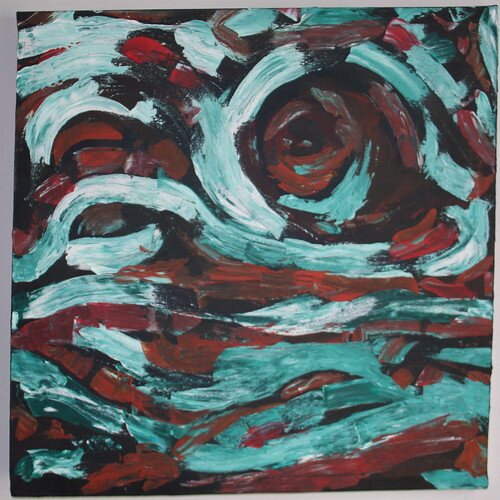} &
     \includegraphics[width=1.4cm]{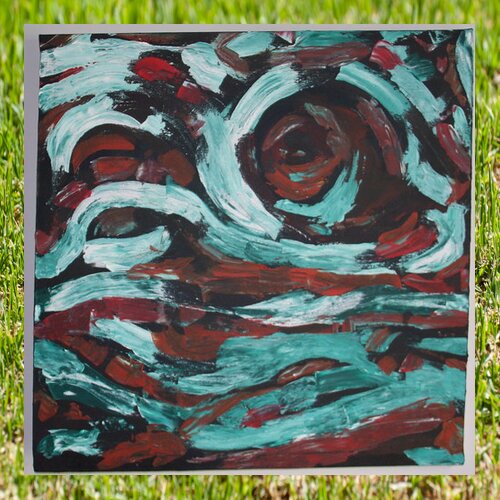} &
     \includegraphics[width=1.4cm]{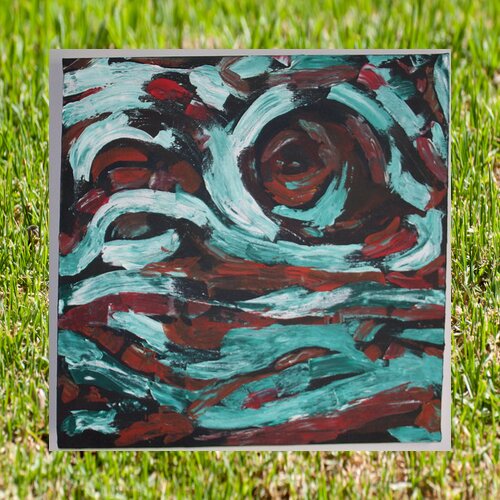} &
     \includegraphics[width=1.4cm]{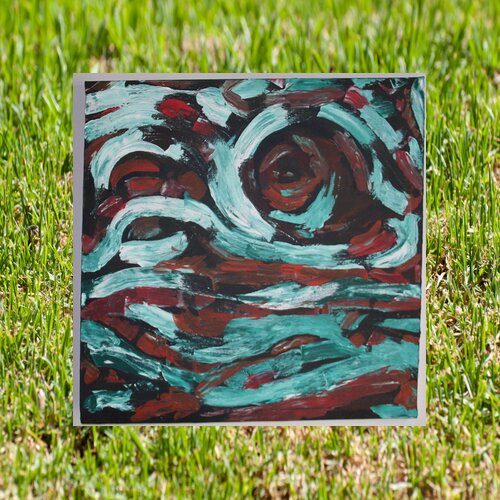} &
     \includegraphics[width=1.4cm]{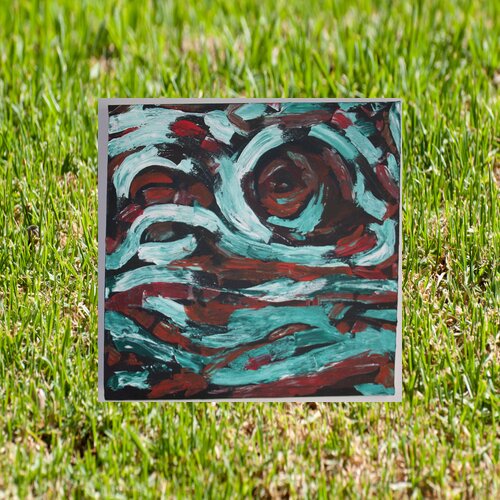} &
     \includegraphics[width=1.4cm]{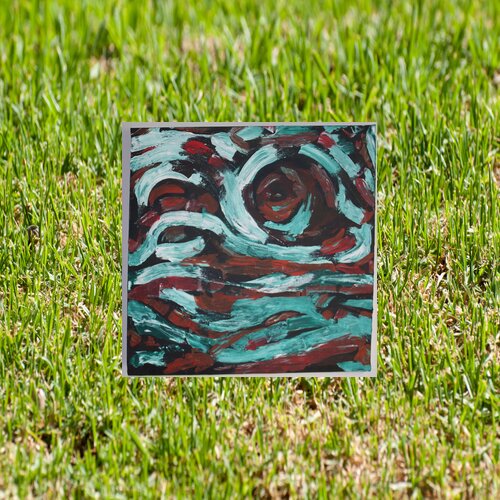} &
     \includegraphics[width=1.4cm]{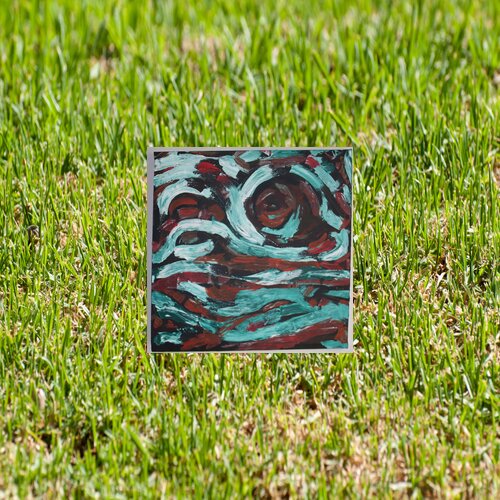} &
     \includegraphics[width=1.4cm]{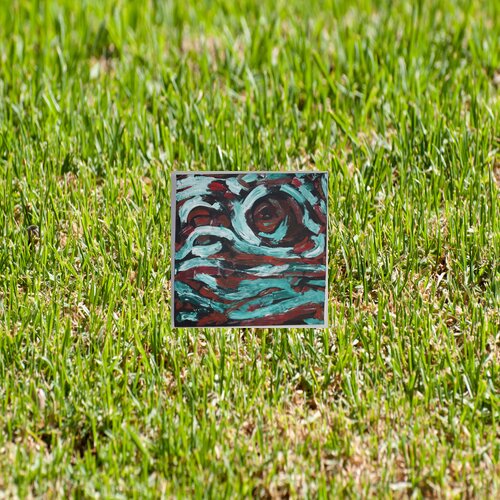} &
     \includegraphics[width=1.4cm]{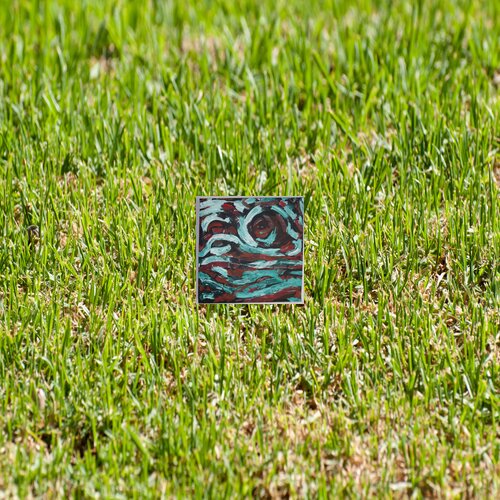} &
     \includegraphics[width=1.4cm]{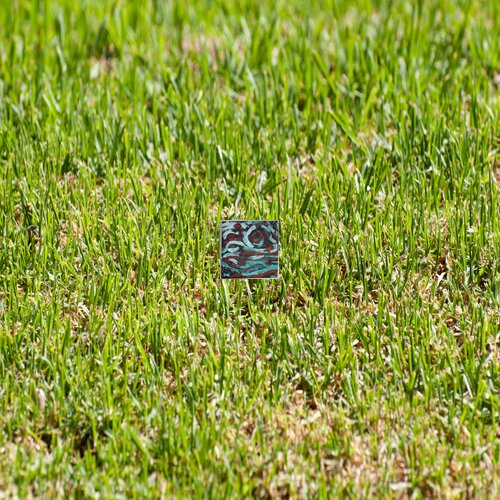} &
     \includegraphics[width=1.4cm]{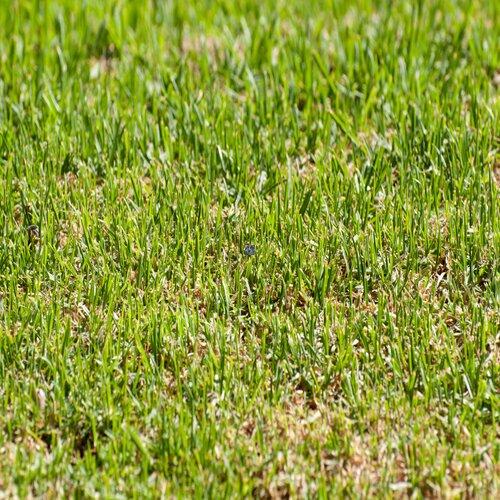}
     \vspace{3pt} \\
     \multicolumn{11}{l}{\textbf{blur}} \\
     \includegraphics[width=1.4cm]{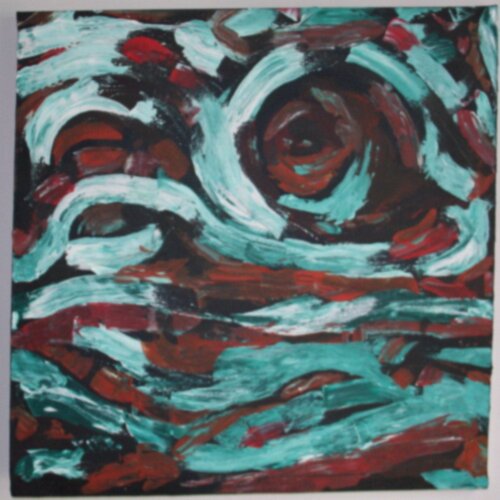} &
     \includegraphics[width=1.4cm]{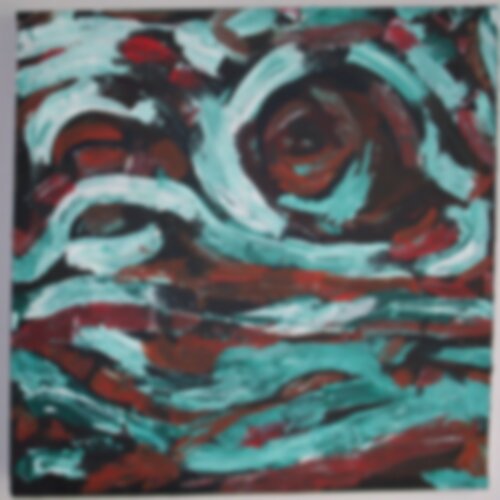} &
     \includegraphics[width=1.4cm]{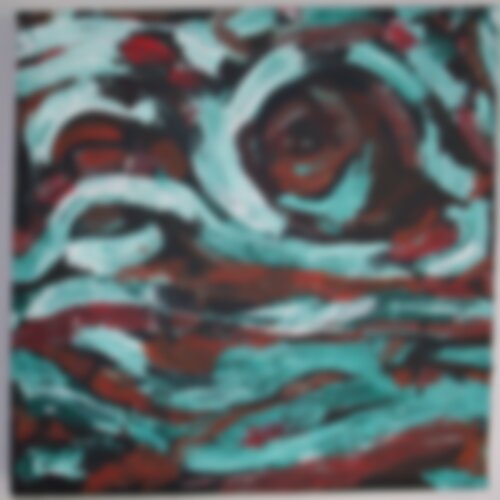} &
     \includegraphics[width=1.4cm]{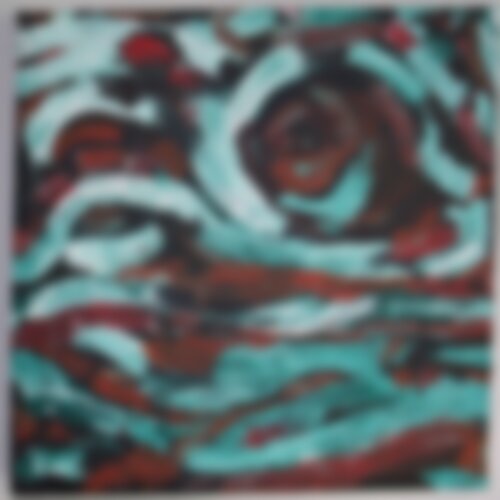} &
     \includegraphics[width=1.4cm]{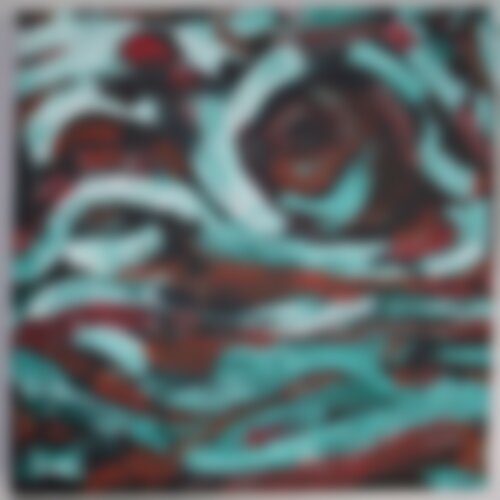} &
     \includegraphics[width=1.4cm]{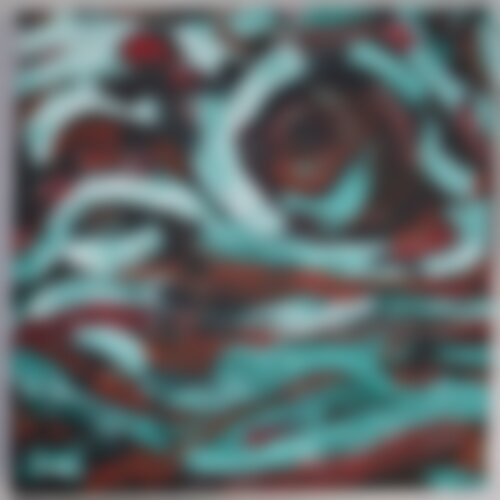} &
     \includegraphics[width=1.4cm]{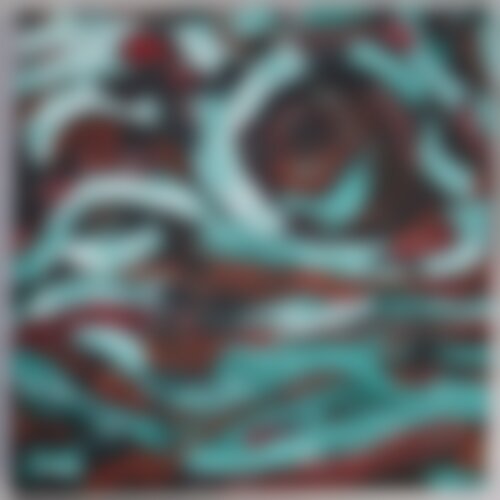} &
     \includegraphics[width=1.4cm]{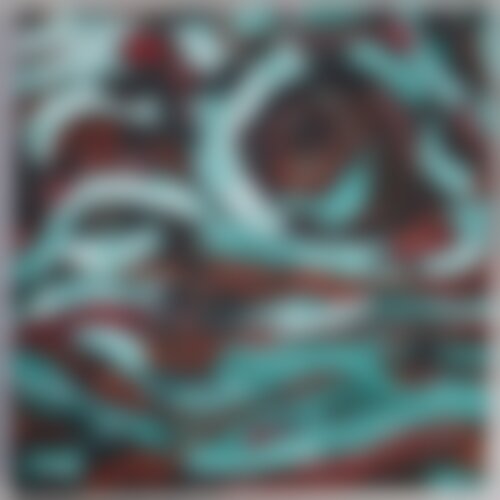} &
     \includegraphics[width=1.4cm]{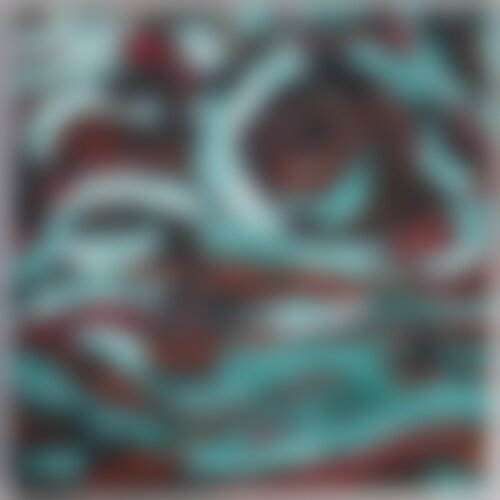} &
     \includegraphics[width=1.4cm]{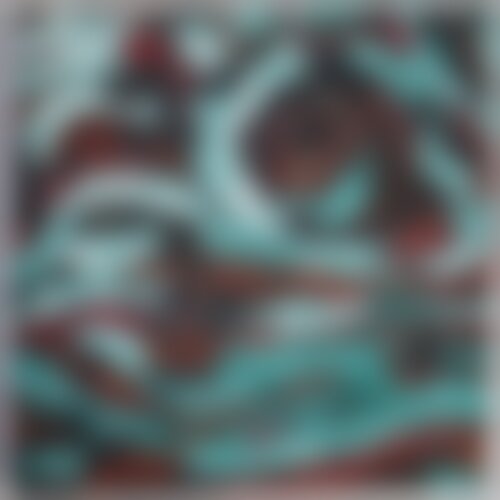} &
     \includegraphics[width=1.4cm]{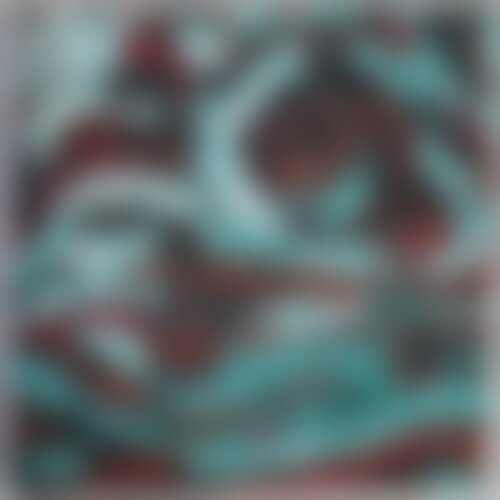}
     \vspace{3pt} \\
     \multicolumn{11}{l}{\textbf{tiling}} \\
     \includegraphics[width=1.4cm]{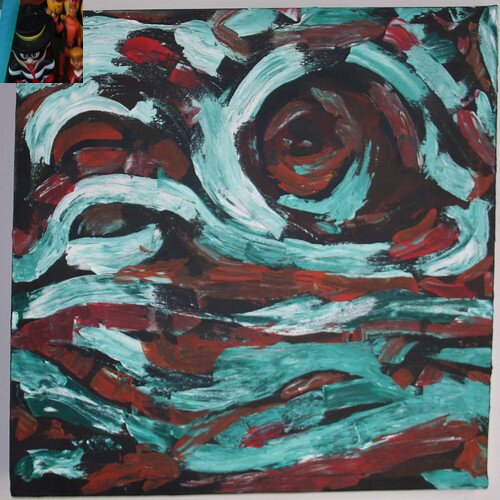} &
     \includegraphics[width=1.4cm]{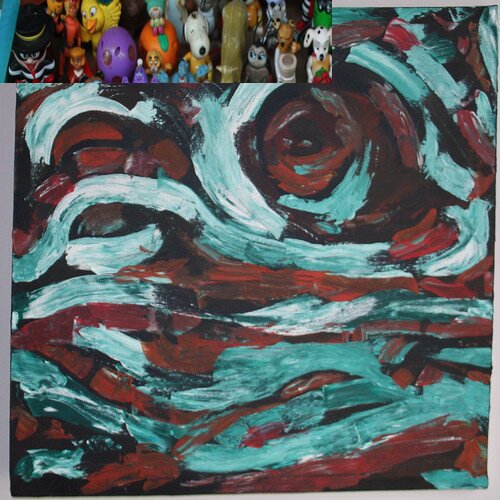} &
     \includegraphics[width=1.4cm]{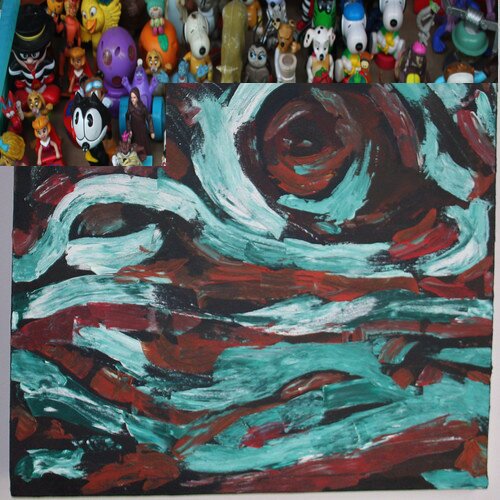} &
     \includegraphics[width=1.4cm]{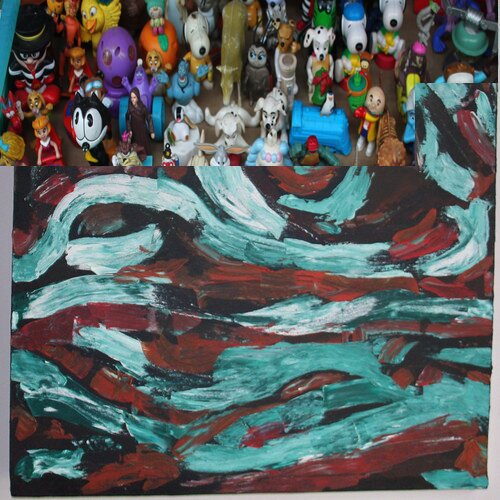} &
     \includegraphics[width=1.4cm]{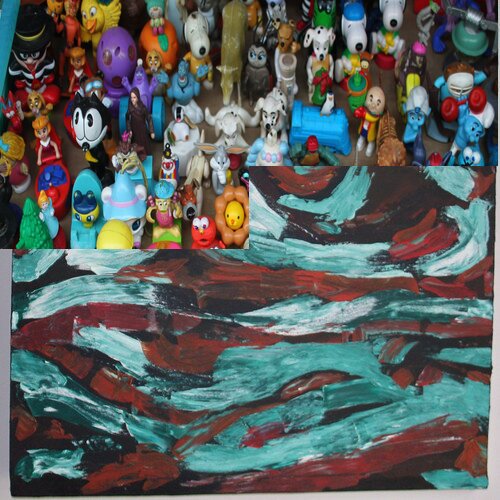} &
     \includegraphics[width=1.4cm]{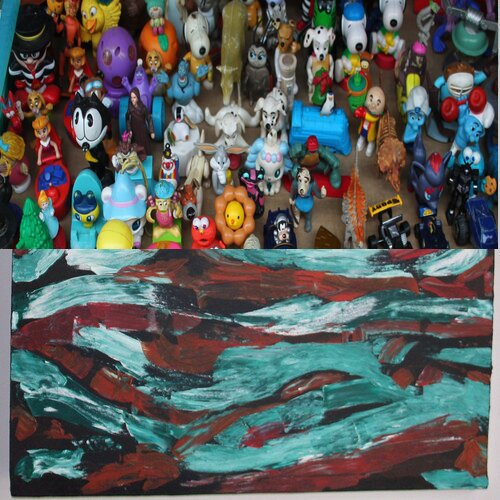} &
     \includegraphics[width=1.4cm]{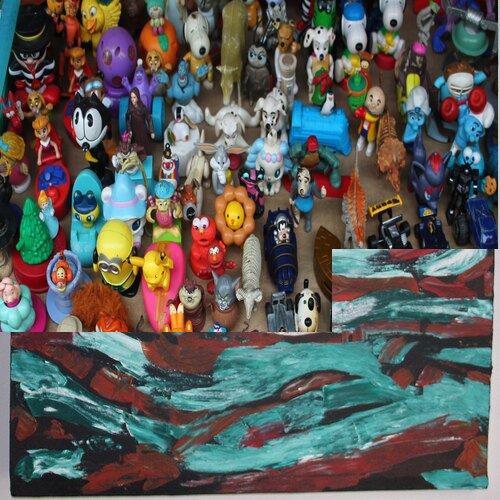} &
     \includegraphics[width=1.4cm]{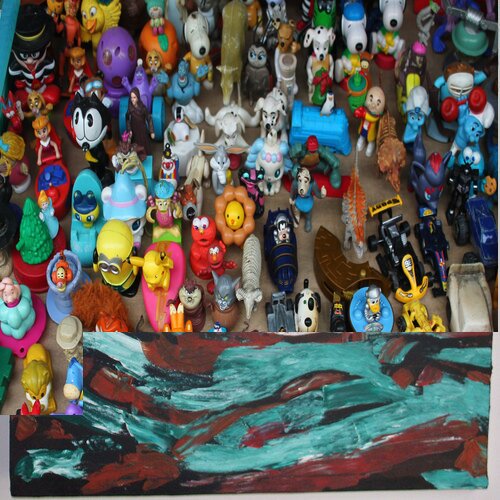} &
     \includegraphics[width=1.4cm]{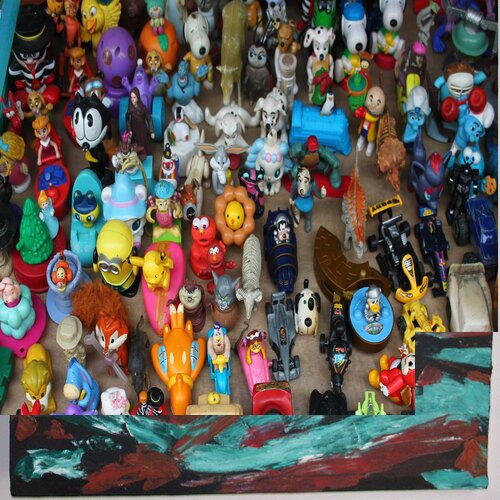} &
     \includegraphics[width=1.4cm]{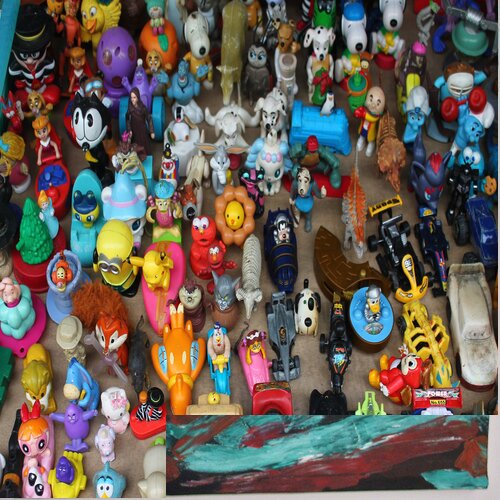} &
     \includegraphics[width=1.4cm]{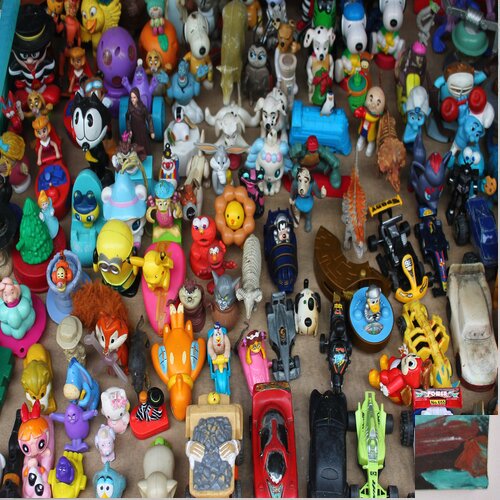}
     \vspace{3pt} \\
     \multicolumn{11}{l}{\textbf{noise}} \\
     \includegraphics[width=1.4cm]{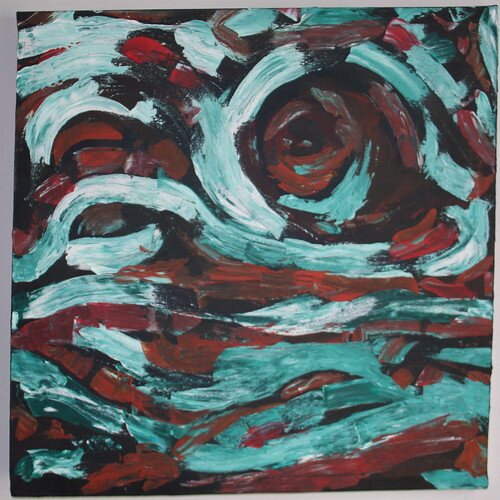} &
     \includegraphics[width=1.4cm]{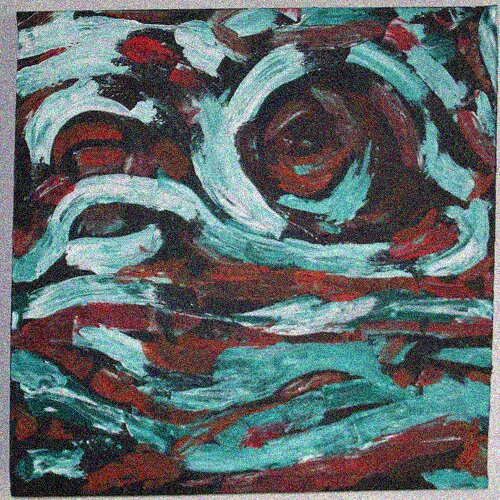} &
     \includegraphics[width=1.4cm]{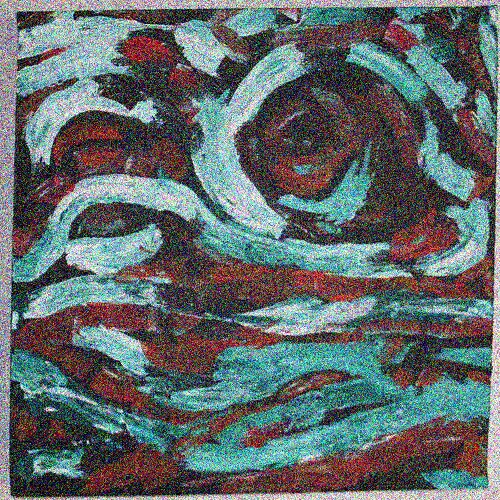} &
     \includegraphics[width=1.4cm]{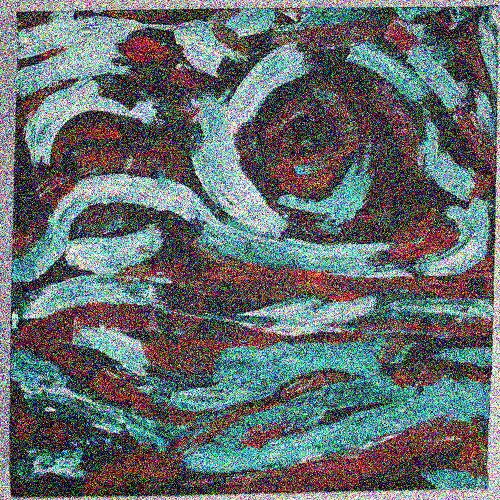} &
     \includegraphics[width=1.4cm]{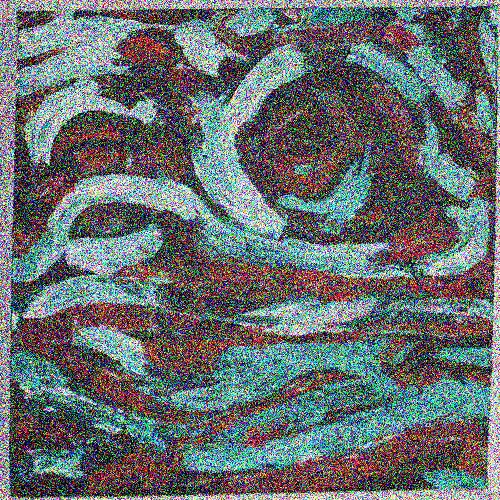} &
     \includegraphics[width=1.4cm]{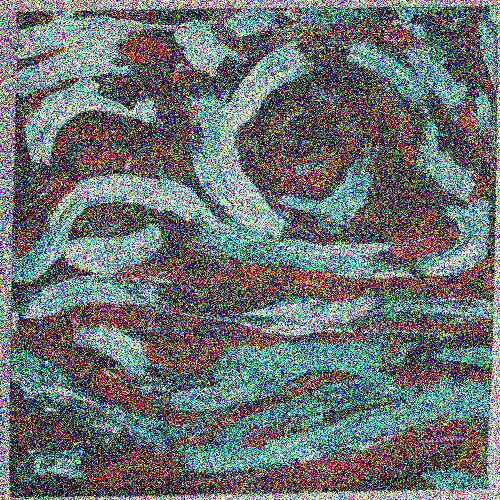} &
     \includegraphics[width=1.4cm]{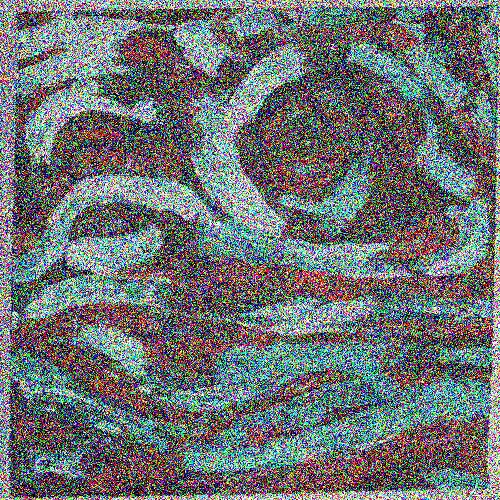} &
     \includegraphics[width=1.4cm]{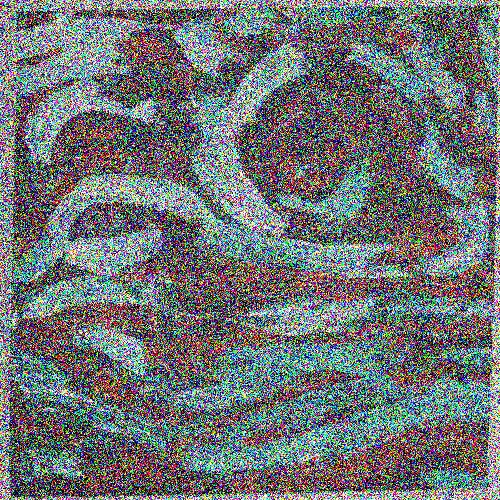} &
     \includegraphics[width=1.4cm]{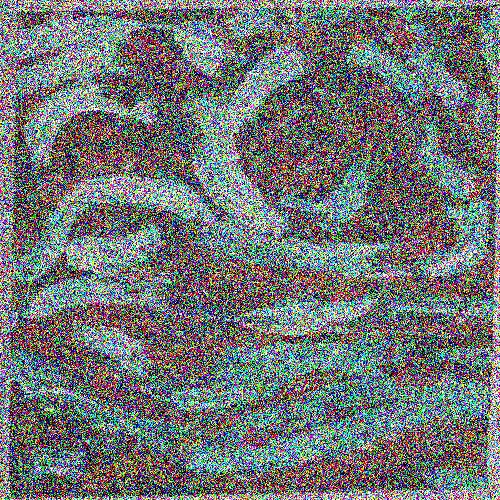} &
     \includegraphics[width=1.4cm]{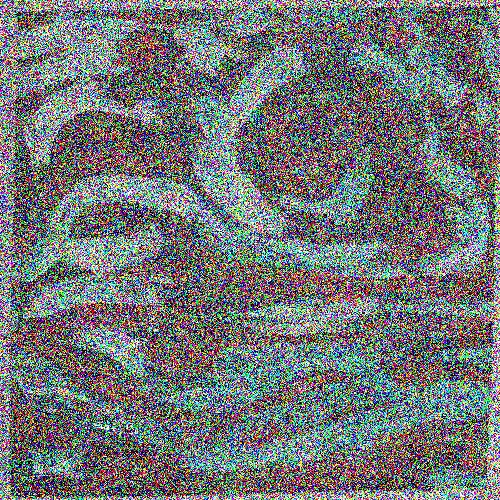} &
     \includegraphics[width=1.4cm]{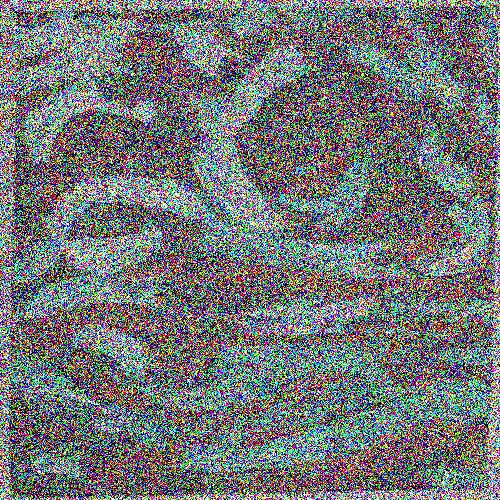}
     \vspace{3pt} \\
     \multicolumn{11}{l}{\textbf{clutter}} \\
     \includegraphics[width=1.4cm]{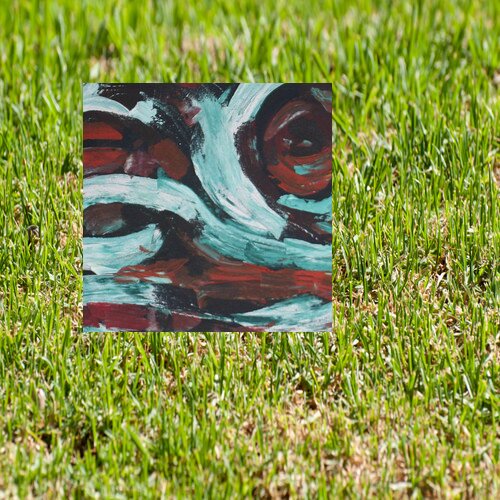} &
     \includegraphics[width=1.4cm]{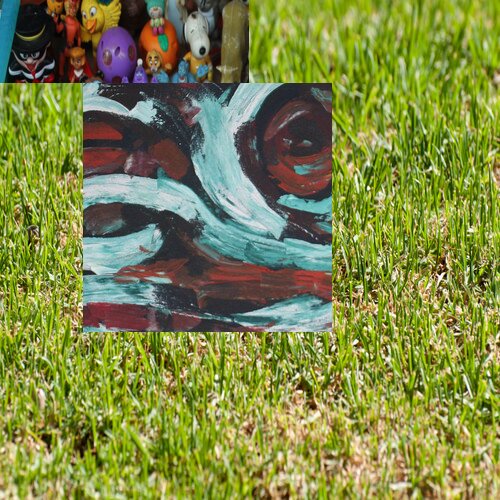} &
     \includegraphics[width=1.4cm]{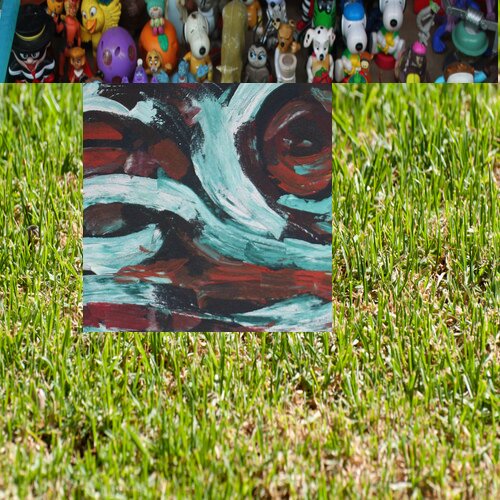} &
     \includegraphics[width=1.4cm]{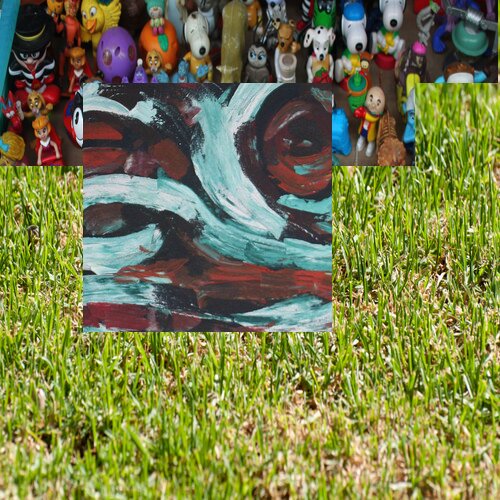} &
     \includegraphics[width=1.4cm]{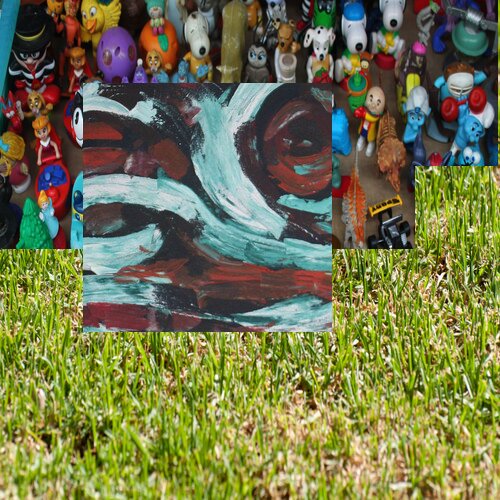} &
     \includegraphics[width=1.4cm]{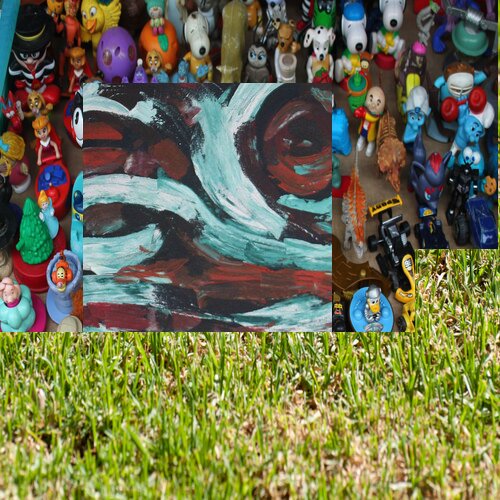} &
     \includegraphics[width=1.4cm]{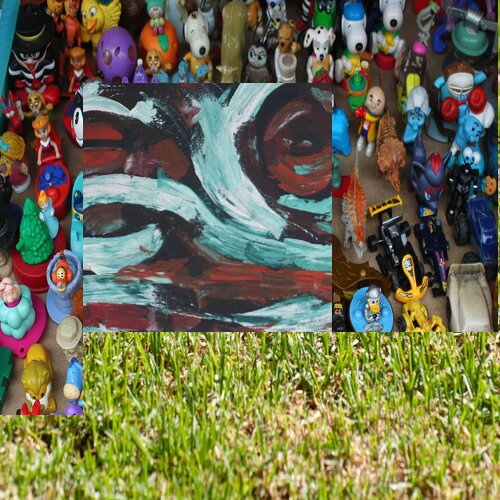} &
     \includegraphics[width=1.4cm]{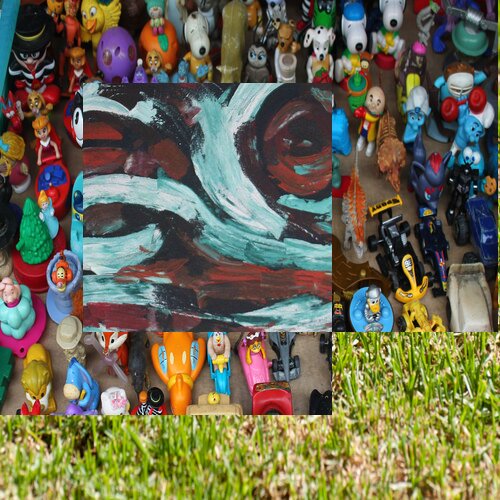} &
     \includegraphics[width=1.4cm]{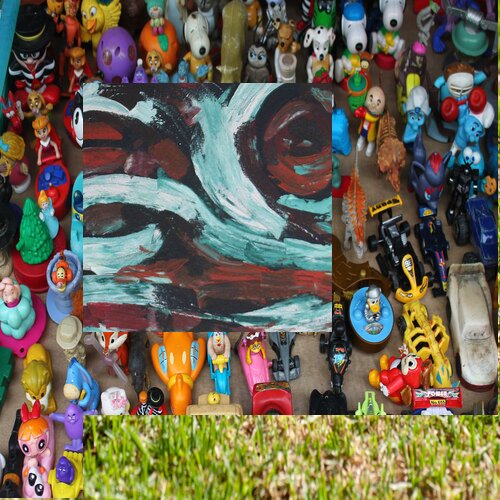} &
     \includegraphics[width=1.4cm]{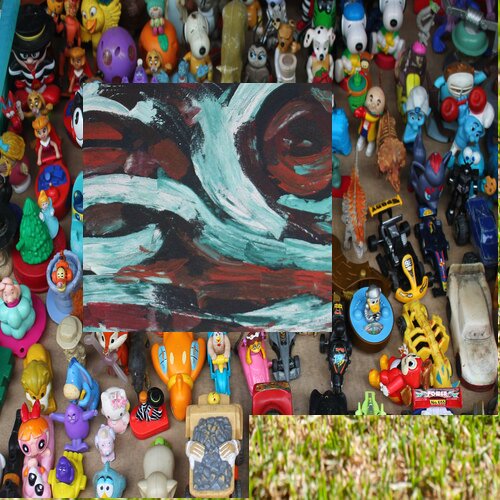} &
     \includegraphics[width=1.4cm]{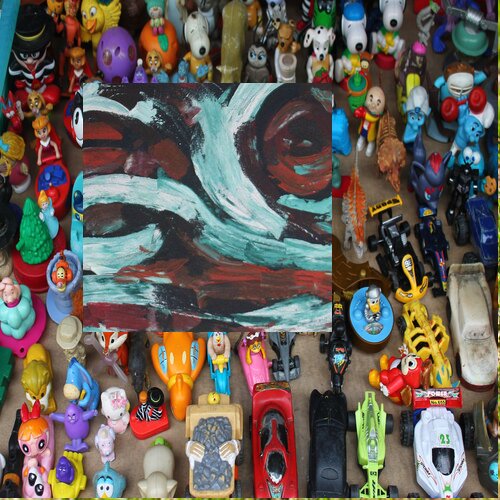}
     \vspace{3pt} \\
     \multicolumn{11}{l}{\textbf{occlusion}} \\
     \includegraphics[width=1.4cm]{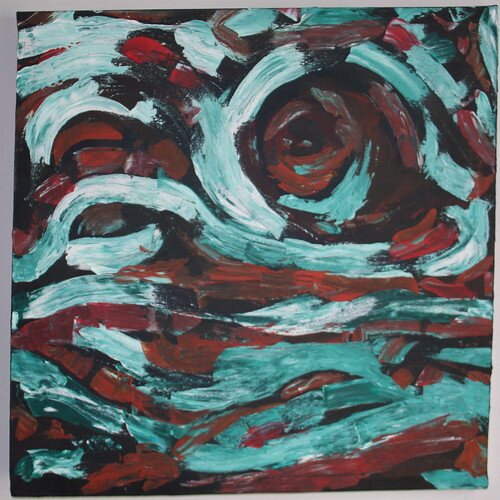} &
     \includegraphics[width=1.4cm]{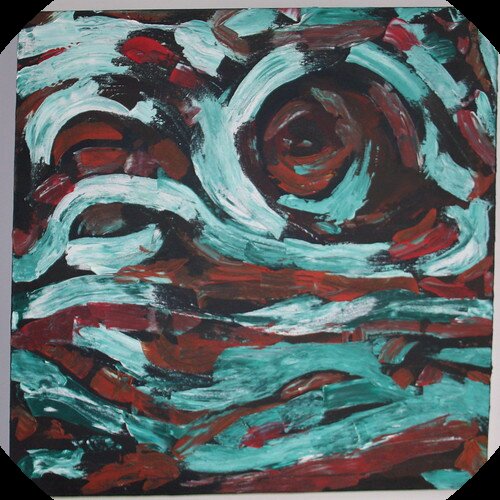} &
     \includegraphics[width=1.4cm]{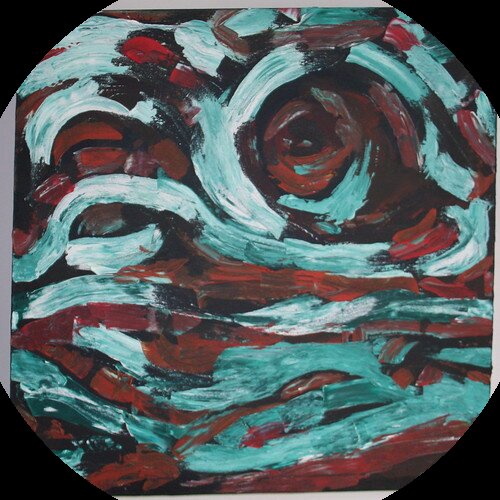} &
     \includegraphics[width=1.4cm]{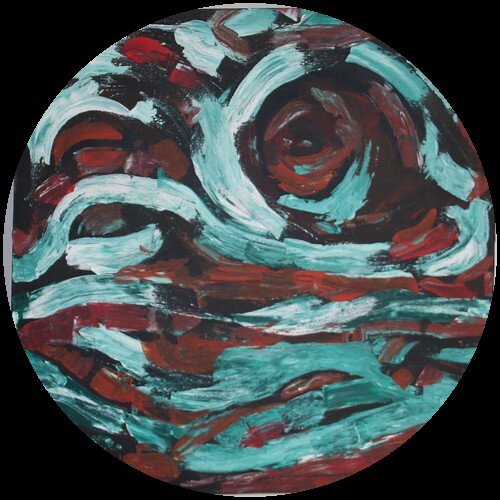} &
     \includegraphics[width=1.4cm]{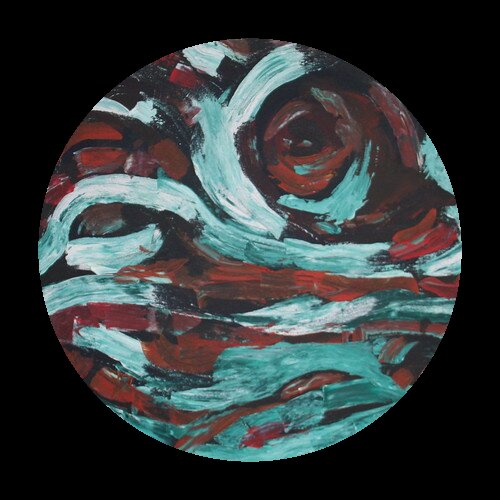} &
     \includegraphics[width=1.4cm]{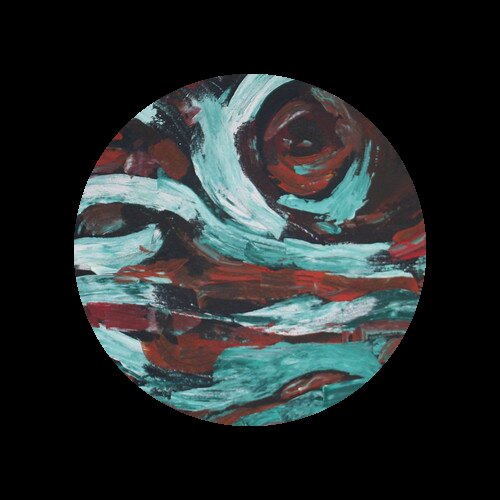} &
     \includegraphics[width=1.4cm]{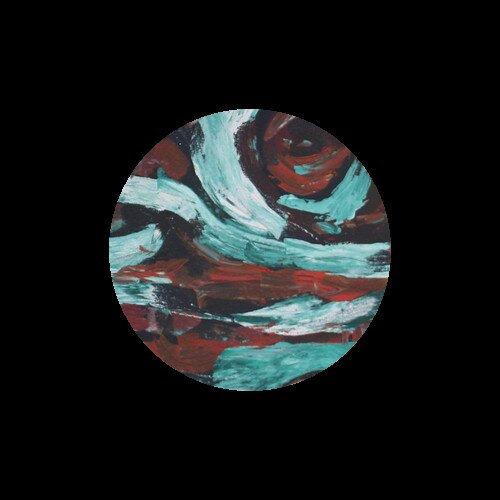} &
     \includegraphics[width=1.4cm]{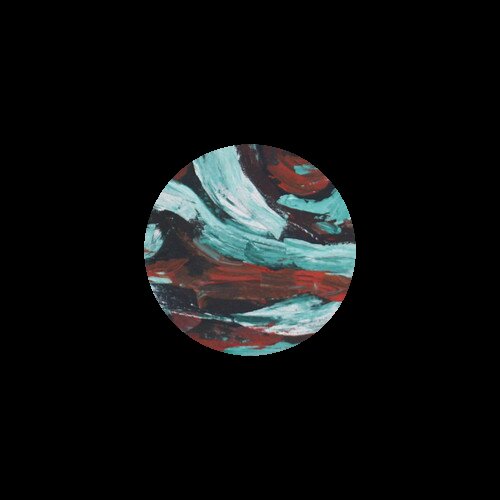} &
     \includegraphics[width=1.4cm]{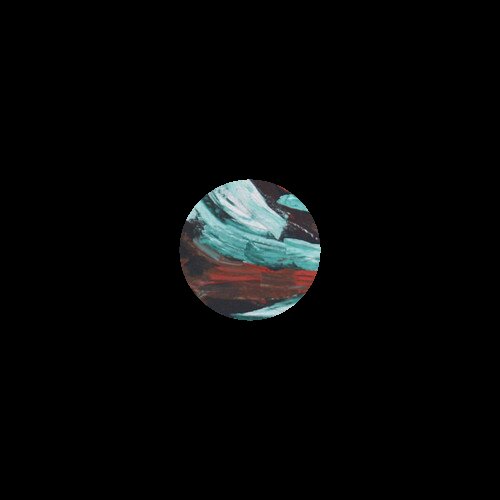} &
     \includegraphics[width=1.4cm]{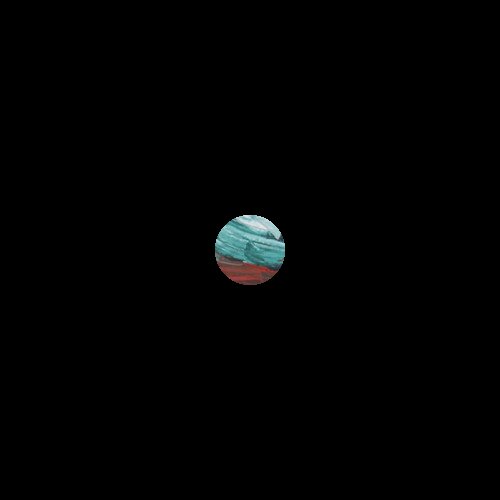} &
     \includegraphics[width=1.4cm]{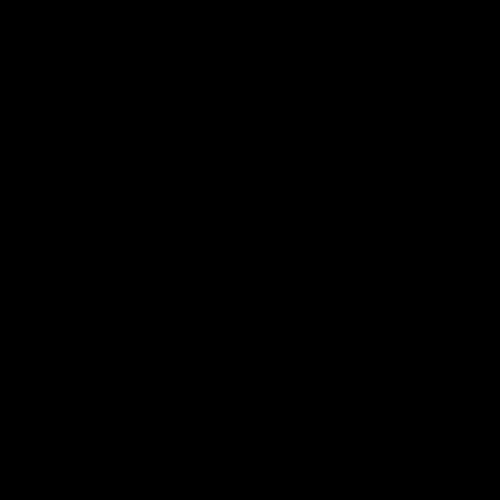}
\end{tabular}
    \caption{\textbf{Examples of transformed images for robustness analysis,} including variations in contrast, brightness, rotation, downscaling, background scaling, blur, tiling, noise, clutter, and occlusion. Each row shows the gradual increase in transformation strength from left to right. These transformations are applied to create positive query–target pairs for controlled robustness evaluation.
    \label{fig:transforms}
    }
\end{figure*}

We use a set of transformations designed to test a wide range of visual challenges, including photometric distortions, geometric and contextual changes, providing a comprehensive robustness evaluation across different retrieval approaches. \cref{fig:transforms} shows examples of the employed transformations. We always apply 20px zero-padding to prevent trivial self-similarity. The following is the list of transformations used in our robustness analysis:
\begin{itemize}
    \item \textbf{contrast:} adjust the image contrast using a scaling factor in the range $[0.05, 20]$, where values below 1 decrease the contrast and values above 1 enhance it.
    \item \textbf{brightness:} adjust the luminance of the image by adding or subtracting a brightness offset in the range $[0.05, 20]$.
    \item \textbf{rotation:} rotate the image by an angle from $0^\circ$ to $180^\circ$, maintaining the image center.
    \item \textbf{downscale:} resize the image by scale factors from $0.5$ to $0.05$, simulating extreme resolution loss.
    \item \textbf{scale-bg:} scales the object down and places it over a random background, with object-to-background ratios ranging from $0$ to $1$.
    \item \textbf{blur:} applies Gaussian blur with kernel sizes increasing from $\sigma=1$ to $\sigma=15$.
    \item \textbf{tiling:} inserts random distractor patches from another image, covering one patch of $1/6$ of the image area to the total area of the image.
    \item \textbf{noise:} adds Gaussian noise with standard deviation $\sigma$ varying from $0$ to $1.0$.
    \item \textbf{clutter:} similar to tiling, but it replaces the original background with a random scene. The object is merged onto this new background with an increase in the clutter density by adding $1$ to $28$ patches from another image.
    \item \textbf{occlusion:} apply circular black occluders that cover from $0\%$ to $100\%$ of the original image.
\end{itemize}

\begin{table}[t]
  \centering
  \hspace{0pt}
  \scalebox{1.}{
    \newcolumntype{C}{>{\centering\arraybackslash}p{6.0em}}

\begin{tabular}{lCC}
\toprule
\textbf{Model} & \textbf{Qwen3R} & \textbf{Qwen3R-C} \\
\midrule
~~\texttt{default}  & 60.2          & 58.8          \\
~~\texttt{object}   & \textbf{62.0} & \textbf{61.2} \\
\bottomrule
\end{tabular}
  }
  \vspace{0pt}
  \caption{\textbf{Qwen3R Prompting.} Retrieval performance (mAP) for Qwen3R and Qwen3R-C with the original generic \texttt{default} prompt, provided in the original repository, and with our task-specific \texttt{object} prompt.
  \vspace{-5pt}
  \label{tab:qwen3R-prompting}}
\end{table}

\section*{Acknowledgments}
This work was supported by the Junior Star GACR (grant no. GM 21-28830M), Horizon MSCA-PF (grant no. 101154126), CTU in Prague (grant no. SGS23/173/OHK3/3T/13), and the Czech National Recovery Plan—CEDMO 2.0 NPO (MPO 60273/24/21300/21000) provided by the Ministry of Industry and Trade. We acknowledge VSB – Technical University of Ostrava, IT4Innovations National Supercomputing Center, Czech Republic, for awarding this project (OPEN-36-1) access to the LUMI supercomputer, owned by the EuroHPC Joint Undertaking, hosted by CSC (Finland) and the LUMI consortium through the Ministry of Education, Youth and Sports of the Czech Republic through the e-INFRA CZ (grant ID: 90254).

{
    \small
    \bibliographystyle{ieeenat_fullname}
    \bibliography{main}
}

\end{document}